\documentclass[manuscript,screen]{acmart}

\AtBeginDocument{%
  \providecommand\BibTeX{{%
    \normalfont B\kern-0.5em{\scshape i\kern-0.25em b}\kern-0.8em\TeX}}}

\setcopyright{acmlicensed}
\copyrightyear{}
\acmYear{}
\acmDOI{XXXXXXX.XXXXXXX}

\newcommand{\lang}[1]{\textsc{#1}}
\usepackage{multirow}
\usepackage{graphicx}

\usepackage{subcaption}
\usepackage{booktabs}
\usepackage{float}
\usepackage{tablefootnote}
\usepackage{xcolor}
\usepackage{hyperref}
\usepackage{changepage}
\usepackage{makecell}
\usepackage{tabularray}
\usepackage[justification=centering]{caption}
\usepackage{placeins}

\usepackage{enumitem}
\newlist{parlist}{enumerate}{1}
\setlist[parlist]{label=(\alph*),wide=0pt,topsep=0pt}

\usepackage{amsmath}
\begin{document}

\title{Exploiting Domain-Specific Parallel Data on Multilingual
Language Models for Low-resource Language Translation}

\author{Surangika Ranathunga}
\email{s.ranathunga@massey.ac.nz}
\orcid{0003-0701-0204 }

\affiliation{%
  \institution{School of Mathematical and Computational Sciences, Massey University} 
  \city{Auckland}
  \country{New Zealand}
  \postcode{102904}
}
\author{Shravan Nayak}
\affiliation{%
  \institution{Mila – Quebec AI Institutes} 
  \city{}
  \country{Canada}
  \postcode{}
}

\author{En-Shiun Annie Lee}
\affiliation{%
  \institution{Computer Science, Ontario Technology University and University of Toronto} 
  \city{}
  \country{Canada}
  \postcode{}
}

\author{Xin Peng}

\author{Shih-Ting Cindy Huang}

\author{Yuchen Zeng}

\author{Yanke Mao }

\author{Tong Su}
\author{Yun-Hsiang Ray Chan}

\author{Songchen Yuan}

\author{Anthony Rinaldi}
\affiliation{%
  \institution{University of Toronto} 
  \city{}
  \country{Canada}
  \postcode{}
}

\renewcommand{\shortauthors}{Ranathunga, et al.}

\begin{abstract}
Neural Machine Translation (NMT) systems built on multilingual sequence-to-sequence Language Models (msLMs) fail to deliver expected results when the amount of parallel data for a language, as well as the language's representation in the model are limited. This {restricts} the capabilities of domain-specific NMT systems for low-resource languages (LRLs).  As a solution, parallel data from auxiliary domains can be used either to fine-tune or to further pre-train the msLM. We present an evaluation of the effectiveness of these two techniques in the {context} of domain-specific LRL-NMT. We also explore the impact of domain divergence on NMT model performance. We recommend several {strategies} for utilizing auxiliary parallel data in building domain-specific NMT models for LRLs.
\end{abstract}

\ccsdesc[500]{Computing methodologies~Machine translation}
\keywords{Neural Machine Translation, Domain divergence, multilingual Language Models, domain adaptation}

\received{xx}
\received[revised]{xx}
\received[accepted]{xx}

\maketitle

\section{Introduction}

The need to rapidly develop translation systems to disseminate information during pandemics, natural disasters, and other major events justifies the need to build domain-specific Neural Machine Translation (NMT) systems with any and all available parallel data. This is particularly true for low-resource languages (LRLs) that do not {possess} large amounts of parallel data for any single domain~\cite{zhou2021family}.
Therefore, when building a domain-specific NMT model for a given LRL pair, we face two possible scenarios: the considered (target, aka test) domain does not have any parallel data to be used in training, or it has a little amount of parallel data. In both cases, it is possible to make use of parallel data available for the same language pair, but in different (auxiliary) domains. This is not an unrealistic assumption - according to Tiedemann~\cite{tiedemann2012parallel}, hundreds of languages have parallel data from more than one domain.

Multilingual sequence-to-sequence Language Models (msLMs) such as mBART~\cite{tang2020multilingual} have been shown to outperform vanilla Transformer models for the LRL-NMT task~\cite{madaan2020transfer,thillainathan2021fine, lee-etal-2022-pre}. There are two possible ways to exploit auxiliary domain parallel data in building domain-specific LRL-NMT systems on top of msLMs: to further pre-train the msLM with self-supervised training objectives (\textbf{continuous pre-training})~\cite{reid-artetxe-2022-paradise} or to \textbf{fine-tune} the msLM with the NMT objective.  

\begin{figure}[!b]
    \centering
    \includegraphics[scale=0.60]{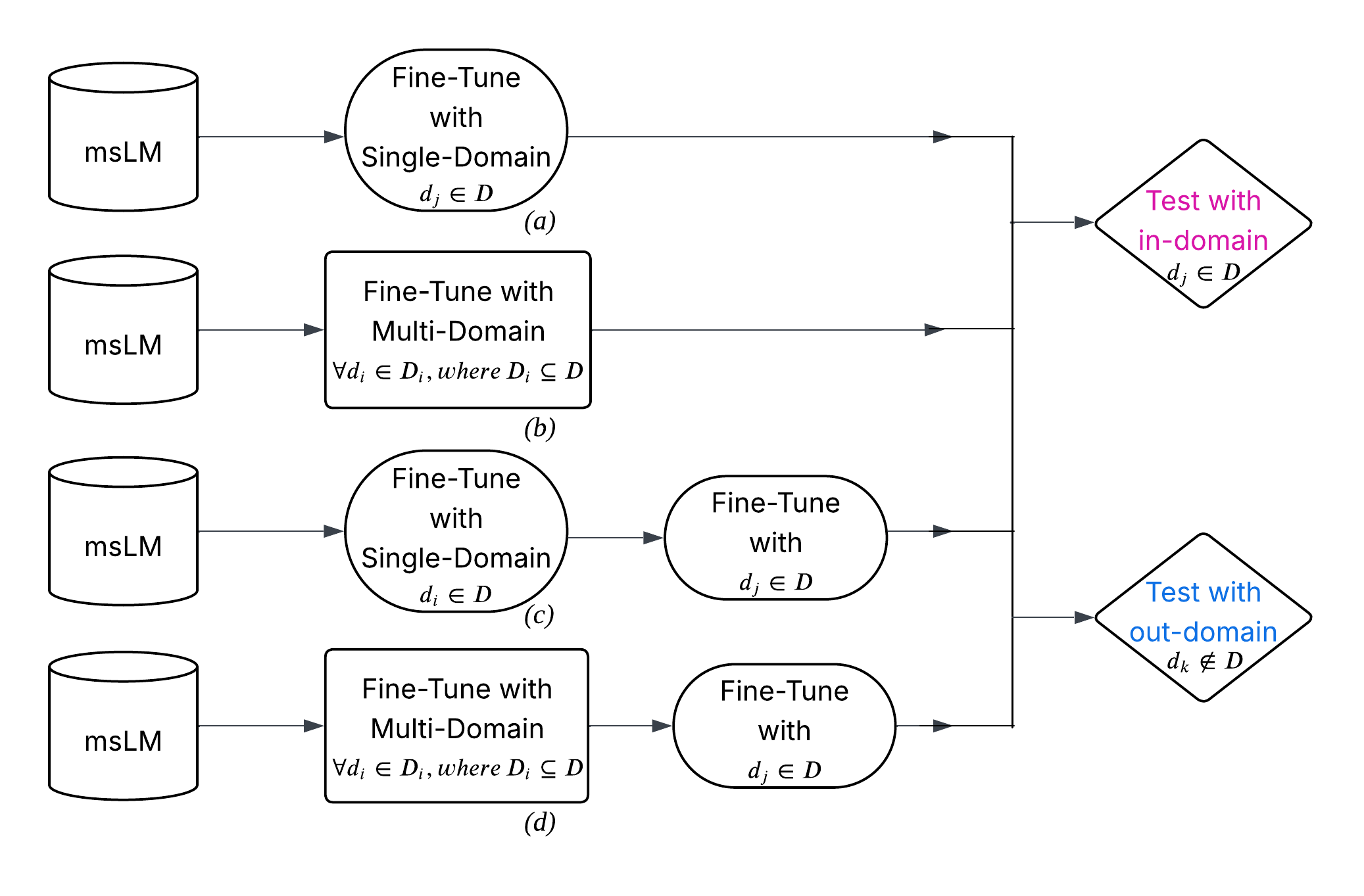}
    \caption{Fine-tuning Strategies. (a) Vanilla FT (b) Mixed-domain FT (c) Single-domain ITTL (d) Multi-domain ITTL. $D$- set of all domain-specific datasets per language pair. $d_i$, $d_j$, $d_k$ - domain-specific datasets.}
    \label{fig:multistage_ft_pt}
\end{figure}

As shown in Figure~\ref{fig:multistage_ft_pt}, there are different ways to use auxiliary domain data in fine-tuning. These can be categorized as single-stage fine-tuning (FT) and Intermediate Task Transfer Learning (ITTL)~\cite{phang2018sentence}. {Similarly, we can introduce another categorization based on the data availability of the target domain. In-domain refers to the case whether target domain has some parallel data for model training. Out-domain refers to the case where there is no such parallel data for the target domain. Note that out-domain testing scenario can be considered as zero-shot testing with respect to the target domain. }

Variations of single-stage FT are: 1) \textbf{single-domain (vanilla) FT}, where  the msLM is fine-tuned once with data from one domain, and 2) \textbf{multi-domain FT}, where the msLM is fine-tuned once with data from multiple domains. In vanilla FT, if the target domain has parallel data, this data is used for fine-tuning (in-domain case). In multi-domain FT, when the target domain has parallel data (in-domain case), it is combined with auxiliary domain data. In ITTL, the msLM is fine-tuned with an intermediate task, before fine-tuning with the final task. It also has two variations: \textbf{single-domain ITTL}, where the intermediate task has data from one auxiliary domain, and \textbf{multi-domain ITTL}, where the intermediate task has data from multiple domains. In both cases, the final task (i.e. stage\footnote{We use the term \textit{stage} interchangeably with \textit{task}, i.e. final task or final stage, and intermediate task or intermediate stage.}) fine-tuning is done using data from a single domain. If target domain has parallel data (in-domain case), the final task always uses this data. In the in-domain case, target domain data is also combined with auxiliary domain data in the intermediate stage of multi-domain ITTL.

Currently, there is no comparative study on the impact of continuous pre-training and fine-tuning techniques when auxiliary domain data is used. Determining the method that best exploits auxiliary domain parallel data is beneficial for LRL-NMT researchers who are unable to run extensive experiments due to limited computational resources. Lee et al.~\cite{lee-etal-2022-pre} and Adelani et al.~\cite{adelani2022few} identified that domain divergence in parallel datasets affects NMT performance. However, they did not conduct a detailed investigation into the impact of this factor. Khiu et al.~\cite{khiu2024predicting} and our previous work~\cite{nayak2023leveraging} evaluated the impact of domain divergence only in the case of single-domain ITTL.  Given the above shortcomings, this paper intends to answer the following questions related to domain-specific LRL-NMT systems built on top of msLMs: 

\begin{enumerate}
    \item Should we use auxiliary domain parallel data for {fine-tuning, continuous pre-training, or for both?}
    \item {What is the impact of domain divergence between different parallel data sets on different FT setups? }  
\end{enumerate}


In order to answer these questions, we conducted an extensive set of experiments with the aforementioned FT techniques as well as the bitext denoising pre-training technique~\cite{reid-artetxe-2022-paradise}. We considered the LRLs Sinhala (\lang{si}), Tamil (\lang{ta}), Kannada (\lang{ka}), and Gujarati (\lang{gu}) as well as Hindi (\lang{hi}) for English-centric NMT. We report results for both the in-domain and out-domain cases. We quantify the impact of domain divergence using Jenson-Shanon Divergence (JSD). 

Based on our analysis, we make the following observations in using parallel data from auxiliary domains when building NMT systems on top of msLMs for LRL pairs:  
\begin{itemize}
    \item When the parallel data set size is small (less than 25k, in our experiments), pre-training with bitext denoising yields no gains for both in- and out-domain setups.
    \item Multi-domain ITTL (with data from two domains in the intermediate step) is the best for in-domain, but its utility diminishes when the target dataset size increases. For out-domain, the best strategy depends on the domain {relatedness} and the size of parallel corpora.
    \item For both in- and out-domain setups, simply combining data from more than two domains does not help in ITTL, mainly due to the divergence between the domains they belong to.
\end{itemize}

\section{Related Work}

\subsection{ITTL on msLMs for NMT}
\label{itft_nmt}
ITTL has roots in Transfer Learning, which has been extensively explored for NMT with vanilla Transformer models~\cite{imankulova2019exploiting, gao2024novel}. However, as mentioned earlier, for LRLs, NMT systems built on top of msLMs outperform these traditional NMT techniques. Thus we do not discuss those Transfer Learning  techniques.  

ITTL~\cite{phang2018sentence, pruksachatkun2020intermediate, park2020scientific} refers to fine-tuning a pre-trained language model with the intermediate task(s), before fine-tuning on the final task. It has been extensively experimented with encoder-based LLMs such as BERT~\cite{phang2018sentence, moghe2021cross}, and has shown promising results even for LRLs~\cite{dhananjaya2024lexicon}. 

With respect to ITTL for NMT, the closest to our work is Adelani et al.~\cite{adelani2022few}. Similar to us, they explored single-domain ITTL and multi-domain ITTL, as well as multi-domain FT. However, they only considered the scenario where the auxiliary task has a large amount of data, while the target task has only a small amount of data. Moreover, they only used religious and news data and tested only for the in-domain setup. They did not quantify the impact of domain divergence. Our prior work~\cite{nayak2023leveraging} and Khiu et al.~\cite{khiu2024predicting} considered only single-domain ITTL. However, they considered both in- and out-domain test setups and investigated the impact of domain divergence. Verma et al.~\cite{verma2022strategies} experimented with single-domain ITTL and multilingual ITTL  (data from the same domain but in multiple languages). They found that the former is better, but did not quantify the impact of domain divergence.

\subsection{Pre-training msLMs with Parallel Data}
Reid and Artetxe~\cite{reid-artetxe-2022-paradise} augmented the existing denoising objective in mBART with three new objectives: replace words in the noised sequence with a bilingual dictionary, predict the reference translation instead of the input sequence (bitext denoising), and a combination of the two. Kale et al.~\cite{kale2021nmt5} used four denoising objective functions on mT5: translation language modeling, standard NMT, denoised NMT and denoised NMT with language model (LM). Chi et al.~\cite{chi2021mt6} used three cross-lingual objective functions on mT5: machine translation, translation pair span corruption, and translation span corruption. 
However, the latter two research did not consider the NMT task. 

\subsection{NMT for Low-resource Languages}
{In the context of the task of Machine Translation, a low-resource language refers to a language for which there are no sufficient parallel corpora to train Machine Translation models~\cite{ranathunga2021neural}. Thus, NMT results obtained for high-resource languages such as English and French cannot be obtained for such low-resource languages~\cite{haddow2022survey}. Therefore, several prominent solutions have been introduced over the past years~\cite{haddow2022survey,ranathunga2021neural}. As mentioned earlier, Transfer Learning is one such technique. Other techniques include unsupervised~\cite{artetxe2018unsupervised} and semi-supervised NMT~\cite{cheng2016semi}, data augmentation~\cite{tennage2018handling} and pivoting~\cite{cheng2019joint}.}

{Early NMT solutions were implemented on either recurrent neural network architectures, or the Transformer architecture~\cite{vaswani2017attention}, where the model is trained from scratch, for the task of NMT~\cite{imankulova2019exploiting}. However, with the introduction of pre-trained Language Models such as BERT~\cite{devlin2019bert} and BART~\cite{lewis2020bart} (as well as their multilingual counterparts mBERT and mBART), researchers were keen on exploiting such models in the context of low-resource NMT, to benefit from the cross-lingual transfer abilities of such models. While encoder models such as BERT were used as language model priors~\cite{baziotis2020language}, encoder-decoder models such as mBART could be directly fine-tuned for the NMT task~\cite{thillainathan2021fine}. With the revelation that such encoder-decoder models significantly outperform the vanilla Transformer models for low-resource NMT~\cite{lee-etal-2022-pre}, encoder-decoder models such as NLLB ( No Language Left Behind)~\cite{costa2022no}, which have been specifically trained for the task of NMT were released. With the introduction of Large Language Models (LLMs) such as ChatGPT~\cite{openai_chatgpt} and Claude~\cite{anthropic_claude}, researchers have conducted empirical studies on their effectiveness against the encoder-decoder models. However, LLMs have not been able to fully outperform encoder-decoder models such as NLLB, in the context of low-resource languages as yet~\cite{enis2024llm,robinson2023chatgpt}.}

\section{Methodology}
The baseline for all our experiments is fine-tuning the msLM once with data from one domain (vanilla FT or single-domain FT). In the in-domain setup, this dataset comes from the target domain.

\subsection{Continuous Pre-training}
For continuous pre-training, we experimented with two of the Reid and Artetxe's~\cite{reid-artetxe-2022-paradise} pre-training objective functions: bitext de-noising with auxiliary domain parallel data {alone ($l_{bitext}$), and  $l_{bitext}$ combined with monolingual denoising ($l_{mono}$)~\cite{reid-artetxe-2022-paradise}. For a given source-target sentence pair ($x, y$), $l_{bitext} = -log P(y|g_\Phi(x))$, where the likelihood of generating the target sentence $y$ is optimized, conditioned on the noised version of the source sentence $g_\Phi(x)$. The noising function removes spans of text and replaces them with a mask token.} For monolingual denoising, we used the source and target side of the parallel data separately and did not use additional monolingual data as done by Reid and Artetxe~\cite{reid-artetxe-2022-paradise}, because our objective is only to exploit the use of parallel data. We also excluded their dictionary denoising, as many LRLs still do not have sufficient bilingual dictionaries~\cite{liyanage2021bilingual}. {During pre-training, msLM is further trained, starting from its current weights.} Once msLM is pre-trained with Reid and Artetxe's objective functions~\cite{reid-artetxe-2022-paradise}, we continue with vanilla FT using the NMT objective function. 

\subsection{Fine-tuning}
{We use the four FT strategies (a, b, c, d) shown in Figure~\ref{fig:multistage_ft_pt}, where strategies c and d are extensions of strategies a and b, respectively:}

\begin{parlist}
    \item {Fine-tune the msLM with data from a single domain }
    \item {Fine-tune the msLM with data from multiple domains}
   \item {First fine-tune the msLM with data from a single domain, then fine-tune the resulting model with data from another single domain}
   \item {First fine-tune the msLM with data from multiple domains, then fine-tune the resulting model with data from a single domain}
\end{parlist}

{All parameters of the encoder, decoder, and embedding layers were updated during all the fine-tuning strategies; i.e.~no components were frozen. The model retained the original architecture with pre-layer normalization and embedding layer normalization.}

We formally define these strategies as follows:  Let $D$ be the set of all domains where there is parallel data available for model training for a considered language pair. $d_k \notin D$ is a domain for which there is no training data, but only test data is available. In the \textbf{single-domain FT (vanilla FT)} baseline (Figure~\ref{fig:multistage_ft_pt}(a)), the model is only fine-tuned with data from one domain $d_j \in D $. {In the in-domain case, both fine-tuning and testing is done using data from domain $d_j \in D $. In the out-domain case, fine-tuning uses data from $d_j $, but testing is done on domain $d_k$}. In \textbf{multi-domain FT} (Figure~\ref{fig:multistage_ft_pt}(b)), the model is fine-tuned with data from multiple domains $\forall d_i \in D$. {In the in-domain case, this includes data from the target domain $d_j$ as well}. In the out-domain case, this trained model is tested on the unseen target domain $d_k$. In \textbf{single-domain ITTL} (Figure~\ref{fig:multistage_ft_pt}(c)), msLM is first fine-tuned with data of domain $d_i \in D $. This is the \textit{intermediate task}, which is also called the \textit{first stage}. Next, this fine-tuned model is further fine-tuned with data from domain $d_j \in D $. This is the \textit{second and last stage}\footnote{The model can always be fine-tuned in more stages if there is data from different domains. However, we do not consider this option due to lack of computational resources.}. { In the in-domain case, testing will be done on the same domain $d_j$. In out-domain case, testing will be done on the unseen domain $d_k$}. In  \textbf{multi-domain ITTL} (Figure~\ref{fig:multistage_ft_pt}(d)), msLM is first fine-tuned with data from multiple domains in $D$. This model is further fine-tuned with $d_j \in D $, which is the target domain. {In the in-domain case, data of $d_j$ is included in the intermediate stage, as well as in the final stage.} In the out-domain case, testing will be done on the unseen domain $d_k$.  

\section{Experimental Setup}
{Our previous research was one of the first to explore the benefits of single-domain ITTL for LRL-NMT, considering the divergence across domains.  We reused the experimental results of single-domain ITTL (strategy c in Figure~\ref{fig:multistage_ft_pt}) of our previous research~\cite{nayak2023leveraging}. We used the same set of languages and the datasets as in our previous research, and since it experimentally showed that for those languages, mBART outperformed mT5~\cite{nayak2023leveraging}, we only experimented with the former.} {NLLB~\cite{costa2022no} was used for an ablation study.} 

We selected the LRL \lang{gu}, \lang{kn}, \lang{si}, \lang{ta} along with \lang{en} and \lang{hi} (see Table~\ref{tab:lang-details}). \lang{kn} is not included in mBART and is at a disadvantage compared to other languages~\cite{ranathunga2022some}. To fine-tune mBART for \lang{kn}, we used related language fine-tuning strategy~\cite{lee-etal-2022-pre} with Telugu being the related language\footnote{{mBART expects the input sentences to be annotated with the language ID. However, these IDs should be from the list of languages that mBART supports. Since Kannada is not supported in mBART, we used the language ID of Telugu to annotate Kannada sentences.}}.{ In order to examine whether this language selection has an impact on Kannada results, we carried out an ablation study using Tamil instead of Telugu. As shown in Table~\ref{tab:fine-tuning-kannada} in Appendix~\ref{sec:resulttables}, this language selection shows no impact on the Kannada results.}

\newcommand{\uniformtablesize}{\fontsize{5.3}{5.8}\selectfont} 
\renewcommand{\arraystretch}{1.15} 

\begin{table}[!htb]
\centering
\uniformtablesize
\setlength\tabcolsep{5pt}

\begin{tabular*}{0.75\textwidth}{@{\extracolsep{\fill}} l l l c r @{}}
\toprule
\textbf{Lang.} & \textbf{Family}      & \textbf{Script}     & \textbf{Joshi} & \textbf{mBART coverage}  \\
    &      &    & \textbf{class}  & \textbf{in Tokens (M)} \\
\midrule
 \lang{\lang{en}}       & Indo European  & Latin & 5 & 55608     \\
\lang{\lang{hi}}       & Indo Aryan  & Devanagari & 4 & 1715     \\
\lang{\lang{gu}} & Indo Aryan & Gujarati & 1 & 140  \\
\lang{\lang{kn}}       & Dravidian       & Kannada  & 1  & --   \\
\lang{\lang{\lang{si}}}       & Indo Aryan  & Sinhala  & 1  & 243      \\
\lang{\lang{ta}}       & Dravidian  & Tamil   & 3   & 595   \\
\bottomrule
\end{tabular*}
\caption{Language details. Smaller the Joshi et al.~\cite{joshi-etal-2020-state} class value, more low-resource the language is (source~\cite{nayak2023leveraging}). Number of tokens are reported in millions.
}
\label{tab:lang-details}
\end{table}

\noindent\textbf{Datasets:} We used both open-domain (CC) and domain-specific (Bible, Gvt, PMI) corpora (see Table~\ref{tab:data-details}). Each language has parallel data from three domains. The rather small FLORES dataset is used for out-domain testing only. Bible is the only domain-specific corpus with data for all our languages. The PMI and Gvt corpora are mutually exclusive in our experiments. Therefore, when describing results, we use PMI/Gvt to denote that we use one of these corpora for the considered experiment. Unlike PMI/Gvt and Bible, CC is larger, even for LRLs. However, it is a web-mined corpus, and is considered noisy~\cite{kreutzer-etal-2022-quality, ranathunga2024quality}. For multi-domain training, we randomly shuffled the training data from different domains.

\noindent\textbf{Evaluation Metrics:} We used SentencePiece BLEU (spBLEU) that was used in FLORES-101 evaluation benchmark~\cite{goyal-etal-2022-flores}.

\noindent\textbf{Domain Divergence:} We used Jenson-Shanon Divergence (JSD) to quantify domain divergence. Kashyap et al.~\cite{ramesh-kashyap-etal-2021-domain} showed that JSD is a very reliable measure to analyze the performance of a model in a new domain.  JSD values between our parallel corpora are given in Table~\ref{tab:result-domaindiv}. 

\begin{table}[!htb]
\centering
\uniformtablesize
\setlength\tabcolsep{5pt}
\renewcommand{\arraystretch}{1.05}

\begin{tabular*}{0.7\textwidth}{@{\extracolsep{\fill}} l c c c c @{}}
\toprule
\textbf{Dataset}     & \textbf{Gvt test} & \textbf{FLORES test} & \textbf{Bib test} & \textbf{PMI test} \\ \hline
Gvt train   & 0.10       & 0.40        & 0.58     & -       \\
CC train    & 0.40       & 0.41        & 0.52     & 0.44     \\
Bib train   & 0.56       & 0.47        & 0.10     & 0.53     \\
PMI train   & -          & 0.33        & 0.52     & 0.15     \\ \hline
\end{tabular*}
\caption{JS divergence between train and test domains (source~\cite{nayak2023leveraging}). The divergence is calculated for each (intermediate and fine-tuning) train and test set pair.}
\label{tab:result-domaindiv}
\end{table}

Test set specifications are in Table~\ref{tab:data-details}. Further details on the data sets, model training, evaluation metrics (including spBLEU signature), and JSD calculation are given in Appendix~\ref{sec:appendixA}.  We carried out experiments across techniques, languages, and domains for all \lang{En}-\lang{XX} pairs. \lang{XX}-\lang{En} experiments were carried out for \lang{si} and \lang{kn} only to keep the number of experiments at a manageable level.
\begin{table}[!htb]
\centering
\uniformtablesize
\setlength\tabcolsep{4pt}
\renewcommand{\arraystretch}{1.05}

\begin{tabular*}{0.8\textwidth}{@{\extracolsep{\fill}} l l l c c @{}}
\toprule
\textbf{Dataset}  & \textbf{Domain} & \textbf{Languages} & \textbf{Train Size} & \textbf{Test Size} \\ \hline
\textsc{Flores}-101 & Open   &  \lang{hi}, \lang{gu}, \lang{ka}, \lang{ta}  & - & 1k \\
\textsc{Flores}v1 & Open   & \lang{si}   & -   & 1k\\
CCAligned (CC) & Open   & \lang{all}   & 100k   & 1k \\
Government (Gvt) & Administrative   & \lang{si}, \lang{ta} & 50k  & 1k \\
PMIndia (PMI) & News   &  \lang{hi}  & 50k   & 1k\\
 &   &  \lang{gu}, \lang{ka} & 25k  & 1k \\
Web-scraped Bible  & Religious   &  \lang{all}   & 25k  & 1k \\
\hline
\end{tabular*}
\caption{Data set statistics (source~\cite{nayak2023leveraging}). No training set means the corresponding dataset was used only for testing. Two textsc{FLORES} versions had to be used because textsc{FLORESv1 includes on \lang{SI} out of the languages we studied.}}
\label{tab:data-details}
\end{table}

\section{Results and Discussion}
\subsection{Pre-training with Bi-text Denoising}
\newcommand{\cl}[1]{\textcolor{lightgray}{#1}}

\newlength\datgap
\setlength{\datgap}{2em}

Three languages were selected for experiments, considering their representation in mBART (Table~\ref{tab:lang-details}): \lang{kn}-not included in mBART, \lang{si}- 243M tokens and \lang{hi}- 1757M tokens. We tested for data sizes 1k, 10k, and 25k from individual domains and report the average results.
Table~\ref{tab:pre-train}\footnote{see Appendix~\ref{sec:pretraining} for raw results.} shows that neither of the pre-training techniques was able to surpass the results of vanilla FT on mBART, except in two cases where only marginal gains are reported over the original mBART. 

Interestingly, additional pre-training performed worst for Kannada, which is {unseen} during the original {pretraining of} mBART.
These results imply that bitext denoising is not a suitable strategy {for utilizing  auxiliary} domain parallel data when the data size is small. Reid and Artetxe~\cite{reid-artetxe-2022-paradise} reported results might improve if additional monolingual data is incorporated in large quantities; however, this is outside the scope of our research and we leave this for future research.  Since pre-training only with parallel data did not yield {satisfactory} results for vanilla FT, we did not carry out further experiments combining pre-training and ITTL. {However, it is possible to further investigate the impact of pre-training, by varying the dataset size, and the noising rate, etc.}  

\subsection{Impact of Data Set Size and Domain Divergence on FT and ITTL}
In these experiments, we varied the data set size of both the intermediate and final stages. The smallest size is 1k {(termed ``small'')} and the maximum size is 25k (termed ``large'')\footnote{These terms for dataset sizes are simply relative. In Ranathunga et al.'s~\cite{ranathunga2021neural}, even 25k parallel data is considered an extremely small data setup. However, the maximum data size we can use is bound by domain data availability.}, where 25k is the largest common size across all three domains (see Table~\ref{tab:data-details}). For clarity, we used a selected set of EN-XX experiment results for our discussion; full results for EN-XX and XX-EN are in Appendix~\ref{sec:resulttables}.


To identify the best-performing technique under different data setups, we {analyzed the results with respect to dataset sizes} of the intermediate (IM) and final (FI) stages in the following four combinations (recall that Adelani et al.~\cite{adelani2022few} considered only the large-small data setup):
1) \textbf{small-small } (IM data 1k - FI data 1k); 2) \textbf{large-small} (IM data 25k - FI data 1k); 3) \textbf{small-large}  (IM data 1k - FI data 25k); 4) \textbf{large-large}  (IM data 25k - FI data 25k).

\subsubsection{In-Domain Test Case}

In-domain results are reported in Tables~\ref{tab:1k-1k-in}-\ref{tab:25k-25k-in}. In this setup, we consider either PMI/Gvt or Bible as the target domain. {Baseline results are reported by training mBART with the target domain, and testing with the same. In ITTF experiments, the second stage of fine-tuning is done using this target domain data.} In each table, the best technique appears in bold, the second best appears in italic. Worst is grayed out.


\begin{table}[!htb]
\centering
\uniformtablesize
\setlength\tabcolsep{4pt}
\renewcommand{\arraystretch}{1.1}

\begin{tabular*}{0.7\textwidth}{@{\extracolsep{\fill}} l cc cc @{}}
\toprule

\multirow{2}{*}{\textbf{Language}} & \multicolumn{2}{c}{\textbf{out domain}} & \multicolumn{2}{c}{\textbf{in domain}} \\ \cline{2-5} 
                          & \textbf{bitex}       & \textbf{bitex+mono}       & \textbf{bitex}       & \textbf{bitex+mono}      \\ \midrule
\lang{ka}                        & -0.2        & -0.6             & -0.8        & -0.6            \\
\lang{si}                        & -0.2        & -0.4             & -0.5        & 0.3             \\
\lang{hi}                        & -0.2        & -0.5             & 0.3         & -0.4            \\ \hline
\end{tabular*}
\caption{Average difference  of NMT (spBLEU) results between continuously pre-trained mBART and original mBART. Results are averaged across different domains and dataset sizes.}
\label{tab:pre-train}
\end{table}

\begin{table*}[!htpb]
 \setlength{\datgap}{20pt}
\setlength\tabcolsep{3pt}
\centering
\uniformtablesize
\begin{tabular*}{\textwidth}{@{\extracolsep{\fill}} @{}ccccc @{\hspace{\datgap}} cccc @{}}
\hline
\multirow{2}{*}{\textbf{Language}} & \multicolumn{4}{c}{\makecell{\textbf{Intermediate task - PMI/Gvt} \\ \textbf{Final task - Bible}}} & \multicolumn{4}{c}{\makecell{\textbf{Intermediate task - Bible} \\ \textbf{Final task - PMI/Gvt}}} \\
\cmidrule(r{\datgap}){2-5} \cmidrule(r{\datgap}){6-8} \cmidrule(l){8-9}
 & \makecell{\textbf{Single} \\ \textbf{ITTL}} & \makecell{\textbf{Multi} \\ \textbf{FT}} & \makecell{\textbf{Multi} \\ \textbf{ITTL}} & \textbf{Baseline} & \makecell{\textbf{Single} \\ \textbf{ITTL}} & \makecell{\textbf{Multi} \\ \textbf{FT}} & \makecell{\textbf{Multi} \\ \textbf{ITTL}} & \textbf{Baseline} \\ \midrule
\lang{si} & \textbf{17.2} & \cl{16.7} & \cl{16.7} & \textit{17.0} & 20.9 & \textit{21.4} & \textbf{22.3} & \cl{20.2} \\ 
\lang{ta} & 11.0 & \textit{14.5} & \textbf{15.0} & \cl{9.3} & \textit{19.9} & 18.9 & \textbf{20.1} & \cl{18.6} \\
\lang{gu} & 13.0 & \textbf{13.3} & \textit{13.2} & \cl{12.9} & \cl{18.8} & 19.4 & \textit{20.1} & \textbf{20.9} \\
\lang{hi} & \textit{15.5} & 15.0 & \textbf{15.7} & \cl{14.0} & 18.3 & \textit{18.5} & \textbf{19.3} & \cl{18.2} \\
\lang{ka} & \textbf{9.4} & \textit{9.1} & 8.9 & \cl{8.2} & \cl{5.9} & 6.2 & \textbf{6.6} & \textit{6.5} \\
\hline
\textit{Avg} & 13.2 & \textit{13.7} & \textbf{13.9} & \cl{12.3} & \cl{16.8} & 16.9 & \textbf{17.7} & 16.9 \\ \hline
\end{tabular*}
\captionsetup[figure]{font=large}
\caption{NMT (spBLEU) results for the in-domain setup. Intermediate task dataset size -1k, final task dataset size-1k.}
\label{tab:1k-1k-in}
\end{table*}


\begin{table*}[!htb]
 \setlength{\datgap}{10pt}
\setlength\tabcolsep{5pt}
\begin{adjustwidth}{0cm}{}
\centering
\uniformtablesize
\begin{tabular*}{\textwidth}{@{\extracolsep{\fill}} @{}c cccc @{\hspace{\datgap}} cccc @{\hspace{\datgap}} cccc @{\hspace{\datgap}} cccc @{}}
\toprule
\multirow{2}{*}{\textbf{Language}} & \multicolumn{4}{c}{\makecell{\textbf{Intermediate task - PMI/Gvt} \\ \textbf{Final task - Bible}}} & \multicolumn{4}{c}{\makecell{\textbf{Intermediate task - Bible} \\ \textbf{Final task - PMI/Gvt}}} & \multicolumn{4}{c}{\makecell{\textbf{Intermediate task - CC} \\ \textbf{Final task - PMI/Gvt}}} & \multicolumn{4}{c}{\makecell{\textbf{Intermediate task - CC} \\ \textbf{Final task - Bible}}} \\
\cmidrule(r{\datgap}){2-5} \cmidrule(r{\datgap}){6-9} \cmidrule(r{\datgap}){10-13} \cmidrule(r{\datgap}){14-16} \cmidrule(l){16-17}
 & \makecell{\textbf{Single} \\ \textbf{ITTL}} & \makecell{\textbf{Multi} \\ \textbf{FT}} & \makecell{\textbf{Multi} \\ \textbf{ITTL}} & \textbf{Baseline} & \makecell{\textbf{Single} \\ \textbf{ITTL}} & \makecell{\textbf{Multi} \\ \textbf{FT}} & \makecell{\textbf{Multi} \\ \textbf{ITTL}} & \textbf{Baseline} & \makecell{\textbf{Single} \\ \textbf{ITTL}} & \makecell{\textbf{Multi} \\ \textbf{FT}} & \makecell{\textbf{Multi} \\ \textbf{ITTL}} & \textbf{Baseline} & \makecell{\textbf{Single} \\ \textbf{ITTL}} & \makecell{\textbf{Multi} \\ \textbf{FT}} & \makecell{\textbf{Multi} \\ \textbf{ITTL}} & \textbf{Baseline} \\ \midrule
\lang{si} & 18.9 & \textbf{19.4} & \textbf{19.4} & \cl{17.0} & \textit{22.1} & \cl{19.8} & \textbf{24.5} & 20.2 & \textit{24.1} & 23.7 & \textbf{25.8} & \cl{20.2} & \textbf{19.3} & 17.4 & \textit{18.8} & \cl{17.0} \\
\lang{ta} & 13.4 & \textit{13.5} & \textbf{14.5} & \cl{9.3} & \cl{18.3} & \cl{18.3} & \textbf{19.7} & 18.6 & 20.0 & 20.0 & \textbf{21.0} & \cl{18.6} & 14.5 & \textit{14.9} & \textbf{16.4} & \cl{9.3} \\
\lang{gu} & 14.3 & \textit{14.6} & \textbf{15.1} & \cl{12.9} & \cl{19.9} & 20.5 & \textbf{21.5} & \textit{20.9} & 23.8 & \textit{24.3} & \textbf{24.6} & \cl{20.9} & \textit{13.9} & 13.3 & \textbf{15.4} & \cl{12.9} \\
\lang{hi} & 15.3 & \textit{15.9} & \textbf{16.7} & \cl{14.0} & \textit{18.7} & \cl{18.1} & \textbf{19.2} & 18.2 & 20.0 & \textit{20.8} & \textbf{21.1} & \cl{18.2} & 16.1 & \textit{16.2} & \textbf{16.3} & \cl{14.0} \\
\lang{ka} & 12.7 & \textit{12.8} & \textbf{13.8} & \cl{8.2} & \textit{12.9} & 12.5 & \textbf{15.5} & \cl{6.5} & 12.9 & \textit{13.8} & \textbf{14.8} & \cl{6.5} & 9.6 & 9.6 & \textbf{10.0} & \cl{8.2} \\
\midrule
\textit{Avg} & 14.9 & \textit{15.2} & \textbf{15.9} & \cl{12.3} & \textit{18.4} & 17.8 & \textbf{20.1} & \cl{16.9} & 20.2 & \textit{20.5} & \textbf{21.5} & \cl{16.9} & \textit{14.7} & 14.3 & \textbf{15.4} & \cl{12.3} \\ \bottomrule
\end{tabular*}
\caption{NMT (spBLEU) results for the in-domain setup. Intermediate task dataset size- 25k, final task dataset size- 1k.}
\label{tab:25k-1k-in}
\end{adjustwidth}
\end{table*}

\begin{table*}[!htpb]
\setlength{\datgap}{20pt}
\setlength\tabcolsep{3pt}
\centering
\uniformtablesize
\renewcommand{\arraystretch}{1.0}
\begin{tabular*}{\textwidth}{@{\extracolsep{\fill}} @{}ccccc @{\hspace{\datgap}} cccc @{}}
\toprule
\multirow{2}{*}{\textbf{Language}} & \multicolumn{4}{c}{\makecell{\textbf{Intermediate task - PMI/Gvt} \\ \textbf{Final task - Bible}}} & \multicolumn{4}{c}{\makecell{\textbf{Intermediate task - Bible} \\ \textbf{Final task - PMI/Gvt}}} \\
\cmidrule(r{\datgap}){2-5} \cmidrule(r{\datgap}){6-8} \cmidrule(l){8-9}
 & \makecell{\textbf{Single} \\ \textbf{ITTL}} & \makecell{\textbf{Multi} \\ \textbf{FT}} & \makecell{\textbf{Multi} \\ \textbf{ITTL}} & \textbf{Baseline} & \makecell{\textbf{Single} \\ \textbf{ITTL}} & \makecell{\textbf{Multi} \\ \textbf{FT}} & \makecell{\textbf{Multi} \\ \textbf{ITTL}} & \textbf{Baseline} \\ \midrule
\lang{si} & \cl{37.6} & 37.7 & \textbf{38.2} &\textit{37.9} & \textit{45.0} & \cl{43.8} & \textbf{45.3} & 44.7 \\
\lang{ta} & \cl{30.5} & \textit{30.9} & 30.7 & \textbf{31.0} & 37.1 & \textit{37.7} & \textbf{38.1} & \cl{36.8} \\
\lang{gu} & \cl{27.3} & \textbf{27.8} & 27.7 & \textbf{27.8} & \textbf{38.0} & \cl{37.8} & \cl{37.8} & \textit{37.9} \\
\lang{hi} & \textit{31.5} & \textbf{31.7} & \cl{31.2} & 31.4 & 35.1 & \textbf{35.3} & \cl{34.9} & \textit{35.2} \\
\lang{ka} & \cl{27.0} & 27.4 & \textit{27.6} & \textbf{27.8} & 33.8 & 33.8 & \textbf{34} & \cl{33.6} \\
\midrule
\textit{Avg} & \cl{30.8} & \textit{31.1 } & 31.1 & \textbf{31.2 } & \textit{37.8} & 37.7 & \textbf{38.0} & \cl{37.6} \\ \bottomrule
\end{tabular*}
\caption{NMT (spBLEU) results for the in-domain setup. Intermediate task dataset size-1k, final task dataset size-25k.}
\label{tab:1k-25k-in}
\end{table*}

\vspace{-2mm}
\begin{table*}[!htb]
 \setlength{\datgap}{20pt}
\setlength\tabcolsep{3pt}
\centering
\uniformtablesize
\renewcommand{\arraystretch}{1.0}
\begin{tabular*}{\textwidth}{@{\extracolsep{\fill}} @{}c cccc @{\hspace{\datgap}} cccc @{\hspace{\datgap}} cccc @{\hspace{\datgap}} cccc @{}}
\toprule
\multirow{2}{*}{\textbf{Language}} & \multicolumn{4}{c}{\makecell{\textbf{Intermediate task - PMI/Gvt} \\ \textbf{Final task - Bible}}} & \multicolumn{4}{c}{\makecell{\textbf{Intermediate task - Bible} \\ \textbf{Final task - PMI/Gvt}}} & \multicolumn{4}{c}{\makecell{\textbf{Intermediate task - CC} \\ \textbf{Final task - PMI/Gvt}}} & \multicolumn{4}{c}{\makecell{\textbf{Intermediate task - CC} \\ \textbf{Final task - Bible}}} \\
\cmidrule(r{\datgap}){2-5} \cmidrule(r{\datgap}){6-9} \cmidrule(r{\datgap}){10-13} \cmidrule(r{\datgap}){14-16} \cmidrule(l){16-17}
 & \makecell{\textbf{Single} \\ \textbf{ITTL}} & \makecell{\textbf{Multi} \\ \textbf{FT}} & \makecell{\textbf{Multi} \\ \textbf{ITTL}} & \textbf{Baseline} & \makecell{\textbf{Single} \\ \textbf{ITTL}} & \makecell{\textbf{Multi} \\ \textbf{FT}} & \makecell{\textbf{Multi} \\ \textbf{ITTL}} & \textbf{Baseline} & \makecell{\textbf{Single} \\ \textbf{ITTL}} & \makecell{\textbf{Multi} \\ \textbf{FT}} & \makecell{\textbf{Multi} \\ \textbf{ITTL}} & \textbf{Baseline} & \makecell{\textbf{Single} \\ \textbf{ITTL}} & \makecell{\textbf{Multi} \\ \textbf{FT}} & \makecell{\textbf{Multi} \\ \textbf{ITTL}} & \textbf{Baseline} \\ \midrule
\lang{si} & \textbf{38.1} & \cl{37.9} & \cl{37.9} & \cl{37.9} & \textit{44.2} & \cl{43.7} & 44.0 & \textbf{44.7} & 44.4 & \cl{43.9} & \textbf{44.7} & \textbf{44.7} & \cl{37.9} & \textbf{38.0} & \cl{37.9} & \cl{37.9} \\
\lang{ta} & \textit{30.9} & \cl{30.4} & 30.7 & \textbf{31.0} & \textit{37.2} & 37.0 & \textbf{38.2} & \cl{36.8} & 37.2 & \textit{37.7} & \textbf{37.9} & \cl{36.8} & \textit{30.8} & \cl{30.6} & 30.7 & \textbf{31.0} \\
\lang{gu} & \cl{27.4} & 27.7 & \textbf{28.2} & \textit{27.8} & \cl{37.6} & 37.7 & \cl{37.6} & \textbf{37.9} & \textbf{38.3} & \cl{37.5} & 37.6 & \textit{37.9} & \cl{26.9} & 27.8 & \textbf{27.9} & 27.8 \\
\lang{hi} & \textbf{31.6} & \cl{31.4} & \cl{31.4} & \cl{31.4} & \cl{34.4} & 35.1 & 35.1 & \textbf{35.2} & \textbf{35.5} & \textit{35.4} & 35.3 & \cl{35.2} & \textbf{31.5} & 31.4 & \cl{31.1} & 31.4 \\
\lang{ka} & \textit{27.5} & \cl{26.2} & 27.2 & \textbf{27.8} & 32.8 & \cl{32.7} & \textbf{34.4} & \textit{33.6} & \textbf{34.3} & \cl{33.5} & \textit{33.7} & 33.6 & 27.3 & \cl{26.5} & \textit{27.4} & \textbf{27.8} \\
\midrule
\textit{Avg} & \textit{31.1 } & 30.7 & \textit{31.1} & \textbf{31.2} & 37.2 & 37.2 & \textbf{37.9} & \textit{37.6 } & \textbf{37.9} & \cl{37.6} & \textit{37.8} & 37.6 & 30.9 & \cl{30.9} & \textit{31.0} & \textbf{31.2} \\ \bottomrule
\end{tabular*}
\caption{NMT (spBLEU) results for the in-domain setup. Intermediate task dataset size- 25k, final task dataset size- 25k.}
\label{tab:25k-25k-in}
\end{table*}

\paragraph{small-small}
According to Table~\ref{tab:1k-1k-in}, multi-domain ITTL is the best-performing technique. However, on average, its gains over the second-best technique (multi-domain FT, which uses less compute) is less than 1 spBLEU. Hence, under limited computing resources, multi-domain FT is the best option.

\paragraph{large-small}
Comparing Table~\ref{tab:25k-1k-in} against Table~\ref{tab:1k-1k-in} (small-small setup), increasing the intermediate task size from 1k to 25k is {consistently} beneficial. In this data setup, multi-domain ITTL performs better than the other two techniques irrespective of the domain differences, thus confirming Adelani et al.'s~\cite{adelani2022few} observations (see Section~\ref{itft_nmt}).  

\paragraph{small-large and large-large}
According to Tables~\ref{tab:1k-25k-in} and~\ref{tab:25k-25k-in}, when there are 25k sentences from the target domain, none of the techniques managed to significantly outperform vanilla FT baseline, thus questioning the benefit of auxiliary domain data.

\begin{figure*}[!t]
     \centering
     \begin{subfigure}[b]{ \textwidth}
         \centering
         \includegraphics[width=\textwidth]{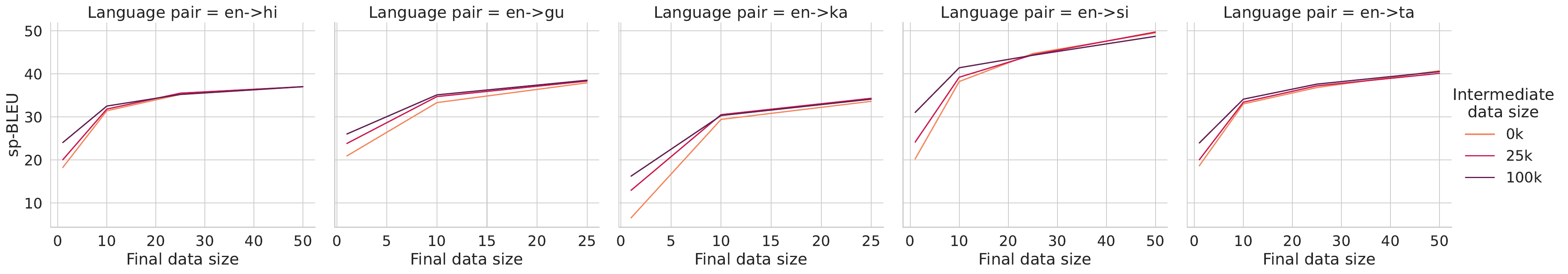}
         \vspace{-6mm}
        \caption{Method - single-domain ITTL. Intermediate dataset - CC, Final dataset - PMI/Gvt, Test dataset - PMI/Gvt.}
        \vspace{4mm}
     \end{subfigure}
     \begin{subfigure}[b]{ \textwidth}
         \centering
         \includegraphics[width=\textwidth]{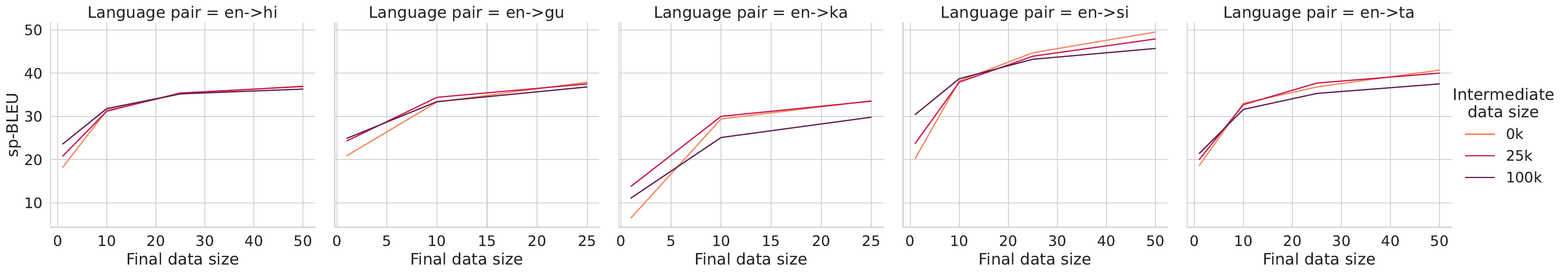}
         \vspace{-6mm}
        \caption{Method - multi-domain FT. Intermediate dataset - CC, Final dataset - PMI/Gvt, Test dataset - PMI/Gvt.}
        \vspace{4mm}
     \end{subfigure}
   \vspace{4mm}
     \begin{subfigure}[b]{ \textwidth}
         \centering
         \includegraphics[width=\textwidth]{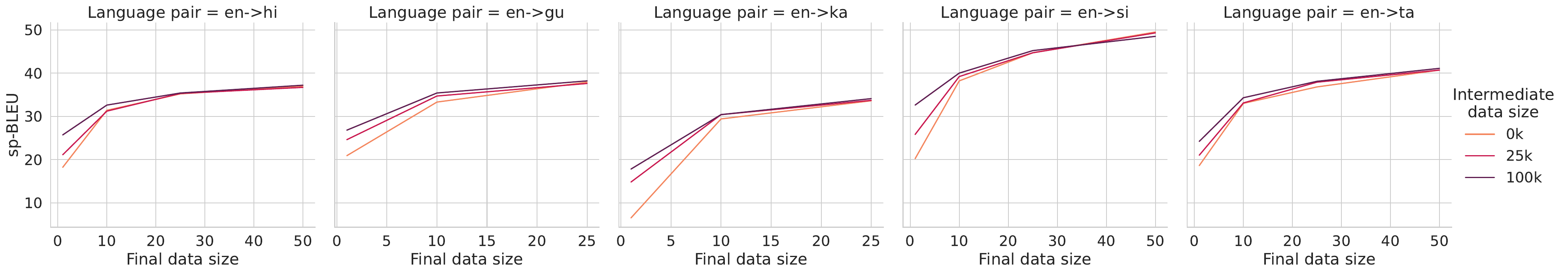}
         \vspace{-6mm}
        \caption{Method - multi-domain ITTL. Intermediate dataset - CC, Final dataset - PMI/Gvt, Test dataset - PMI/Gvt.}
     \end{subfigure}
      \vspace{-9mm}
     \caption{ NMT (spBLEU) results for different methods used in the in-domain setup.}
     \label{fig:mixed1-in-domain-cc}
\end{figure*}
The {above} observations are supported by Figure~\ref{fig:mixed1-in-domain-cc}  as well \footnote{Graphs for all the experiments are in ~\ref{sec:graphicalresults}.}. Note that the line corresponding to 0k in the graphs is the vanilla FT baseline. When the target (final) task size is less than 10k, all three techniques are effective over the vanilla FT baseline, but to varying degrees. However, {as} the target task data set size increases, their results converge to the baseline. Even using 100k CC data was not able to {make gains in performance} when the target domain data set is at least 25k. {Thus our results emphasize that Adelani et al.'s~\cite{adelani2022few} observations do not always hold, and it is important to consider the size of the datasets.}

An important observation is the results for \lang{ka}, which is {unseen} in mBART. Adding an intermediate task with 1k auxiliary data only results in a less than 1 spBLEU gain for this language in the small-small data setup. However, when adding 25k auxiliary data at the intermediate stage results in a significant gain (5.9 spBLEU on average) resulting in the largest gain among the languages. 

{In summary, for the small-small setup, multi-domain FT is the best option. For the large-small setup, multi-domain ITTL is the best. In both the small-large setup and the large-large setup, vanilla FT baseline is the best. Contrary to what Adelani et al.~\cite{adelani2022few} reported, our experiments confirm that the best model depends on the dataset size.}

\subsubsection{Out-domain Test Case}

\renewcommand{\arraystretch}{1.15} 
\setlength{\datgap}{20pt}

\begin{table*}[!htpb]
\setlength\tabcolsep{3pt}
\centering
\uniformtablesize
\begin{tabular*}{\textwidth}{@{\extracolsep{\fill}} @{}c cccc @{\hspace{\datgap}} cccc @{}}
\toprule
\multirow{2}{*}{\textbf{Language}} & \multicolumn{4}{c}{\makecell{\textbf{Intermediate task - PMI/Gvt} \\ \textbf{Final task - Bible}}} & \multicolumn{4}{c}{\makecell{\textbf{Intermediate task - Bible} \\ \textbf{Final task - PMI/Gvt}}} \\
\cmidrule(r{\datgap}){2-5} \cmidrule(r{\datgap}){6-8}  \cmidrule(l){8-9}
     & \makecell{\textbf{Single} \\ \textbf{ITTL}}      & \makecell{\textbf{Multi} \\ \textbf{FT}}      & \makecell{\textbf{Multi} \\ \textbf{ITTL}}     & \textbf{Baseline}     & \makecell{\textbf{Single} \\ \textbf{ITTL}}      & \makecell{\textbf{Multi} \\ \textbf{FT}}      & \makecell{\textbf{Multi} \\ \textbf{ITTL}}     & \textbf{Baseline}    \\ \midrule
\lang{si}   & 2.4     & \textbf{4.4}  & \textit{4.3}   & \cl{0.9}    & \textbf{4.9}     & 4.4  & \textit{4.5}   & \cl{3.9}    \\
\lang{ta}   & 2.2     & \textbf{3.7}  & \textit{3.6}   & \cl{0.7}    & \textbf{3.9}     & 3.7  & \textbf{3.9}   & \cl{2.6}    \\
\lang{gu}   & 6.0     & \textit{7.8}  & \textbf{7.9}   & \cl{2.2}    & \textbf{8.3}     & 7.8  & \textit{8.1}   & \cl{7.3}    \\
\lang{hi}   & 4.8     & \textbf{8.0}  & \textit{7.6}   & \cl{2.6}    & \textbf{8.4}     & 8.0  & \textit{8.2}   & \cl{7.1}    \\
\lang{ka}   & 0.5     & \textbf{1.1}  & \textbf{1.1}   & \cl{0.3}    & \cl{1.0}     & 1.1  & \textbf{1.2}   & \cl{1.0}    \\
\midrule
\textit{Avg}  & 3.2     & \textbf{5.0 }    & \textit{4.9}   & \cl{1.3}    & \textbf{5.3}     & 5.0  & \textit{5.2}   & \cl{4.4}    \\ \bottomrule
\end{tabular*}
\caption{NMT (spBLEU) results for the out-domain setup. Intermediate task dataset size-1k, final task dataset size-1k.}
\label{tab:1k-1k-out}
\end{table*}

\begin{table*}[!htb]
\setlength\tabcolsep{3pt}
\centering
\uniformtablesize
\begin{tabular*}{\textwidth}{@{\extracolsep{\fill}} @{}c cccc @{\hspace{\datgap}} cccc @{}}
\hline
\multirow{2}{*}{\textbf{Language}} & \multicolumn{4}{c}{\makecell{\textbf{Intermediate task - PMI/Gvt}\\\textbf{Final task - Bible}}} & \multicolumn{4}{c}{\makecell{\textbf{Intermediate task - Bible} \\ \textbf{Final task - PMI/Gvt}}} \\
\cmidrule(r{\datgap}){2-5} \cmidrule(r{\datgap}){6-8}  \cmidrule(l){8-9}
     & \makecell{\textbf{Single} \\ \textbf{ITTL}}      & \makecell{\textbf{Multi} \\ \textbf{FT}}      & \makecell{\textbf{Multi} \\ \textbf{ITTL}}     & \textbf{Baseline}     & \makecell{\textbf{Single} \\ \textbf{ITTL}}      & \makecell{\textbf{Multi} \\ \textbf{FT}}      & \makecell{\textbf{Multi} \\ \textbf{ITTL}}     & \textbf{Baseline}     \\ \midrule
\lang{si}   & \cl{1.9}     & \textbf{4.6}  & 2.8   & \cl{1.9}    & \textbf{11.3}     & \textbf{11.3}  & \textbf{11.3}   & \cl{11.2}    \\
\lang{ta}   & 2.2     & \textbf{3.8}  & \textit{2.9}   & \cl{2.1}    & 9.5     & 9.5  & \textbf{9.8}   & \cl{9.0}    \\
\lang{gu}   & 5.2     & \textbf{9.4}  & \textit{9.0}   & \cl{4.4}    & \textbf{20.2}     & \cl{19.8}  & 20.0   & \cl{19.8}    \\
\lang{hi}   & 3.4     & \textbf{6.9}  & \textit{6.5}   & \cl{3.2}    & \cl{16.9}     & \textit{17.4}  & \textbf{17.8}   & 17.3    \\
\lang{ka}   & 2.3     & \textbf{3.9}  & \textit{3.2}   & \cl{2.2}    & \cl{14.0}     & \textbf{14.5}  & \textbf{14.5}   & 14.1    \\
\hline
\textit{Avg}  & 3.0     & \textbf{5.7  }   & \textit{4.9}   & \cl{2.8}    & 14.4     & \textit{14.5}  & \textbf{14.7}   & \cl{14.3}    \\ \hline
\end{tabular*}
\caption{NMT (spBLEU) results for the out-domain setup. Intermediate task dataset size-1k, final task dataset size-25k.}
\label{tab:1k-25k}
\end{table*}

\begin{table*}[!htb]
\setlength\tabcolsep{3pt}
\centering
\uniformtablesize
\begin{tabular*}{\textwidth}{@{\extracolsep{\fill}} @{}c cccc @{\hspace{\datgap}} cccc @{\hspace{\datgap}} cccc @{\hspace{\datgap}} cccc @{}}
\toprule
\multirow{2}{*}{\textbf{Language}} & \multicolumn{4}{c}{\makecell{\textbf{Intermediate task - PMI/Gvt} \\ \textbf{Final task - Bible}}} & \multicolumn{4}{c}{\makecell{\textbf{Intermediate task - Bible} \\ \textbf{Final task - PMI/Gvt}}}& \multicolumn{4}{c}{\makecell{\textbf{Intermediate task - CC} \\ \textbf{Final task - PMI/Gvt}}} & \multicolumn{4}{c}{\makecell{\textbf{Intermediate task - CC} \\ \textbf{Final task - Bible}}} \\
\cmidrule(r{\datgap}){2-5} \cmidrule(r{\datgap}){6-9} \cmidrule(r{\datgap}){10-13} \cmidrule(r{\datgap}){14-16}  \cmidrule(l){16-17}
     & \makecell{\textbf{Single} \\ \textbf{ITTL}}      & \makecell{\textbf{Multi} \\ \textbf{FT}}      & \makecell{\textbf{Multi} \\ \textbf{ITTL}}     & \textbf{Baseline}     & \makecell{\textbf{Single} \\ \textbf{ITTL}}      & \makecell{\textbf{Multi} \\ \textbf{FT}}      & \makecell{\textbf{Multi} \\ \textbf{ITTL}}     & \textbf{Baseline} & \makecell{\textbf{Single} \\ \textbf{ITTL}}      & \makecell{\textbf{Multi} \\ \textbf{FT}}      & \makecell{\textbf{Multi} \\ \textbf{ITTL}}     & \textbf{Baseline}     & \makecell{\textbf{Single} \\ \textbf{ITTL}}      & \makecell{\textbf{Multi} \\ \textbf{FT}}      & \makecell{\textbf{Multi} \\ \textbf{ITTL}}     & \textbf{Baseline}   \\ \midrule
\lang{si}   & 7.9     & \textbf{11.3}    & \textbf{11.3}  & \cl{0.9}    & \textbf{5.8}     & 4.2  & \textit{5.6}   & \cl{3.9}    & 8.9     & \textit{9.7}  & \textbf{10.6}   & 3.9  & 3.5      & \textbf{8.8}   & \textit{8.4}    & \cl{0.9}  \\
\lang{ta}   & 5.4     & \textbf{9.4}  & \textit{8.7}   & \cl{0.7}    & \textbf{4.9}     & 3.8  & \textit{4.3}   & \cl{2.6}    & 10.1    & \textit{12.2}    & \textbf{12.4}   & \cl{2.6}  & 7.2      & \textbf{10.8}  & \textit{10.6}   & \cl{0.7}  \\
\lang{gu}   & 15.3    & \textbf{19.8}    & \textit{19.7}  & \cl{2.2}    & \textbf{10.5}    & 9.3  & \textit{10.2}  & \cl{7.3}    & \textbf{16.8}    & 16.1    & \textit{16.5}   & \cl{7.3}  & 9.6      & \textbf{12.2}  & \textit{11.9}   & \cl{2.2}  \\
\lang{hi}   & 12.8    & \textbf{17.1}    & \textbf{17.1}  & \cl{2.6}    & \textbf{8.8}     & \cl{7.1}  & 8.2   & \cl{7.1}    & 13.3    & \textbf{16.3}    & \textit{16.0}   & \cl{7.1}  & 7.1      & \textbf{14.9}  & \textbf{14.9}   & \cl{2.6}  \\
\lang{ka}   & 7.6     & \textbf{14.3}    & \textit{13.7}  & \cl{0.3}    & \textit{4.5}     & 4.4  & \textbf{5.1}   & \cl{1.0}    & 4.5     & \textit{4.7}  & \textbf{5.6}    &  \cl{1.0}  & 0.9      & \textit{1.7}   & \textbf{1.9}    & \cl{0.3}  \\
\midrule
\textit{Avg}  & 9.8    & \textbf{14.4}    & \textit{14.1}  & \cl{1.34}   & \textbf{6.9}     & 5.8  & \textit{6.7}   & \cl{4.4}    & 10.7    & \textit{11.8}    & \textbf{12.2}   & \cl{4.4}  & 5.7      & \textbf{9.7}   & \textit{9.5}    & \cl{1.3}      \\ \bottomrule
\end{tabular*}
\caption{NMT (spBLEU) results for the out-domain setup. Intermediate task dataset size- 25k, final task dataset size- 1k. }
\label{tab:25k-1k-out}
\end{table*}
\begin{table*}[!htb]
\setlength\tabcolsep{3pt}
\centering
\uniformtablesize
\begin{tabular*}{\textwidth}{@{\extracolsep{\fill}} @{}c cccc @{\hspace{\datgap}} cccc @{\hspace{\datgap}} cccc @{\hspace{\datgap}} cccc @{}}
\toprule
\multirow{2}{*}{\textbf{Language}} & \multicolumn{4}{c}{\makecell{\textbf{Intermediate task - PMI/Gvt} \\ \textbf{Final task - Bible}}} & \multicolumn{4}{c}{\makecell{\textbf{Intermediate task - Bible} \\ \textbf{Final task - PMI/Gvt}}} & \multicolumn{4}{c}{\makecell{\textbf{Intermediate task - CC} \\ \textbf{Final task - PMI/Gvt}}} & \multicolumn{4}{c}{\makecell{\textbf{Intermediate task - CC} \\ \textbf{Final task - Bible}}} \\
\cmidrule(r{\datgap}){2-5} \cmidrule(r{\datgap}){6-9} \cmidrule(r{\datgap}){10-13} \cmidrule(r{\datgap}){14-16}  \cmidrule(l){16-17}
     & \makecell{\textbf{Single} \\ \textbf{ITTL}}      & \makecell{\textbf{Multi} \\ \textbf{FT}}      & \makecell{\textbf{Multi} \\ \textbf{ITTL}}     & \textbf{Baseline}     & \makecell{\textbf{Single} \\ \textbf{ITTL}}      & \makecell{\textbf{Multi} \\ \textbf{FT}}      & \makecell{\textbf{Multi} \\ \textbf{ITTL}}     & \textbf{Baseline} & \makecell{\textbf{Single} \\ \textbf{ITTL}}      & \makecell{\textbf{Multi} \\ \textbf{FT}}      & \makecell{\textbf{Multi} \\ \textbf{ITTL}}     & \textbf{Baseline}     & \makecell{\textbf{Single} \\ \textbf{ITTL}}      & \makecell{\textbf{Multi} \\ \textbf{FT}}      & \makecell{\textbf{Multi} \\ \textbf{ITTL}}     & \textbf{Baseline}      \\ \midrule
\lang{si}   & 2.4     & \textbf{11.6}    & \textit{11.5}  & \cl{1.9}    & \textbf{11.9}     & \cl{9.9}  & 10.5   & \textit{11.2}    & 12.1     & \textbf{14.1}  & \textit{13.3}   & \cl{11.2}  & 2.0      & \textbf{9.5}   & \textit{8.8}    & \cl{1.9}  \\
\lang{ta}   & 2.8     & \textbf{9.9}  & \textit{6.6}   & \cl{2.1}    & 10.0     & \textit{11.5}  & \textbf{11.8}   & \cl{9.0}    & 11.1    & \textbf{14.8}    & \textit{14.7}   & \cl{9.0}  & 2.9      & \textbf{11.0}  & \textit{6.1}   & \cl{2.1}  \\
\lang{gu}   & 8.0    & \textbf{20.8}    & \textit{20.0}  & \cl{4.4}    & 20.5    & \textbf{20.8}  & \textit{20.6}  & \cl{19.8}    & 21.5    & \textit{23.7}    & \textbf{24.1}   & \cl{19.8}  & 5.6      & \textit{13.3}  & \textbf{14.6}   & \cl{4.4}  \\
\lang{hi}   & 4.3    & \textbf{17.6}    & \textit{16.5}  & \cl{3.2}    & \textbf{18.4}     & \cl{17.2}  & \textit{18.0}   & 17.3    & 18.3    & \textbf{21.5}    & \textit{21.3}   & \cl{17.3}  & 3.8      & \textbf{14.5}  & \textit{12.6}   & \cl{3.2}  \\
\lang{ka}   & 3.2     & \textbf{14.3}    & \textit{8.6}  & \cl{2.2}    & 14.2     & \textit{14.3}  & \textbf{15.5}   & \cl{14.1}    & 15.0     & \textbf{16.6}  & \textit{16.2}    & \cl{14.1}  & 2.4      & \textbf{4.2}   & \textit{3.1}    & \cl{2.2}  \\
\midrule
\textit{Avg}  & 4.1    & \textbf{14.8}    & \textit{12.6}  & \cl{2.8}   & \textit{15.0}     & 14.7  & \textbf{15.2}   & \cl{14.2}    & 15.6    & \textbf{18.1}    & \textit{17.9}   & \cl{14.3}  & 3.3      & \textbf{10.5}   & \textit{9.0}    & \cl{2.8}      \\ \bottomrule
\end{tabular*}
\caption{NMT (spBLEU) results for the out-domain setup. Intermediate task dataset size- 25k, final task dataset size- 25k.}
\label{tab:25k-25k-out}
\end{table*}

We use the results for the FLORES test set to explain our observations in Tables~\ref{tab:1k-1k-out}-\ref{tab:25k-25k-out}. In ITTL experiments, either PMI/Gvt or Bible is used as the final stage of fine-tuning. PMI/Gvt, Bible, and/or CC are used at the intermediate stage\footnote{Unlike domain-specific datasets s.a Bible, CC corpus is larger even for LRLs. Thus we do not run 1k setup for CC.}. Here, baseline refers to vanilla FT with the dataset used for the final task {(e.g. if Bible is used as the final task in the ITTL experiments, baseline refers to vanilla FT with Bible and testing with FLORES)}.

Unlike in the in-domain test case, in the out-domain test case, there exists a large impact of domain divergence. To begin with, vanilla FT baseline  {performs} significantly better (average gain 11.5 BLEU) when PMI/Gvt is used for fine-tuning, compared to using Bible. This observation can be explained with the domain divergence results in Table~\ref{tab:result-domaindiv}, {where} the divergence between FLORES and Bible ($0.47$) is higher than that between FLORES and PMI/Gvt ($0.33$/$0.4$).  

Similarly, for all four data size setups, the best results from both the ITTL strategies are reported when a domain more closer to FLORES is used in final stage fine-tuning.  The best result is reported when both the intermediate dataset and the final dataset are from domains closer to the target domain (e.g.~ CC and PMI/Gvt).

\begin{figure*}[!ht]
     \centering
    
     \begin{subfigure}[b]{ \textwidth}
         \centering
         \includegraphics[width=\textwidth]{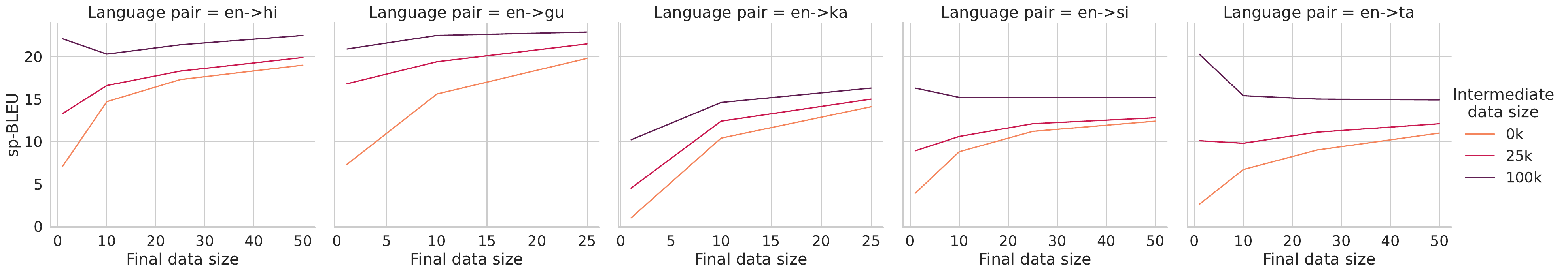}
        \vspace{-6mm}
        \caption{Method - single-domain ITTL. Intermediate dataset - CC, final dataset - PMI/Gvt, test set - FLORES.}
        \label{out-a}
        \vspace{4mm}
     \end{subfigure}
     \begin{subfigure}[b]{ \textwidth}
         \centering
         \includegraphics[width=\textwidth]{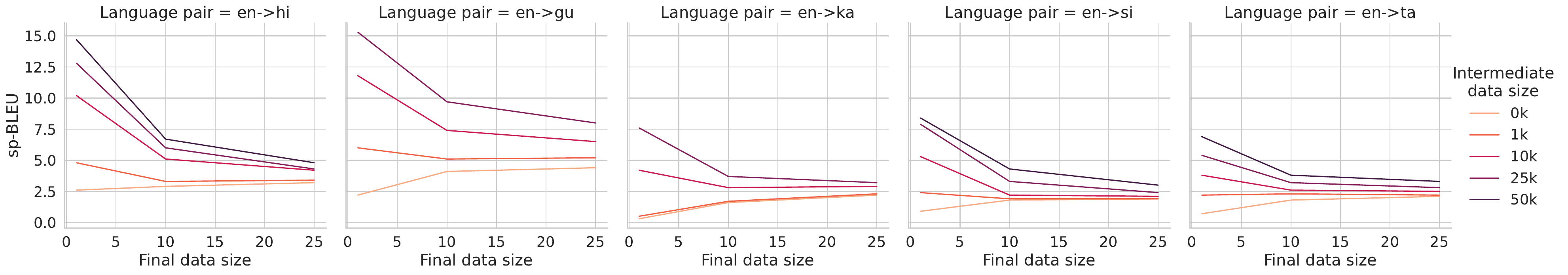}
         \vspace{-6mm}
        \caption{Method - single-domain ITTL. Intermediate dataset - PMI/Gvt, final dataset - Bible, test set - FLORES.}
        \label{out-b}
        \vspace{4mm}
     \end{subfigure}
     \begin{subfigure}[b]{ \textwidth}
         \centering
         \includegraphics[width=\textwidth]{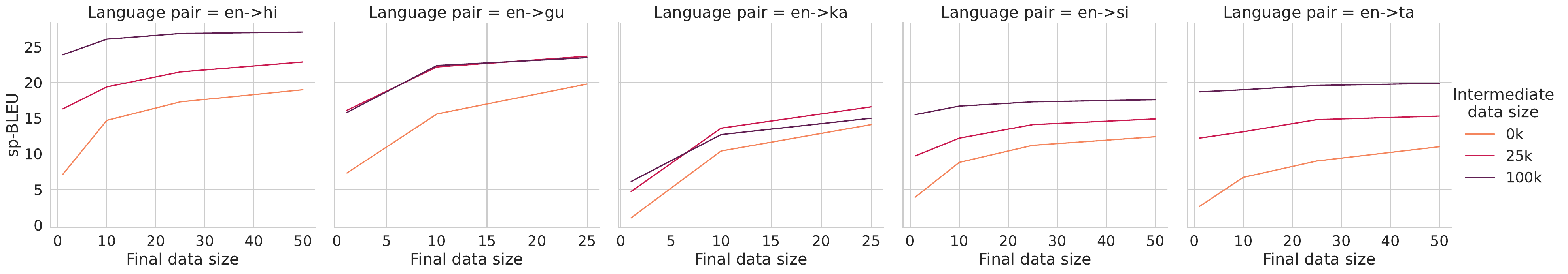}
         \vspace{-6mm}
        \caption{Method - multi-domain FT. Intermediate dataset - CC, final dataset - PMI/Gvt, test set - FLORES.}
        \vspace{4mm}
     \end{subfigure}
     \begin{subfigure}[b]{ \textwidth}
         \centering
         \includegraphics[width=\textwidth]{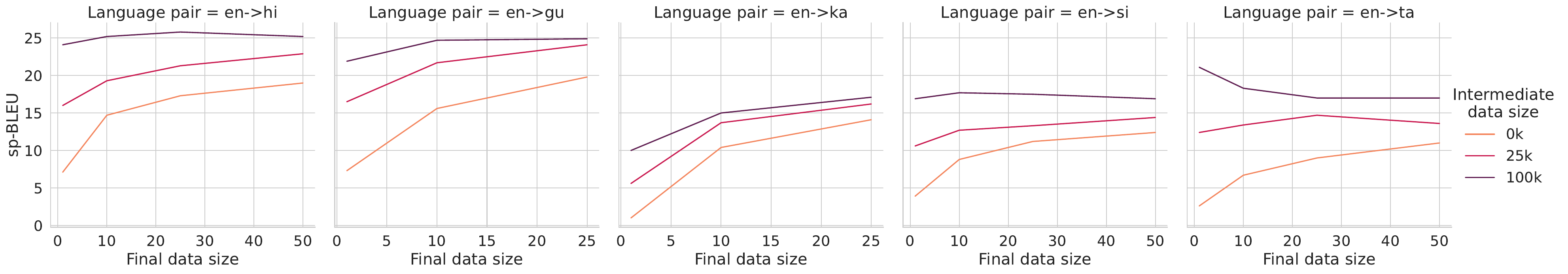}
         \vspace{-6mm}
        \caption{Method - multi-domain ITTL. Intermediate dataset - CC, final dataset - PMI/Gvt, test set - FLORES.}
        \vspace{4mm}
     \end{subfigure}
     \vspace{-9mm}
     \caption{ NMT (spBLEU) results for different methods used in the out-domain setup.}
     \label{fig:mixed1-out-domain}
\end{figure*}

The impact of domain divergence is visible in Figure~\ref{fig:mixed1-out-domain} as well. For example, in Figure~\ref{out-a}, when PMI/Gvt is used in the final stage, results for FLORES mostly increase when PMI/Gvt size increases. In contrast, in the single-domain ITTL in Figure~\ref{out-b}, results for FLORES decrease to baseline when Bible dataset size increases in the final stage.

For all four data size setups, all three techniques are better than the vanilla FT baseline, but in varying degrees. In order to further explain this observation, we calculated the correlation ($R^2$) between JSD (of the final stage dataset and test set) and the corresponding spBLEU score. For the baseline, the average $R^2$ is $0.78$, while the same is $0.57$, $0.2$0 and $0.59$ for single-domain ITTL, multi-domain ITTL and multi-domain FT (respectively). This indicates that,  compared to the vanilla FT baseline, domain divergence is less correlated to the performance when auxiliary domain data is used. 

{Overall, both multi-domain ITTL and multi-domain FT are better than single-domain ITTL, except in two cases. However, the results difference between multi-domain ITTL and multi-domain FT is marginal. In fact, in more than half of the experiments reported in Tables~\ref{tab:1k-1k-out}-\ref{tab:25k-25k-out}, multi-domain ITTL underperforms multi-domain FT.}
Our summary results also show that the variance of performance for multi-domain FT and multi-domain ITTL is $17.2$ and $28.00$ (respectively). The higher variance of multi-domain ITTL indicates that the final stage fine-tuning in the multi-domain ITTL model is more susceptible to domain divergence compared to multi-domain FT.

{In summary, all three techniques (single ITTL, multi-FT and multi-ITTL) outperform the vanilla FT baseline for all the data sizes. Among the three strategies, multi-FT and multi-ITTL outperform single ITTL in general. However, there is no clear winner between multi-FT and multi-ITTL. Considering the less computational cost involved with multi-FT and its less susceptibility to domain divergence, multi-FT would be the better choice than multi-ITTL. However, the exact results of all the strategies depend on the divergence between domains. Therefore, domains closer to the target domain should be selected to train the msLM.}
\section{Ablation Study}

\subsection{Impact of Up-sampling}
In both small-large and large-small dataset setups, one dataset size is significantly smaller than the other. In order to dilute the impact of size difference, we up-sampled\footnote{random oversampling (\url{https://t.ly/MLrhd}), we use the term up-sampling to mean oversampling} the smaller data set to match with the larger dataset. For this experiment, we used CC (the largest dataset) and Bible, and experimented with {three} languages. As shown in Table \ref{tab:oversampling-table}, we do not see any gains from upsampling.

\begin{table}[!htp]
\centering
\uniformtablesize
\setlength\tabcolsep{2pt}
\aboverulesep=0ex
\belowrulesep=0ex

\begin{tabular*}{\columnwidth}{@{\extracolsep{\fill}} lll|lll|lll|lll @{}}
\toprule \rule{0pt}{1.2EM}
\multirow{2}{*}{\textbf{Method}}  & \multirow{2}{*}{\makecell{\textbf{CC}\\\textbf{Size}}} & \multirow{2}{*}{\makecell{\textbf{Bible}\\\textbf{Size}}}  & \multicolumn{3}{c}{\textbf{\lang{hi}}} & \multicolumn{3}{c}{\textbf{\lang{ka}}} & \multicolumn{3}{c}{\textbf{\lang{si}}} \\[1pt]
  &   &   & \textbf{FLORES}  & \textbf{Bible} & \textbf{PMI}  & \textbf{FLORES}  & \textbf{Bible} & \textbf{PMI}  & \textbf{FLORES}  & \textbf{Bible}& \textbf{Gvt}  \\[2pt] \midrule \rule{0pt}{1.1EM}
\multirow{5}{*}{\makecell{Multi\\FT}}      & \multirow{2}{*}{25k}  & 1k  & \cl{-6.8}   & \textbf{0.6}  & \cl{-4.4} & \cl{-1.3}	 & \cl{-0.6}	 & \cl{-0.9} & \cl{-4.2} & \textbf{0.2} & \cl{-7.5}\\ 

    & & 10k  & \cl{-1.8}   & \textbf{0.3}  & \cl{-1.8} & \cl{-0.8} & \textbf{0.4} & \cl{-0.5} & \cl{-1.5} & \textbf{0.2} & \cl{-2.9} \\ 
    
  & \multirow{3}{*}{100k} & 1k  & \cl{-13.1}   & \textbf{1.2}    & \cl{-9.2} & \cl{-3.8} & \textbf{3.9} &	\cl{-2.5} & \cl{-8.5}	& \textbf{1.3}	& \cl{-13.6} \\ 
    &      & 10k & \cl{-7.9}   & \textbf{0.6}  & \cl{-5.2} & \cl{-0.7}	& \textbf{6.0} & 	\cl{-0.6} & \cl{-4.8} &	\textbf{1.1} &	\cl{-7.9} \\  
    &    & 25k   & \cl{-4.9} & \textbf{0.3} & \cl{-3.4} & \cl{-1.4} & \textbf{2.7} & \cl{-0.7} & \cl{-2.7} & \textbf{1.4} & \cl{-2.6} \\ \midrule \rule{0pt}{1.1EM} 
\multirow{5}{*}{\makecell{Multi\\ITTL}}    & \multirow{2}{*}{25k}  & 1k  & \cl{-7.1}   & \textbf{0.5}  & \cl{-4.7} & \cl{-1.5} &	\cl{-0.4} &	\cl{-0.9}	& \cl{-3.9} &	\cl{-0.9} &	\cl{-7.2}\\ 
  &      & 10k  & \cl{-1.6}   & \textbf{0.2}  & \cl{-1.6} & \cl{-0.4} & \textbf{0.3} & \cl{-0.1} &	\cl{-1.7} &	\cl{-0.1} &	\cl{-3.2}\\ 
 & \multirow{3}{*}{100k} & 1k  & \cl{-14}   & \textbf{0.6}  & \cl{-9.5} & \cl{-4.8} & \cl{-0.1} & \cl{-2.9} &	\cl{-9.1} &	\textbf{0.1}	& \cl{-13.8}\\
  &      & 10k & \cl{-8.1}   & \textbf{0.6}  & \cl{-5.7} & \cl{-0.6}	& \textbf{0.2} &	0.0 &  \cl{-2.0}	& \cl{-0.6} & \cl{-1.7}\\
  &      &   25k  & \textbf{2.1} & \cl{-0.7} & \textbf{1.6} & \textbf{0.2} & \textbf{1.2} &	\cl{-0.2} &	\textbf{0.5} &	\textbf{0.9}&	\textbf{0.8}\\[3pt]
  \bottomrule
\end{tabular*}
\caption{Performance difference of NMT (spBLEU) results with/without minority class upsampling. Grey textindicates where upsampling under-performs.}
\label{tab:oversampling-table}
\end{table}

\subsection{Mixing more than two domains}

All our previous experiments involved only two domains. In order to investigate the impact of combining multiple domains, we mixed data from CC, Bible, and PMI/Gvt separately for \lang{si} and \lang{ka}. As seen in Tables \ref{tab:diff_mixed_FT_stage1} and \ref{tab:diff_mixed_FT_stage2}, the inclusion of an extra domain during training does not consistently improve performance due to domain divergence. 

\begin{table}[!htp]
\centering
\uniformtablesize  
\setlength\tabcolsep{3pt}
\aboverulesep=0ex
\belowrulesep=0ex
\renewcommand{\arraystretch}{1.1}

\begin{tabular*}{\columnwidth}{@{\extracolsep{\fill}} ll|lll|lll @{}}

\toprule \rule{0pt}{1.1EM} 
\multirow{2}{*}{\textbf{Multi-Domains}} & \multirow{2}{*}{\textbf{Domain Size}} &  \multicolumn{3}{c}{\lang{ka}} & \multicolumn{3}{c}{\lang{si}} \\[1pt] 
        &  & \textbf{FLORES} & \textbf{Bible} & \textbf{PMI} & \textbf{FLORES} & \textbf{Bible} & \textbf{Gvt} \\[2pt] \midrule

CC+Bible      & 25k+25k      & 4.2   & 26.5  & 2.1 & 9.5	 & 38.0	 & 16.6 \\[2pt] 

CC+PMI/Gvt      & 25k+25k      & 16.6   & 1.1  & 33.5 & 14.1	 & 2.7	 & 43.9 \\[2pt] 

CC+PMI/Gvt+Bible      & 25k+25k+25k      & 14.5   & 25.1  & 31.9 & 13.6	 & 37.2	 & 43.5 \\ 
\bottomrule
\end{tabular*}
\caption{{NMT (spBLEU) results for Multi-domain FT with different domain combination for \lang{ka} and \lang{si}.}}
\label{tab:diff_mixed_FT_stage1}
\end{table}

\begin{table}[!htp]
\centering
\uniformtablesize
\setlength\tabcolsep{3pt}
\aboverulesep=0ex
\belowrulesep=0ex
\renewcommand{\arraystretch}{1.1}

\begin{tabular*}{\columnwidth}{@{\extracolsep{\fill}} ll|ll|lll|lll @{}}
\toprule 

\multicolumn{2}{c}{\textbf{Intermediate Task}} & \multicolumn{2}{c}{\makecell{\textbf{ Final Stage}\\}} & \multicolumn{3}{c}{\textbf{\lang{ka}}} & \multicolumn{3}{c}{\textbf{\lang{si}}} \\
\textbf{Multi-Domains} & \textbf{Domain Size} & \textbf{Domain} & \textbf{Size} & \textbf{FLORES} & \textbf{Bible} & \textbf{PMI} & \textbf{FLORES} & \textbf{Bible} & \textbf{Gvt} \\[2pt] \midrule \rule{0pt}{1.1EM} 

CC+Bible & 25k+25k & \multirow{2}{*}{Bible} & \multirow{2}{*}{25k} & \textbf{3.1} &\textbf{27.4} & \cl{1.4} & \textbf{8.8} &\textbf{37.9} & \cl{15.6} \\ 

CC+PMI/Gvt+Bible & 25k+25k+25k & & & \textbf{7.0} &\textbf{27.8} & \cl{14.4} & \textbf{9.1} &\textbf{37.6} & \cl{34.3} \\[2pt]
\multicolumn{2}{c}{\textbf{\hspace{15mm} Improvement}} & & & \textbf{3.9} &\textbf{0.4} & \cl{13.0} & \textbf{0.3} &\textbf{-0.3} & \cl{18.7} \\[2pt] \midrule \rule{0pt}{1.0EM} 

CC+PMI/Gvt & 25k+25k & \multirow{2}{*}{PMI/Gvt} & \multirow{2}{*}{25k} & \textbf{16.2} &\cl{1.8} & \textbf{33.7} & \textbf{13.3} &\cl{2.3} & \textbf{44.7} \\[2pt]

CC+PMI/Gvt+Bible & 25k+25k+25k & & & \textbf{16.6} &\cl{19.8} & \textbf{34.3} & \textbf{13.8} &\cl{31.0} & \textbf{45.0} \\[2pt]
\multicolumn{2}{c}{\textbf{\hspace{15mm} Improvement}} & & & \textbf{0.4} &\cl{18.0 } & \textbf{0.6 } & \textbf{0.5} &\cl{28.7} & \textbf{0.3 } \\[2pt]
\end{tabular*}
\caption{NMT(spBLEU) results for Multi-domain ITTL with different domain combinations for \lang{ka} and \lang{si}. Last row shows the improvement of using three domains.}
\label{tab:diff_mixed_FT_stage2}
\end{table}

In multi-domain FT (Table \ref{tab:diff_mixed_FT_stage1}), when testing on FLORES, combining Bible with CC+PMI/Gvt yielded inferior results. A similar observation holds for the in-domain test case---adding Bible to CC+PMI/Gvt drops the result on PMI/Gvt.

In multi-domain ITTL (Table \ref{tab:diff_mixed_FT_stage2}), when testing on FLORES, adding PMI/Gvt to CC+Bible significantly improves the results, but adding Bible to CC+PMI/Gvt yields  marginal gains. Similarly, in the in-domain case, adding PMI/Gvt to CC+Bible does not increase the result on Bible, and adding Bible to CC+PMI/Gvt does not increase the result on PMI/Gvt.

\begin{table*}[!htpb]
\centering
\uniformtablesize
\setlength{\datgap}{20pt}
\setlength\tabcolsep{3pt}
\aboverulesep=0ex
\belowrulesep=0ex
\renewcommand{\arraystretch}{1.1}

\begin{tabular*}{\textwidth}{@{\extracolsep{\fill}} ccccc@{\hspace{\datgap}}cccc @{}}
\toprule
\multirow{2}{*}{\textbf{Language}} & \multicolumn{4}{c}{\makecell{\textbf{Intermediate task - PMI/Gvt} \\ \textbf{Final task - Bible}}} & \multicolumn{4}{c}{\makecell{\textbf{Intermediate task - Bible} \\ \textbf{Final task - PMI/Gvt}}} \\
\cmidrule(r{\datgap}){2-5} \cmidrule(r{\datgap}){6-8} \cmidrule(l){8-9}
 & \makecell{\textbf{Single} \\ \textbf{ITTL}} & \makecell{\textbf{Multi} \\ \textbf{FT}} & \makecell{\textbf{Multi} \\ \textbf{ITTL}} & \textbf{Baseline} & \makecell{\textbf{Single} \\ \textbf{ITTL}} & \makecell{\textbf{Multi} \\ \textbf{FT}} & \makecell{\textbf{Multi} \\ \textbf{ITTL}} & \textbf{Baseline} \\ \midrule
\lang{si} & \cl{12.00} & \textbf{12.25} & 12.04 & \textit{12.08} & \textit{24.18} & \cl{23.99} & \textbf{24.19} & 24.15 \\ 
\lang{ta} & 8.65 & \textbf{8.87} & \cl{8.44} & \textit{8.75} & \cl{14.55} & \textit{14.91} & 14.75 & \textbf{14.96} \\
\lang{gu} & \cl{10.84} & \textbf{11.34} & \textit{11.09} & \textit{11.09} & \cl{16.91} & \textbf{18.08} & 16.94 & \textit{17.36} \\
\lang{hi} & \textit{18.53} & \textbf{19.03} & \cl{18.41} & 18.49 & 24.49 & \textbf{25.49} & \cl{24.15} & \textit{24.99} \\
\lang{ka} & \cl{6.27} & \textbf{6.80} & \textit{6.47} & 6.42 & \cl{14.85} & \textit{15.07} & 14.91 & \textbf{15.45} \\
\hline
\textit{Avg} & \cl{11.26} & \textbf{11.66} & 11.29 & \textit{11.37} & 19.00 & \textbf{19.51} & \cl{18.99} & \textit{19.38} \\ \hline
\end{tabular*}
\captionsetup[figure]{font=large}
\caption{NMT (spBLEU) results for the in-domain setup using the \textit{NLLB-200-distilled-600M} model. Intermediate task dataset size -1k, final task dataset size-1k.}
\label{tab:1k-1k-in-nllb}
\end{table*}

\begin{table*}[!htpb]
\centering
\uniformtablesize
\setlength{\datgap}{20pt}
\setlength\tabcolsep{3pt}
\aboverulesep=0ex
\belowrulesep=0ex
\renewcommand{\arraystretch}{1.1}

\begin{tabular*}{\textwidth}{@{\extracolsep{\fill}} ccccc@{\hspace{\datgap}}cccc @{}}
\toprule
\multirow{2}{*}{\textbf{Language}} & \multicolumn{4}{c}{\makecell{\textbf{Intermediate task - PMI/Gvt} \\ \textbf{Final task - Bible}}} & \multicolumn{4}{c}{\makecell{\textbf{Intermediate task - Bible} \\ \textbf{Final task - PMI/Gvt}}} \\
\cmidrule(r{\datgap}){2-5} \cmidrule(r{\datgap}){6-8}  \cmidrule(l){8-9}
     & \makecell{\textbf{Single} \\ \textbf{ITTL}}      & \makecell{\textbf{Multi} \\ \textbf{FT}}      & \makecell{\textbf{Multi} \\ \textbf{ITTL}}     & \textbf{Baseline}     & \makecell{\textbf{Single} \\ \textbf{ITTL}}      & \makecell{\textbf{Multi} \\ \textbf{FT}}      & \makecell{\textbf{Multi} \\ \textbf{ITTL}}     & \textbf{Baseline}    \\ \midrule
\lang{si}   & 5.88     & \textbf{8.63}  & \textit{7.68}   & \cl{5.26}    & \cl{7.13}     & \textbf{8.63}  & 7.67   & \textit{7.94}    \\
\lang{ta}   & \cl{5.41}     & \textbf{8.47}  & \textit{7.00}   & 6.49    & \cl{7.13}     & \textbf{8.47}  & 7.70   & \textit{7.78}   \\
\lang{gu}   & \cl{13.12}     & \textbf{15.24}  & \textit{14.86}   & 13.38    & \cl{13.74}     & \textbf{15.24}  & 14.40   & \textit{15.11}    \\
\lang{hi}   & \cl{17.02}     & \textbf{21.55}  & \textit{21.08}   & 18.86    & \cl{19.62}     & \textbf{21.55}  & 20.12   & \textit{21.38}    \\
\lang{ka}   & \cl{6.91}     & \textbf{11.10}  & \textit{9.31}   & 7.05    & \cl{9.17}     & \textbf{11.10}  & 9.64   & \textit{10.44}    \\
\midrule
\textit{Avg}  & \cl{9.67}     & \textbf{13.00}    & \textit{11.99}   & 10.21    & \cl{11.36}    & \textbf{13.00}  & 11.91   & \textit{12.53}    \\ \bottomrule
\end{tabular*}
\caption{NMT (spBLEU) results for the out-domain setup using the \textit{NLLB-200-distilled-600M} model. Intermediate task dataset size-1k, final task dataset size-1k.}
\label{tab:1k-1k-out-nllb}
\end{table*}

\subsection{Impact of the msLM}
{In order to determine whether our observations hold across msLMs, we conducted an ablation study with the NLLB-200-distilled-600M model for both the in-domain and the out-domain cases, considering the small-small data setup. Results are reported in Tables ~\ref{tab:1k-1k-in-nllb} and ~\ref{tab:1k-1k-out-nllb}. Here, Table ~\ref{tab:1k-1k-in-nllb} has the same setup as Table~\ref{tab:1k-1k-in} for the in-domain case, and Table ~\ref{tab:1k-1k-out-nllb} has the same setup as Table~\ref{tab:1k-1k-out} for the out-domain case.
In this data setup, reaffirming the recommendation we made for mBART, multiFT emerges as the most promising technique for the in-domain case, despite there being a lesser results variation among the techniques compared to that of mBART. As for the out-domain case, the observations made with the mBART models are mostly echoed with NLLB as well. To begin with, vanilla FT baseline performs better when PMI/Gvt is used for fine-tuning, compared to using Bible. Similarly,  the best results from Single ITTL are reported when a domain closer to FLORES is used in the final fine-tuning stage. However, for Multi ITTL, there is no result variation with respect to the domain used in the final stage. Overall, both multi-domain ITTL and multi-domain FT are better than single-domain ITTL. However, some individual ITTL results fall below the baseline. Here also we can conclude that multiFT is better than multiITTL.}

\section{Discussion}
\subsection{Recommendations}
{Based on our extensive experiments, the following recommendations can be made:}
\begin{itemize}
    \item {When the parallel data set size is small (less than 25k), do not pre-train mBART with bitext denoising, as it may not result in any gains.}
    \item {In the in-domain case, when two domains are concerned, and (1) if both have small amounts of data (about 1k), use the multi-domain FT option, (2) use the multi-domain ITTL option when the intermediate dataset size increases, (3) Simply use vanilla FT if the target domain has at least 25k data samples.}
    \item {In the out-domain test case, when two domains are concerned, results are heavily impacted by the divergence of domains. Therefore, before carrying any experiments, calculate the divergence between domains and use data from a domain that is closest to the target domain. We recommend multi-FT in this case, since its results are less susceptible to domain divergence and it is computationally more efficient. }
    \item {Do not blindly mix data belonging to more than two domains, as the effective gains are highly subjective to the domain divergence.}
\end{itemize}

\subsection{Limitations}
{In our experiments, we considered only mBART, with an ablation study on NLLB. Other msLMs could not be considered due to the limitations in computational resources. Similarly, the number of languages had to be kept at 5, due to the difficulty in finding data belonging to multiple domains. Due to this, it was not possible to carry out a language-wise analysis. The maximum dataset size for most domains/languages was 25k, a common problem with LRLs. To keep the results discussion concise, only a selected set of EN-XX experiment results were used for the discussion. For clarity, only spBLEU results are reported. We could not report results on human evaluations, due to the lack of funds to employ professional translators.}

\section{Conclusion}

The optimal way of using auxiliary domain parallel data to build domain-specific LRL-NMT systems on msLMs {was} an under-explored problem. We presented the first comparative study of two ways to utilize this parallel data---fine-tuning and additional pre-training. {Our key findings are: (1) with small parallel data ($\le$ 25k per domain in our experiments), continuous pre-training with bitext denoising does not provide visible gains; (2) for in-domain translation, multi-domain FT is better for small amounts of data and multi-domain ITTL is the best when the dataset size increases, however when the target domain has $\approx$ 25k parallel sentences, vanilla FT is usually sufficient; (3) for out-of-domain evaluation, auxiliary-domain strategies generally improve over vanilla FT, but the best choice depends on the divergence between auxiliary and target domains; (4) multi-domain FT offers a strong performance/compute trade-off and is generally less sensitive to domain divergence than multi-domain ITTL; and (5) blindly mixing data from multiple domains do not deliver expected results due to domain divergence}. We hope our findings and recommendation will assist NMT researchers in selecting the optimal way to use data from auxiliary domains, without having to run extensive sets of experiments. In the future, we plan to experiment with multilingual FT and ITTL, as well as multilingual and multi-domain FT and ITTL. It would also be worthwhile to explore new pre-training objectives on parallel data as well.


\bibliographystyle{ACM-Reference-Format}
\bibliography{sample-base}
\newpage
\appendix
\section{Experiment Setup}
\label{sec:appendixA}

\subsection{Corpus Details (from~\cite{nayak2023leveraging})}

\textbf{PMIndia corpus (PMI)}  ~\cite{haddow2020pmindia} is a multi-way parallel corpus of 14 languages (13 Indian languages, plus English). The data set contains Indian Prime Minister's speeches extracted from the Prime Minister of India website\footnote{\url{https://www.pmindia.gov.in/en/}}.

\textbf{Government corpus (Gvt)} ~\cite{fernando2020data} is a multi-way parallel corpus for Sinhala, Tamil, and English. This is a manually curated data set. It consists of data from official government documents (annual reports, committee reports, government institutional websites, and acts of the Parliament).

\textbf{Bible corpus} We scrape Bible data from web\footnote{Sinhala: https://www.wordproject.org/bibles/si/index.htm; and others: https://ebible.org/download.php} and then automatically align the sentences (at verse level) using the scripts provided by Nayak et al.~\cite{nayak2023leveraging} This curated multi-way parallel corpus has 25k parallel sentences in \lang{kn}, \lang{gu}, \lang{hi}, \lang{ta}. Note that Sinhala was scraped from a different website, thus has different content\footnote{ We will be releasing the scripts to create the corpus on acceptance of the paper.}.

\textbf{CCAligned (CC)}  is an automatically sentence-aligned (bi-text mined) corpus based on the CommonCrawl. 

\textbf{FLORES-101} ~\cite{goyal-etal-2022-flores}  is a multi-way parallel corpus. The source is English Wikipedia, which has been manually translated into 101 languages. Data comes from a variety of topics and domains, thus can be considered open-domain. We use \textsc{Flores}v1~\cite{guzman2019flores} for Sinhala since it is not present in \textsc{Flores}-101.

{When the size of a training dataset exceeded the amount of samples needed for the experiment, we randomly sampled data from the source corpus.}

\subsection{Model Training}

{The mBART experiments were implemented using the Fairseq library and conducted on an NVIDIA Volta GPU with 32 GB of memory. \\
The NLLB experiments were implemented using the Hugging Face Transformers library and conducted on an NVIDIA H100 GPU with 80 GB of memory. \\
All experiments were conducted with a fixed random seed of 222.}

Training details are as follows: number of epochs =$3$, learning rate = $3 \cdot 10^{-5}$, dropout = $0.3$, attention dropout = $0.1$, and a batch size = $32$ for training and evaluation. {mBART tokenizer was used to tokenize text.}

\subsection{Evaluation Metric}

spBLEU calculates the BLEU score for the text tokenized using the sentence-piece sub-word model, which has been trained for all the 101 languages in \textsc{Flores}-101 data set (including the languages (except \lang{kn}) considered in our experiments). Furthermore, Goyal et al.~\cite{goyal-etal-2022-flores} showed that spBLEU functions similarly to BLEU and has a strong correlation with the Chrf++ metric.~\cite{popovic-2017-chrf} \footnote{We use the official implementation provided in the sacreBLEU library\footnote{https://github.com/mjpost/sacreBLEU} ~\cite{post-2018-call}}. 
The sacreBLEU signature for pre-training with bi-text denoising is: 
nrefs:1|case:mixed|eff:no|tok:\\spm-flores|smooth:none|version:2.1.0. 
The sacreBLEU signature for Single-domain ITTL is: 
nrefs:\\1|case:mixed|eff:no|tok:spm-flores|smooth:none|\\version:2.2.1. 
\\
The signature for Multi-domain ITTL is: 
nrefs:1|case:mixed|eff:no|tok:flores101|\\smooth:none|version:2.3.1.

\subsection{Jenson-Shannon Divergence}

Jenson-Shannon Divergence (JSD) is calculated between two distributions $P$ and $Q$ using the formula given below: 
$$JSD(P||Q) = \frac{1}{2}KL(P||M) + \frac{1}{2}KL(Q||M)$$

Here, $M$ represents an equally weighted sum of the two distributions (i.e., $M = \frac{1}{2}P + \frac{1}{2}Q$). $KL(\cdot || \cdot)$ represents the Kullback–Leibler divergence. JSD ranges from 0 to 1 - lower the value, more similar the distributions are.

Divergence is calculated between each pair of train and test sets, for a given language pair. These language-specific values are then averaged to obtain a similarity score independent of the target language.

Following our previous work~\cite{nayak2023leveraging}, each corpus was first tokenized (times and numbers were also converted into the tokens \verb|<TIME>| and \verb|<NUMBER>|, respectively) using the \verb|NLTK| package,\footnote{https://www.nltk.org/} and stopwords were removed. The resulting corpora were transformed into a (discrete) frequency distribution over all word tokens. The frequency distributions of each train and test set are then compared using the formula above.

\section{Pre-training results}%
\label{sec:pretraining}


\begin{table*}[!htb] 
\centering
\uniformtablesize
\setlength{\datgap}{10pt}
\setlength\tabcolsep{3pt}
\renewcommand{\arraystretch}{1.1}

\begin{tabular*}{\textwidth}{@{\extracolsep{\fill}} %
l l l l l @{\hspace{\datgap}} lll @{\hspace{5pt}} lll @{\hspace{5pt}} lll @{}}
\toprule

\multirow{2}{*}{\textbf{Pre-training Set}}  & \multirow{2}{*}{\textbf{Pre-training}} & \multirow{2}{*}{\textbf{Pre-training Set Size}} & \multirow{2}{*}{\textbf{Fine-tuning Set}}
& \multirow{2}{*}{\textbf{Fine-tuning Set Size}} & \multicolumn{3}{c}{\textbf{\lang{ka}}} & \multicolumn{3}{c}{\textbf{\lang{si}}} & \multicolumn{3}{c}{\textbf{\lang{hi}}} \\
\cmidrule(lr){6-8} \cmidrule(lr){9-11} \cmidrule(lr){12-14} 
 & &   & &  & \textbf{FLORES}  & \textbf{Bible} & \textbf{PMI} & \textbf{FLORES}  & \textbf{Bible} & \textbf{Gvt}& \textbf{FLORES}& \textbf{Bible} & \textbf{PMI} \\ 
 \hline
\multirow{4}{*}{CC} & \multirow{2}{*}{bitext}  & 25k  &   \multirow{4}{*}{CC}  & 25k &   1.0&	0.0	&0.6&8.5&	1.9&	14.1	&12.3	&5.0&	11.3   \\
&  & 100k &  & 100k  & 6.2&	0.1	&4.5	&15.4	&4.1	&26.4	&22.7	&6.8&	17.8 \\
 & \multirow{2}{*}{bitext + monolignual}  & 25k  & & 25k & 0.5&	0.0&	0.4&	8.4&	1.8&	13.3	&14.0&	5.4&	11.9 \\ \vspace{.7em}
&  & 100k &  & 100k   & 5.8&	0.2	&4.1	&14.6&	4.1&	25.9	&22.9	&6.9&	18.4\\

\multirow{8}{*}{PMI/Gvt} & \multirow{2}{*}{bitext}  & 1k  &   \multirow{8}{*}{PMI/Gvt}  & 1k   &0.5	&0.1&	3.7	&3.7	&0.5	&19.5	&6.5	&1.4   &17.3\\
&  & 10k &   & 10k  &9.6	&1.1&	28.9	&8.8&	0.9&	39.7&	14	&2.2 & 31.7\\
&  & 25k &   & 25k & 14.3	&1.9	&34.3	&10.8	&1.1	&44.9&	17.2&	3 &35.6\\
&  & 50k &   & 50k  &N/A & N/A&N/A & 12.2	&1.3	&48.7&	19.1&	3.3 &37.2\\
 & \multirow{2}{*}{bitext + monolingual}  & 1k  & & 1k  &0.5	&0.1	&3.3	&3.2	&0.4	&21.2	&7.1	&1.7	&18.6 \\
&  & 10k &   & 10k  &9.1&	0.9	&29.1	&8.3&	0.8&	38.3&	14	&2.4&	31.8\\
&  & 25k &   & 25k & 12.8&	1.6&	33.8&	10.7	&1.1&	44.5	&17.2	&3	&35.6\\ \vspace{.7em}
&  & 50k &   & 50k &N/A & N/A&N/A  &11.7	&1.3&	49.2	&18.7&	3.3&	36.8 \\

\multirow{6}{*}{Bible} & \multirow{3}{*}{bitext}  & 1k  &   \multirow{6}{*}{Bible}  & 1k &0.3	&6.7	&0.2	&0.7	&12.6	&1.1	&2.4&	15.9	&2.1  \\
&  & 10k &   & 10k & 1.5	&21.9&	0.6&	1.8	&33.7	&0.8&	2.9	&26.9&	2.2 \\
&  & 25k &   & 25k   &2.1&28&	0.8	&1.6&	38&	0.7	&3	&31.6&	2.1 \\
 & \multirow{3}{*}{bitext + monolignual}  & 1k  & & 1k  &0.3&	7.4	&0.2&	0.8&	17.9&1.2	&0.9&	10.5	&1.2\\
&  & 10k &   & 10k  & 1.3&	22.4	&0.6&	1.8&	33.9	&0.8&	2.9	&26.8&	2 \\
&  & 25k &   & 25k  & 1.8&	28	&0.7&	1.7	&37.9	&0.7&	2.7	&31.4&1.9 \\\hline
\end{tabular*}
\caption{Results (spBLEU) for pre-training with bi-text denoising.}
\label{tab:paradise-full}
\end{table*}

\clearpage
\onecolumn

\section{Tabular Results for Fine-tuning Techniques}%
\label{sec:resulttables}
\begin{table}[!htb]
\centering
\uniformtablesize
\setlength\tabcolsep{2.5pt}
\renewcommand{\arraystretch}{1.1}

\begin{tabular*}{\textwidth}{@{\extracolsep{\fill}} l l ccc ccc @{}}
\toprule
\multirow{2}{*}{\textbf{Fine-tuning Set}} 
& \multirow{2}{*}{\textbf{Fine-tuning Set Size}} 
& \multicolumn{3}{c}{\textbf{Telugu ID}} 
& \multicolumn{3}{c}{\textbf{Tamil ID}} \\
\cmidrule(lr){3-5} \cmidrule(lr){6-8}
& & \textbf{FLORES} & \textbf{Bible} & \textbf{PMO} 
  & \textbf{FLORES} & \textbf{Bible} & \textbf{PMO} \\
\midrule

\multirow{3}{*}{PMO}
 & 1k  & 1.0 & 0.1 & 6.5 & 1.0 & 0.0 & 6.3 \\
 & 25k & 14.1 & 1.7 & 33.6 & 14.1 & 1.8 & 33.8 \\

\multirow{3}{*}{Bible}
 & 1k  & 0.3 & 8.2 & 0.2 & 0.2 & 7.8 & 0.2 \\
 & 25k & 2.2 & 27.8 & 0.9 & 2.1 & 27.8 & 0.9 \\

\bottomrule
\end{tabular*}

\caption{Results (spBLEU) for vanilla fine-tuning on Kannada using Telugu and Tamil language IDs.}
\label{tab:fine-tuning-kannada}
\end{table}

\subsection{{Result Tables for EN-XX}}

\begin{table*}[!htb]
\centering
\uniformtablesize
\setlength{\datgap}{10pt}
\setlength\tabcolsep{1.5pt}
\renewcommand{\arraystretch}{1.1}

\begin{tabular*}{\textwidth}{@{\extracolsep{\fill}} %
l l l l
@{\hspace{\datgap}} lll
@{\hspace{\datgap}} lll
@{\hspace{\datgap}} lll
@{\hspace{\datgap}} lll
@{\hspace{\datgap}} lll @{}}
\toprule
\multirow{2}{*}{\textbf{D1}}  & \multirow{2}{*}{\textbf{D1 size}} & \multirow{2}{*}{\textbf{D2}}  & \multirow{2}{*}{\textbf{D2 size}} & \multicolumn{3}{c}{\textbf{\lang{ka}}} & \multicolumn{3}{c}{\textbf{\lang{gu}}} & \multicolumn{3}{c}{\textbf{\lang{hi}}} & \multicolumn{3}{c}{\textbf{\lang{si}}} & \multicolumn{3}{c}{\textbf{\lang{ta}}} \\
\cmidrule(r{\datgap}){5-7} \cmidrule(r{\datgap}){8-10} \cmidrule(r{\datgap}){11-13} \cmidrule(r{\datgap}){14-16} \cmidrule(){17-19}
  &   &   &    & \textbf{FLORES}  & \textbf{Bible} & \textbf{PMI}  & \textbf{FLORES}  & \textbf{Bible} & \textbf{PMI}  & \textbf{FLORES}  & \textbf{Bible} & \textbf{PMI}  & \textbf{FLORES}  & \textbf{Bible} & \textbf{Gvt}  & \textbf{FLORES}  & \textbf{Bible} & \textbf{Gvt}  \\ \midrule
\multirow{8}{*}{CC} & \multirow{4}{*}{25k}   & \multirow{8}{*}{PMI/Gvt}  & 1k    & 4.5  & 0.4   & 12.9 & 16.8    & 3.6   & 23.8 & 13.3    & 3.6   & 20   & 8.9  & 1.7   & 24.1 & 10.1    & 5.9   & 20.0 \\
&   &   & 10k   & 12.4    & 1.6   & 30.5 & 19.4    & 4.6   & 34.7 & 16.6    & 3.3   & 31.8 & 10.6    & 1.9   & 39.2 & 9.8  & 3.6   & 33.4 \\
&   &   & 25k   & 15.0    & 1.9   & 34.3 & 21.5    & 5.1   & 38.3 & 18.3    & 3.3   & 35.5 & 12.1    & 1.9   & 44.4 & 11.1    & 3.3   & 37.2 \\
&   &   & 50k   & N/A  & N/A   & N/A  & N/A  & N/A   & N/A  & 19.9    & 3.8   & 37.0 & 12.8    & 2.0   & 49.7 & 12.1    & 2.8   & 40.1 \\
 & \multirow{4}{*}{100k}  &   & 1k    & 10.2    & 1.1   & 16.2 & 20.9    & 3.7   & 26.0 & 22.1    & 6.1   & 24.0 & 16.3    & 4.6   & 31.0 & 20.3    & 9.4   & 23.9 \\
&   &   & 10k   & 14.6    & 1.8   & 30.3 & 22.5    & 5.2   & 35.1 & 20.3    & 4.9   & 32.5 & 15.2    & 4.1   & 41.4 & 15.4    & 6.2   & 34.1 \\
&   &   & 25k   & 16.3    & 2.1   & 34.1 & 22.9    & 5.4   & 38.5 & 21.4    & 4.4   & 35.2 & 15.2    & 3.6   & 44.3 & 15.0    & 5.0   & 37.6 \\ \vspace{0.7em}
 &   &   & 50k   & N/A  & N/A   & N/A  & N/A  & N/A   & N/A  & 22.5    & 4.4   & 37.0 & 15.2    & 3.5   & 48.7 & 14.9    & 4.5   & 40.5 \\ 
\multirow{6}{*}{CC}   & \multirow{3}{*}{25k}   & \multirow{6}{*}{Bible}    & 1k    & 0.9  & 9.6   & 0.7  & 9.6  & 13.9  & 8.5  & 7.1  & 16.1  & 6.8  & 3.5  & 19.3  & 6.2  & 7.2  & 14.5  & 6.7  \\
  &   &   & 10k   & 2.1  & 23.1  & 0.9  & 6.6  & 23.7  & 5.1  & 4.0  & 26.4  & 3.0  & 2.2  & 33.5  & 2.2  & 3.0  & 26.1  & 2.5  \\
  &   &   & 25k   & 2.4  & 27.3  & 1    & 5.6  & 26.9  & 4.1  & 3.8  & 31.5  & 2.4  & 2.0  & 37.9  & 1.0  & 2.9  & 30.8  & 1.6  \\
  & \multirow{3}{*}{100k}  &   & 1k    & 4.1  & 11.2  & 2.3  & 14.6    & 13.3  & 11.9 & 16.9    & 16.5  & 14.5 & 11.0    & 20.1  & 19.0 & 17.2    & 16.3  & 15.3 \\
  &   &   & 10k   & 2.8  & 22.9  & 1.4  & 8.3  & 24.2  & 6.9  & 6.6  & 26.8  & 5.5  & 3.9  & 33.4  & 5.4  & 5.9  & 26.2  & 6.0  \\ \vspace{0.7em}
  &   &   & 25k   & 2.8  & 27.4  & 1.2  & 6.9  & 27.8  & 5.4  & 4.8  & 31.7  & 3.3  & 3.0  & 37.9  & 2.6  & 4.2  & 31.0  & 3.2  \\
\multirow{12}{*}{PMI/Gvt}  & \multirow{3}{*}{1k}    & \multirow{12}{*}{Bible} & 1k    & 0.5  & 9.4   & 1.0  & 6.0  & 13.0  & 15.6 & 4.8  & 15.5  & 11.4 & 2.4  & 17.2  & 13.1 & 2.2  & 11.0  & 13.7 \\
&   & \multicolumn{1}{c}{}   & 10k   & 1.7  & 22.0  & 0.9  & 5.1  & 24.2  & 6.6  & 3.3  & 27.2  & 2.9  & 1.9  & 34.1  & 3.2  & 2.3  & 26.3  & 3.7  \\
&   & \multicolumn{1}{c}{}   & 25k   & 2.3  & 27.0  & 1.1  & 5.2  & 27.3  & 5.1  & 3.4  & 31.5  & 2.5  & 1.9  & 37.6  & 1.8  & 2.2  & 30.5  & 2.7  \\
& \multirow{3}{*}{10k}   & \multicolumn{1}{c}{}   & 1k    & 4.2  & 12.4  & 12.9 & 11.8    & 14.7  & 26.3 & 10.2    & 15.8  & 24.7 & 5.3  & 17.5  & 29.8 & 3.8  & 13.6  & 23.3 \\
&   & \multicolumn{1}{c}{}   & 10k   & 2.8  & 23.1  & 2.9  & 7.4  & 24.2  & 13.6 & 5.1  & 26.9  & 8.9  & 2.2  & 32.4  & 8.0  & 2.6  & 25.8  & 8.1  \\
&   & \multicolumn{1}{c}{}   & 25k   & 2.9  & 27.1  & 2.1  & 6.5  & 27.7  & 9.1  & 4.2  & 31.5  & 4.6  & 2.1  & 37.8  & 4.4  & 2.5  & 30.6  & 4.8  \\
& \multirow{3}{*}{25k}   & \multicolumn{1}{c}{}   & 1k    & 7.6  & 12.7  & 21.3 & 15.3    & 14.3  & 31.8 & 12.8    & 15.3  & 29.3 & 7.9  & 18.9  & 36.7 & 5.4  & 13.4  & 28.4 \\
&   & \multicolumn{1}{c}{}   & 10k   & 3.7  & 23.1  & 6.1  & 9.7  & 23.9  & 20.1 & 6.0  & 27.0  & 12.8 & 3.3  & 33.1  & 14.5 & 3.2  & 24.8  & 13.8 \\
&   & \multicolumn{1}{c}{}   & 25k   & 3.2  & 27.5  & 2.8  & 8.0  & 27.4  & 13.4 & 4.3  & 31.6  & 6.4  & 2.4  & 38.1  & 6.5  & 2.8  & 30.9  & 7.3  \\
& \multirow{3}{*}{50k}   & \multicolumn{1}{c}{}   & 1k    & N/A  & N/A   & N/A  & N/A  & N/A   & N/A  & 14.7    & 16.0  & 31.3 & 8.4  & 19.2  & 40.5 & 6.9  & 13.7  & 32.5 \\
&   & \multicolumn{1}{c}{}   & 10k   & N/A  & N/A   & N/A  & N/A  & N/A   & N/A  & 6.7  & 26.7  & 15.8 & 4.3  & 33.5  & 17.8 & 3.8  & 25.0  & 17.3 \\ \vspace{0.7em}
&   & \multicolumn{1}{c}{}   & 25k   & N/A  & N/A   & N/A  & N/A  & N/A   & N/A  & 4.8  & 31.8  & 8.2  & 3.0  & 37.3  & 8.6  & 3.3  & 30.5  & 9.8  \\
\multirow{12}{*}{Bible}   & \multirow{4}{*}{1k}    & \multirow{12}{*}{PMI/Gvt}    & 1k    & 1.0  & 1.6   & 5.9  & 8.3  & 9.6   & 18.8 & 8.4  & 8.6   & 18.3 & 4.9  & 10.3  & 20.9 & 3.9  & 6.1   & 19.9 \\
  &   &   & 10k   & 10.0    & 1.7   & 28.5 & 17.3    & 7.5   & 34.5 & 14.7    & 5.3   & 31.4 & 9.3  & 5.5   & 38.6 & 7.0  & 4.6   & 32.7 \\
  &   &   & 25k   & 14.0    & 2.0   & 33.8 & 20.2    & 6.8   & 38.0 & 16.9    & 4.7   & 35.1 & 11.3    & 3.5   & 45.0 & 9.5  & 3.0   & 37.1 \\
  &   &   & 50k   & N/A  & N/A   & N/A  & N/A  & N/A   & N/A  & 19.1    & 4.9   & 37.2 & 12.3    & 3.5   & 49.5 & 11.2    & 2.6   & 40.7 \\
  & \multirow{4}{*}{10k}   &   & 1k    & 4.0  & 12.0  & 12.8 & 10.3    & 19.4  & 20.2 & 9.1  & 20.1  & 19.2 & 6.1  & 25.1  & 22.4 & 4.8  & 19.3  & 18.9 \\
  &   &   & 10k   & 11.4    & 6.0   & 29.7 & 18.1    & 13.9  & 34.8 & 15.4    & 14.2  & 31.4 & 10.0    & 15.1  & 39.3 & 7.9  & 12.2  & 33.2 \\
  &   &   & 25k   & 14.3    & 4.3   & 34.2 & 20.5    & 12.0  & 37.9 & 17.9    & 10.8  & 34.7 & 11.8    & 12.2  & 44.2 & 9.9  & 10.2  & 37.3 \\
  &   &   & 50k   & N/A  & N/A   & N/A  & N/A  & N/A   & N/A  & 19.9    & 10.1  & 36.7 & 12.7    & 9.3   & 49.0 & 11.6    & 8.5   & 39.8 \\
  & \multirow{4}{*}{25k}   &   & 1k    & 4.5  & 16.3  & 12.9 & 10.5    & 22.6  & 19.9 & 8.8  & 23.9  & 18.7 & 5.8  & 28.7  & 22.1 & 4.9  & 24.3  & 18.3 \\
  &   &   & 10k   & 11.6    & 8.4   & 29.8 & 17.8    & 17.4  & 33.7 & 15.4    & 17.2  & 31.2 & 10.0    & 19.6  & 39.1 & 8.0  & 14.9  & 33.6 \\
  &   &   & 25k   & 14.2    & 6.5   & 32.8 & 20.5    & 14.7  & 37.6 & 18.4    & 15.2  & 34.4 & 11.9    & 14.9  & 44.2 & 10.0    & 13.2  & 37.2 \\
  &   &   & 50k   & N/A  & N/A   & N/A  & N/A  & N/A   & N/A  & 20   & 12.7  & 36.8 & 12.9    & 12.7  & 48.6 & 11.2    & 11.4  & 39.9 \\ 
  \bottomrule
\end{tabular*}
\caption{Results (spBLEU) for single-domain ITTL. Here D1 refers to Domain 1 and D2 refers to Domain 2.}
\label{tab:single-domainITFT}
\end{table*}


\begin{table*}[!htb]
\centering
\uniformtablesize
\setlength{\datgap}{10pt}
\setlength\tabcolsep{1.5pt}
\renewcommand{\arraystretch}{1.1}

\begin{tabular*}{\textwidth}{@{\extracolsep{\fill}} %
l l
@{\hspace{\datgap}} lll
@{\hspace{\datgap}} lll
@{\hspace{\datgap}} lll
@{\hspace{\datgap}} lll
@{\hspace{\datgap}} lll @{}}
\toprule
\multirow{2}{*}{\textbf{D1 + D2}}  & \multirow{2}{*}{\textbf{D1 + D2 size}} & \multicolumn{3}{c}{\textbf{\lang{ka}}} & \multicolumn{3}{c}{\textbf{\lang{gu}}} & \multicolumn{3}{c}{\textbf{\lang{hi}}} & \multicolumn{3}{c}{\textbf{\lang{si}}} & \multicolumn{3}{c}{\textbf{\lang{ta}}} \\
\cmidrule(r{\datgap}){3-5} \cmidrule(r{\datgap}){6-8} \cmidrule(r{\datgap}){9-11} \cmidrule(r{\datgap}){12-14} \cmidrule(){15-17}
    &    & \textbf{FLORES}  & \textbf{Bible}   & \textbf{PMI}  & \textbf{FLORES}  & \textbf{Bible} & \textbf{PMI}  & \textbf{FLORES}  & \textbf{Bible} & \textbf{PMI}  & \textbf{FLORES}  & \textbf{Bible} & \textbf{Gvt}  & \textbf{FLORES}  & \textbf{Bible} & \textbf{Gvt}  \\ \midrule
\multirow{8}{*}{CC + PMI/Gvt}   & 25k + 1k    & 4.7  & 0.0   & 13.8    & 16.1    & 2.1   & 24.3 & 16.3    & 5.6   & 20.8 & 9.7  & 1.9   & 23.7 & 12.2    & 6.6   & 20.0 \\
     & 25k + 10k   & 13.6    & 0.4   & 30.0    & 22.2    & 3.4   & 34.4 & 19.4    & 5.9   & 31.2 & 12.2    & 2.4   & 37.9 & 13.1    & 7.0   & 32.7 \\
     & 25k + 25k   & 16.6    & 1.1   & 33.5    & 23.7    & 5.1   & 37.5 & 21.5    & 6.4   & 35.4 & 14.1    & 2.7   & 43.9 & 14.8    & 7.5   & 37.7 \\
     & 25k + 50k   & N/A  & N/A   & N/A  & N/A  & N/A   & N/A  & 22.9    & 6.6   & 36.9 & 14.9    & 2.8   & 47.9 & 15.3    & 7.4   & 40.0 \\
     & 100k + 1k   & 6.1  & 0.2   & 11.1    & 15.8    & 1.6   & 24.9 & 23.9    & 7.2   & 23.6 & 15.5    & 4.5   & 30.4 & 18.7    & 9.3   & 21.4 \\
     & 100k + 10k  & 12.7    & 0.7   & 25.1    & 22.4    & 3.7   & 33.4 & 26.1    & 7.4   & 31.8 & 16.7    & 4.9   & 38.7 & 19.0    & 9.8   & 31.6 \\
     & 100k + 25k  & 15.0    & 1.0   & 29.8    & 23.5    & 5.0   & 36.8 & 26.9    & 7.5   & 35.2 & 17.3    & 4.6   & 43.2 & 19.6    & 9.4   & 35.3 \\ \vspace{.7em}
     & 100k + 50k  & N/A  & N/A   & N/A  & N/A  & N/A   & N/A  & 27.1    & 7.8   & 36.3 & 17.6    & 4.4   & 45.7 & 19.9    & 8.8   & 37.5 \\
\multirow{6}{*}{CC + Bible}    & 25k + 1k  & 1.7  & 9.6   & 1.2  & 12.2    & 13.3  & 8.2  & 14.9    & 16.2  & 12.2 & 8.8  & 17.4  & 14.8 & 10.8    & 14.9  & 8.9  \\
     & 25k + 10k   & 3.6  & 22.4  & 2.0  & 12.8    & 23.9  & 10.7 & 14.8    & 27.0  & 12.6 & 9.8  & 33.1  & 16.9 & 11.3    & 26.1  & 9.4  \\
     & 25k + 25k   & 4.2  & 26.5  & 2.1  & 13.3    & 27.8  & 11.5 & 14.5    & 31.4  & 11.9 & 9.5  & 38    & 16.6 & 11.0    & 30.6  & 9.5  \\
     & 100k + 1k    & 4.3  & 7.3   & 2.9  & 12.1    & 13.1  & 11.2 & 23.9    & 16.4  & 19.0 & 15.4    & 18.8  & 26.8 & 18.7    & 16.0  & 17.0 \\
     & 100k + 10k  & 4.1  & 16.7  & 2.8  & 12.9    & 22.2  & 12.6 & 24.1    & 27.0  & 19.1 & 15.4    & 31.9  & 27.4 & 18.8    & 24.5  & 17.1 \\ \vspace{.7em}
     & 100k + 25k  & 5.8  & 22.6  & 3.5  & 11.9    & 25.0  & 10.2 & 23.8    & 30.6  & 19.3 & 15.4    & 35.7  & 26.8 & 18.1    & 28.3  & 16.2 \\
\multirow{12}{*}{PMI/Gvt + Bible (1st)}   & \multirow{1}{*}{1k + 1k}   
     &  1.1  & 9.1   & 6.2  & 7.7  & 13.1  & 19.4 & 8.0  & 15.0  & 18.5 & 4.4  & 18.5  & 21.4 & 3.7  & 14.5  & 18.9 \\
     & \multirow{1}{*}{1k + 10k}    
       & 3.7  & 22.7  & 11.8    & 8.8  & 23.8  & 20.3 & 7.5  & 26.7  & 18.1 & 4.9  & 33.4  & 22.4 & 3.9  & 25.1  & 17.4 \\
     & \multirow{1}{*}{1k + 25k}    
       & 4.4  & 27.8  & 12.5    & 9.3  & 27.6  & 20.5 & 7.1  & 30.8  & 18.1 & 4.2  & 36.6  & 19.8 & 3.8  & 30.9  & 18.3 \\
     & \multirow{1}{*}{10k + 1k}   
        & 11.2    & 12.6  & 29.3    & 17.2    & 14.2  & 34.0 & 15.0    & 16.7  & 31.0 & 9.1  & 19.4  & 39.3 & 7.5  & 14.9  & 33.7 \\
     & \multirow{1}{*}{10k + 10k}    & 11.9    & 22.8  & 29.3    & 17.9    & 24.2  & 34.1 & 14.7    & 27.2  & 30.5 & 9.5  & 33.8  & 39.0 & 7.9  & 25.9  & 33.0 \\
     & \multirow{1}{*}{10k + 25k}     & 11.8    & 27.3  & 29.3    & 18.2    & 27.8  & 34.2 & 14.5    & 31.1  & 30.8 & 9.4  & 37.5  & 38.5 & 7.6  & 30.4  & 33.0 \\
     & \multirow{1}{*}{25k + 1k}    & 14.5    & 13.0  & 33.8    & 19.8    & 15.0  & 37.8 & 17.4    & 16.5  & 35.3 & 11.3    & 19.2  & 43.8 & 9.5  & 14.7  & 37.7 \\
     & \multirow{1}{*}{25k + 10k}     & 14.9    & 23.0  & 33.7    & 20.9    & 24.1  & 37.4 & 17.4    & 27.2  & 35.6 & 11.9    & 33.8  & 44.9 & 9.9  & 25.9  & 37.3 \\
     & \multirow{1}{*}{25k + 25k}     & 14.3    & 26.2  & 32.7    & 20.8    & 27.7  & 37.7 & 17.2    & 31.4  & 35.1 & 11.5    & 37.7  & 43.7 & 9.9  & 30.4  & 37.0 \\
     & \multirow{1}{*}{50k + 1k}    & N/A  & N/A   & N/A  & N/A  & N/A   & N/A  & 19.4    & 17.2  & 37.4 & 12.4    & 20.0  & 48.6 & 11.2    & 14.1  & 40.1 \\
     & \multirow{1}{*}{50k + 10k}     & N/A  & N/A   & N/A  & N/A  & N/A   & N/A  & 20.1    & 26.8  & 37.1 & 12.7    & 33.4  & 48.2 & 11.3    & 25.6  & 39.7 \\ \vspace{.7em}
     & \multirow{1}{*}{50k + 25k}     & N/A  & N/A   & N/A  & N/A  & N/A   & N/A  & 19.2    & 31.0  & 36.2 & 12.6    & 37.4  & 46.9 & 11.2    & 29.5  & 39.4 \\
\multirow{12}{*}{PMI/Gvt + Bible (2nd)}   & \multirow{1}{*}{1k + 1k}    & 1.1  & 9.1   & 6.2  & 7.9  & 13.3  & 19.2 & 8.0  & 15.0  & 18.5 & 4.3  & 16.7  & 19.5 & 3.7  & 14.5  & 18.9 \\

     & \multirow{1}{*}{1k + 10k}  & 3.2  & 22.2  & 10.8    & 9.2  & 24.2  & 21.0 & 7.7  & 27.0  & 18.5 & 4.8  & 33.7  & 22.7 & 3.6  & 24.6  & 16.9 \\
     
     & \multirow{1}{*}{1k + 25k}  & 3.9  & 27.4  & 11.3    & 9.4  & 27.8  & 21.2 & 6.9  & 31.7  & 18.2 & 4.6  & 37.7  & 22.3 & 3.8  & 30.9  & 18.3 \\
     
     & \multirow{1}{*}{10k + 1k}   & 10.1    & 11.6  & 28.7    & 16.7    & 14.1  & 33.3 & 15.0    & 16.7  & 31.0 & 8.8  & 18.1  & 37.2 & 7.3  & 13.9  & 32.1 \\
     
     & \multirow{1}{*}{10k + 10k}    & 11.5    & 22.8  & 29.1    & 17.9    & 24.2  & 34.1 & 14.7    & 27.2  & 31.0 & 9.1  & 32.4  & 37.8 & 7.7  & 25.4  & 32.3 \\
     
     & \multirow{1}{*}{10k + 25k}    & 11.7    & 27.0  & 28.9    & 18.2    & 27.8  & 34.2 & 14.6    & 31.6  & 31.2 & 9.3  & 38.2  & 39.3 & 7.6  & 30.7  & 33.1 \\
     
     & \multirow{1}{*}{25k + 1k}   & 14.3    & 12.8  & 33.5    & 19.8    & 14.6  & 37.6 & 17.1    & 15.9  & 34.8 & 11.3    & 19.4  & 43.4 & 9.4  & 13.5  & 36.1 \\
     
     & \multirow{1}{*}{25k + 10k}    & 14.8    & 23.1  & 33.6    & 20.9    & 24.4  & 37.3 & 17.4    & 27.2  & 35.6 & 11.8    & 33.2  & 43.8 & 9.9  & 25.9  & 37.3 \\
     
     & \multirow{1}{*}{25k + 25k}    & 14.3    & 26.2  & 32.7    & 20.8    & 27.7  & 37.7 & 17.6    & 31.4  & 35.1 & 11.6    & 37.9  & 44.3 & 9.9  & 30.4  & 37.0 \\
     
     & \multirow{1}{*}{50k + 1k}   & N/A  & N/A   & N/A  & N/A  & N/A   & N/A  & 18.8    & 15.6  & 36.7 & 12.4    & 18.1  & 45.8 & 11.1    & 13.7  & 39.7 \\
     
     & \multirow{1}{*}{50k + 10k}    & N/A  & N/A   & N/A  & N/A  & N/A   & N/A  & 19.8    & 26.7  & 36.8 & 12.7    & 33.4  & 48.2 & 11.3    & 25.8  & 40.0 \\
     
     & \multirow{1}{*}{50k + 25k}    & N/A  & N/A   & N/A  & N/A  & N/A   & N/A  & 19.2    & 30.5  & 36.9 & 12.6    & 37.4  & 46.9 & 11.2    & 29.5  & 39.4 \\
      \bottomrule
\end{tabular*}
\caption{Results (spBLEU) for multi-domain FT. Here D1 refers to Domain 1 and D2 refers to Domain 2. The experiment for PMI/Gvt+Bible ran twice.}
\label{tab:mixed-domainFT}
\end{table*}
\begin{table*}[!htb]
\centering
\uniformtablesize
\setlength{\datgap}{2pt}
\setlength\tabcolsep{1.5pt}
\renewcommand{\arraystretch}{1.05}

\begin{tabular*}{\textwidth}{@{\extracolsep{\fill}} %
l l l l lll lll lll lll lll @{}}
\toprule
\multirow{2}{*}{\textbf{D1 + D2}}  & \multirow{2}{*}{\textbf{D1 + D2 size}} & \multirow{2}{*}{\textbf{D2}}  & \multirow{2}{*}{\textbf{D2 size}} & \multicolumn{3}{c}{\textbf{\lang{ka}}} & \multicolumn{3}{c}{\textbf{\lang{gu}}} & \multicolumn{3}{c}{\textbf{\lang{hi}}} & \multicolumn{3}{c}{\textbf{\lang{si}}} & \multicolumn{3}{c}{\textbf{\lang{ta}}} \\
\cmidrule(lr){5-7} \cmidrule(lr){8-10} \cmidrule(lr){11-13} \cmidrule(lr){14-16} \cmidrule(lr){17-19}
  &  &  &   & \textbf{FLORES}  & \textbf{Bible}    & \textbf{PMI}  & \textbf{FLORES}  & \textbf{Bible} & \textbf{PMI}  & \textbf{FLORES}  & \textbf{Bible} & \textbf{PMI}  & \textbf{FLORES}  & \textbf{Bible} & \textbf{Gvt}  & \textbf{FLORES}  & \textbf{Bible} & \textbf{Gvt}  \\ \midrule
\multirow{8}{*}{CC + PMI/Gvt}   & 25k + 1k    & \multirow{8}{*}{PMI/Gvt}  & 1k    & 5.6  & 0.1   & 14.8    & 16.5    & 2.2   & 24.6 & 16.0    & 5.6   & 21.1 & 10.6    & 2.2   & 25.8 & 12.4    & 6.7   & 21   \\
  & 25k + 10k   &  & 10k   & 13.7    & 0.4   & 30.4    & 21.7    & 4.7   & 34.7 & 19.3    & 5.9   & 31.2 & 12.7    & 2.8   & 39.2 & 13.4    & 7.1   & 33.1 \\
  & 25k + 25k   &  & 25k   & 16.2    & 1.8   & 33.7    & 24.1    & 5.3   & 37.6 & 21.3    & 6.2   & 35.3 & 13.3    & 2.3   & 44.7 & 14.7    & 7.4   & 37.9 \\
  & 25k + 50k   &  & 50k   & N/A  & N/A   & N/A  & N/A  & N/A   & N/A  & 22.9    & 6.6   & 36.7 & 14.4    & 2.7   & 49.3 & 13.6    & 6.1   & 40.7 \\
  & 100k + 1k   &  & 1k    & 10.0    & 0.6   & 17.8    & 21.9    & 4.0   & 26.8 & 24.1    & 7.0   & 25.7 & 16.9    & 4.8   & 32.6 & 21.1    & 9.9   & 24.2 \\
  & 100k + 10k  &  & 10k   & 15.0    & 2.1   & 30.4    & 24.7    & 5.6   & 35.4 & 25.2    & 7.3   & 32.6 & 17.7    & 5.4   & 40.0 & 18.3    & 8.5   & 34.3 \\
  & 100k + 25k  &  & 25k   & 17.1    & 2.2   & 34.1    & 24.9    & 5.9   & 38.2 & 25.8    & 6.9   & 35.4 & 17.5    & 5.0   & 45.2 & 17.0    & 7.1   & 38.1 \\ \vspace{.7em}
  & 100k + 50k  &  & 50k   & N/A  & N/A   & N/A  & N/A  & N/A   & N/A  & 25.2    & 6.4   & 37.2 & 16.9    & 4.7   & 48.5 & 17.0    & 6.5   & 41.1 \\
\multirow{6}{*}{CC + Bible}    & 25k + 1k    & \multirow{6}{*}{Bible}   & 1k    & 1.9  & 10.0  & 1.2  & 11.9    & 15.4  & 10.0 & 14.9    & 16.3  & 12.3 & 8.4  & 18.8  & 14.5 & 10.6    & 16.4  & 8.8  \\
  & 25k + 10k   &  & 10k   & 3.1  & 22.7  & 1.6  & 14.6    & 24.6  & 11.9 & 13.9    & 27.3  & 11.8 & 9.3  & 33.7  & 16.1 & 10.2    & 26.4  & 9.1  \\
  & 25k + 25k   &  & 25k   & 3.1  & 27.4  & 1.4  & 14.6    & 27.9  & 12.6 & 12.6    & 31.1  & 10.7 & 8.8  & 37.9  & 15.6 & 6.1  & 30.7  & 6.7  \\
  & 100k + 1k   &  & 1k    & 5.4  & 11.8  & 3.3  & 15.3    & 14.7  & 15.1 & 23.8    & 17.3  & 18.7 & 15.6    & 20.4  & 26.3 & 19.3    & 17.4  & 17.1 \\
  & 100k + 10k  &  & 10k   & 3.7  & 22.9  & 1.7  & 17.3    & 23.9  & 14.8 & 24.0    & 27.2  & 18.9 & 11.3    & 34.2  & 18.7 & 12.9    & 26.3  & 12.7 \\ \vspace{.7em}
  & 100k + 25k  &  & 25k   & 3.6  & 27.1  & 1.7  & 11.0    & 27.9  & 9.1  & 14.1    & 31.6  & 12.2 & 9.1  & 37.4  & 15.7 & 9.8  & 30.1  & 9.8  \\
\multirow{24}{*}{PMI/Gvt + Bible}   & \multirow{2}{*}{1k + 1k}   & Bible  & 1k    & 1.1  & 8.9   & 6.1  & 7.9  & 13.2  & 19.4 & 7.6  & 15.7  & 18.2 & 4.3  & 16.7  & 19.4 & 3.6  & 15.0  & 19.1 \\
  &    & PMI/Gvt    & 1k    & 1.2  & 8.6   & 6.6  & 8.1  & 13.0  & 20.1 & 8.2  & 14.9  & 19.3 & 4.5  & 18.3  & 22.3 & 3.9  & 12.0  & 20.1 \\
  & \multirow{2}{*}{1k + 10k}  & Bible  & 10k   & 2.6  & 22.6  & 7.6  & 9.5  & 24.1  & 21.0 & 6.9  & 27.2  & 18.1 & 4.3  & 34.1  & 21.4 & 3.0  & 25.7  & 13.3 \\
  &    & PMI/Gvt    & 1k    & 4.0  & 21.7  & 13.7    & 8.9  & 23.8  & 20.3 & 8.3  & 26.3  & 19.5 & 5.2  & 33.2  & 23.1 & 4.4  & 25.0  & 20.0 \\
  & \multirow{2}{*}{1k + 25k}  & Bible  & 25k   & 3.2  & 27.6  & 5.6  & 9.0  & 27.7  & 19.7 & 6.9  & 31.2  & 17.2 & 2.8  & 38.2  & 14.1 & 2.9  & 30.7  & 12.6 \\
  &    & PMI/Gvt    & 1k    & 5.1  & 25.6  & 15.5    & 10.2    & 27.2  & 21.5 & 8.2  & 30.1  & 19.2 & 5.6  & 35.6  & 24.5 & 4.3  & 30.2  & 19.7 \\
  & \multirow{2}{*}{10k + 1k}  & Bible  & 1k    & 9.3  & 12.7  & 26.8    & 16.7    & 14.5  & 33.5 & 14.8    & 16.6  & 31.0 & 8.4  & 19.5  & 37.0 & 7.2  & 14.1  & 32.1 \\
  &    & PMI/Gvt    & 10k   & 11.4    & 12.2  & 29.3    & 16.9    & 13.2  & 34.1 & 15.0    & 16.3  & 31.6 & 9.2  & 20.1  & 39.1 & 7.3  & 14.8  & 33.9 \\
  & \multirow{2}{*}{10k + 10k} & Bible  & 10k   & 9.6  & 23.4  & 25.9    & 17.0    & 24.3  & 33.7 & 14.3    & 27.5  & 30.5 & 6.6  & 33.2  & 31.7 & 7.0  & 25.8  & 31.6 \\
  &    & PMI/Gvt    & 10k   & 12.0    & 22.4  & 29.4    & 17.7    & 23.1  & 34.2 & 15.4    & 27.2  & 31.5 & 9.6  & 33.8  & 39.1 & 7.7  & 25.6  & 33.2 \\
  & \multirow{2}{*}{10k + 25k} & Bible  & 25k   & 7.1  & 27.6  & 17.0    & 17.1    & 27.9  & 33.1 & 13.5    & 31.2  & 30.2 & 9.1  & 37.9  & 39.0 & 5.0  & 30.6  & 25.8 \\
  &    & PMI/Gvt    & 10k   & 12.2    & 26.0  & 29.5    & 18.5    & 27.6  & 34.6 & 14.9    & 31.1  & 31.8 & 9.7  & 37.1  & 39.7 & 8.0  & 29.9  & 33.6 \\
  & \multirow{2}{*}{25k + 1k}  & Bible  & 1k    & 13.7    & 13.8  & 32.7    & 19.7    & 15.1  & 37.4 & 17.1    & 16.7  & 34.8 & 11.3    & 19.4  & 43.4 & 8.7  & 14.5  & 36.1 \\
  &    & PMI/Gvt    & 25k   & 14.5    & 12.9  & 34.0    & 20.0    & 14.0  & 37.8 & 17.8    & 15.4  & 34.9 & 11.3    & 15.4  & 45.3 & 9.8  & 14.7  & 38.1 \\
  & \multirow{2}{*}{25k + 10k} & Bible  & 10k   & 13.9    & 23.1  & 32.3    & 20.2    & 24.7  & 36.5 & 17.2    & 27.5  & 35.0 & 11.0    & 33.8  & 43.2 & 8.8  & 26.3  & 36.5 \\
  &    & PMI/Gvt    & 25k   & 14.7    & 20.8  & 33.7    & 21.2    & 23.1  & 37.8 & 17.7    & 26.8  & 35.4 & 11.7    & 32.8  & 44.8 & 9.9  & 25.7  & 37.3 \\
  & \multirow{2}{*}{25k + 25k} & Bible  & 25k   & 8.6  & 27.2  & 20.0    & 20.0    & 28.2  & 36.5 & 16.5    & 31.4  & 34.3 & 11.5    & 37.9  & 44.4 & 6.6  & 30.7  & 29.5 \\
  &    & PMI/Gvt    & 25k   & 15.5    & 21.4  & 34.4    & 20.6    & 25.0  & 37.6 & 18.0    & 29.5  & 35.1 & 11.8    & 35.7  & 44.0 & 10.5    & 25.3  & 38.2 \\
  & \multirow{2}{*}{50k + 1k}  & Bible  & 1k    & N/A  & N/A   & N/A  & N/A  & N/A   & N/A  & 18.0    & 17.6  & 36.5 & 11.9    & 20.1  & 45.5 & 10.1    & 14.8  & 39.5 \\
  &    & PMI/Gvt    & 50k   & N/A  & N/A   & N/A  & N/A  & N/A   & N/A  & 19.9    & 16.8  & 36.9 & 12.3    & 16.5  & 48.6 & 10.7    & 11.4  & 40.8 \\
  & \multirow{2}{*}{50k + 10k} & Bible  & 10k   & N/A  & N/A   & N/A  & N/A  & N/A   & N/A  & 15.7    & 27.5  & 32.6 & 12.2    & 34.3  & 47.4 & 10.3    & 26.0  & 39.2 \\
  &    & PMI/Gvt    & 50k   & N/A  & N/A   & N/A  & N/A  & N/A   & N/A  & 20.0    & 26.7  & 37.0 & 12.8    & 26.5  & 48.9 & 11.5    & 20.8  & 40.7 \\
  & \multirow{2}{*}{50k + 25k} & Bible  & 25k   & N/A  & N/A   & N/A  & N/A  & N/A   & N/A  & 18.0    & 30.8  & 35.9 & 8.1  & 38.0  & 35.9 & 7.1  & 30.5  & 31.4 \\
  &    & PMI/Gvt    & 50k   & N/A  & N/A   & N/A  & N/A  & N/A   & N/A  & 19.8    & 26.6  & 37.3 & 12.8    & 29.9  & 48.5 & 11.6    & 23.9  & 40.5 \\ \bottomrule
\end{tabular*}
\caption{Results (spBLEU) for multi-domain ITTL. Here D1 refers to Domain 1 and D2 refers to Domain 2.}
\label{tab:mixed-domainITFT}
\end{table*}

\begin{table*}[!htb]
\centering
\uniformtablesize
\setlength\tabcolsep{1.5pt}
\renewcommand{\arraystretch}{1.05}

\begin{tabular*}{\textwidth}{@{\extracolsep{\fill}} %
l l l l lll lll lll lll lll @{}}
\toprule
\multirow{2}{*}{\textbf{D1}}  & \multirow{2}{*}{\textbf{D1 size}} & \multirow{2}{*}{\textbf{D2}}   & \multirow{2}{*}{\textbf{D2 size}} & \multicolumn{3}{c}{\textbf{\lang{ka}}} & \multicolumn{3}{c}{\textbf{\lang{gu}}} & \multicolumn{3}{c}{\textbf{\lang{hi}}} & \multicolumn{3}{c}{\textbf{\lang{si}}} & \multicolumn{3}{c}{\textbf{\lang{ta}}} \\
\cmidrule(lr){5-7} \cmidrule(lr){8-10} \cmidrule(lr){11-13} \cmidrule(lr){14-16} \cmidrule(lr){17-19}
  &   &   &    & \textbf{FLORES}  & \textbf{Bible} & \textbf{PMI}  & \textbf{FLORES}  & \textbf{Bible} & \textbf{PMI}  & \textbf{FLORES}  & \textbf{Bible}& \textbf{PMI}  & \textbf{FLORES}  & \textbf{Bible} & \textbf{Gvt}  & \textbf{FLORES}  & \textbf{Bible} & \textbf{Gvt}  \\ \midrule
\multirow{8}{*}{CC} & 25k    & \multirow{8}{*}{PMI/Gvt}  & 1k  & 0.2  & -0.4  & 0.9  & -0.7    & -1.5  & 0.5  & 3.0  & 2.0   & 0.8  & 0.8  & 0.2   & -0.4 & 2.1  & 0.7   & 0.0  \\
& 25k    &   & 10k    & 1.2  & -1.2  & -0.5 & 2.8  & -1.2  & -0.3 & 2.8  & 2.6   & -0.6 & 1.6  & 0.5   & -1.3 & 3.3  & 3.4   & -0.7 \\
& 25k    &   & 25k    & 1.6  & -0.8  & -0.8 & 2.2  & 0.0   & -0.8 & 3.2  & 3.1   & -0.1 & 2.0  & 0.8   & -0.5 & 3.7  & 4.2   & 0.5  \\
& 25k    &   & 50k    & N/A  & N/A   & N/A  & N/A  & N/A   & N/A  & 3.2  & 2.8   & -0.1 & 2.1  & 0.8   & -1.8 & 3.2  & 4.6   & -0.1 \\
& 100k   &   & 1k  & -4.1    & -0.9  & -5.1 & -5.1    & -2.1  & -1.1 & 3.0  & 1.1   & -0.4 & -0.8    & -0.1  & -0.6 & -1.6    & -0.1  & -2.5 \\
& 100k   &   & 10k    & -1.9    & -1.1  & -5.2 & -0.1    & -1.5  & -1.7 & 1.8  & 2.5   & -0.7 & 1.5  & 0.8   & -2.7 & 3.6  & 3.6   & -2.5 \\
& 100k   &   & 25k    & -1.3    & -1.1  & -4.3 & 0.6  & -0.4  & -1.7 & 5.8  & 3.1   & 0.0  & 2.1  & 1.0   & -1.1 & 4.6  & 4.4   & -2.3 \\ \vspace{.7em}
& 100k   &   & 50k    & N/A  & N/A   & N/A  & N/A  & N/A   & N/A  & 4.6  & 3.4   & -0.7 & 2.4  & 0.9   & -3.0 & 5.0  & 4.3   & -3.0 \\
\multirow{6}{*}{CC}   & 25k    & \multirow{6}{*}{Bible}    & 1k  & 0/8  & 0.0   & 0.5  & 2.6  & -0.6  & -0.3 & 7.8  & 0.1   & 5.4  & 5.3  & -1.9  & 8.6  & 3.6  & 0.4   & 2.2  \\
  & 25k    &   & 10k    & 1.5  & -0.7  & 1.1  & 6.2  & 0.2   & 5.6  & 10.8    & 0.6   & 9.6  & 7.6  & -0.4  & 14.7 & 8.3  & 0.0   & 6.9  \\
  & 25k    &   & 25k    & 1.8  & -0.8  & 1.1  & 7.7  & 0.9   & 7.4  & 10.7    & -0.1  & 9.5  & 7.5  & 0.1   & 15.6 & 8.1  & -0.2  & 7.9  \\
  & 100k   &   & 1k  & 0.2  & -3.9  & 0.6  & -2.5    & -0.2  & -0.7 & 7.0  & -0.1  & 4.5  & 4.4  & -1.3  & 7.8  & 1.5  & -0.3  & 1.7  \\
  & 100k   &   & 10k    & 1.3  & -6.2  & 1.4  & 4.6  & -2.0  & 5.7  & 17.5    & 0.2   & 13.6 & 11.5    & -1.5  & 22.0 & 12.9    & -1.7  & 11.1 \\ \vspace{.7em}
  & 100k   &   & 25k    & 3.0  & -4.8  & 2.3  & 5.0  & -2.8  & 4.8  & 19.0    & -1.1  & 16.0 & 12.4    & -2.2  & 24.2 & 13.9    & -2.7  & 13.0 \\
\multirow{12}{*}{PMI/Gvt}  & 1k  & \multirow{12}{*}{Bible} & 1k  & 0.6  & -0.3  & 5.2  & 1.7  & 0.1   & 3.8  & 3.2  & -0.5  & 7.1  & 2.0  & 1.3   & 8.3  & 1.5  & 3.5   & 5.2  \\
& 1k  & \multicolumn{1}{c}{}   & 10k    & 2.0  & 0.7   & 10.9 & 3.7  & -0.4  & 13.7 & 4.2  & -0.5  & 15.2 & 3.0  & -0.7  & 19.2 & 1.6  & -1.2  & 13.7 \\
& 1k  & \multicolumn{1}{c}{}   & 25k    & 2.1  & 0.8   & 11.4 & 4.1  & 0.3   & 15.4 & 3.7  & -0.7  & 15.6 & 2.3  & -1.0  & 18.0 & 1.6  & 0.4   & 15.6 \\
& 10k    & \multicolumn{1}{c}{}   & 1k  & 7.0  & 0.2   & 16.4 & 5.4  & -0.5  & 7.7  & 4.8  & 0.9   & 6.3  & 3.8  & 1.9   & 9.5  & 3.7  & 1.3   & 10.4 \\
& 10k    & \multicolumn{1}{c}{}   & 10k    & 9.1  & -0.3  & 26.4 & 10.5    & 0.0   & 20.5 & 9.6  & 0.3   & 21.6 & 7.3  & 1.4   & 31.0 & 5.3  & 0.1   & 24.9 \\
& 10k    & \multicolumn{1}{c}{}   & 25k    & 8.9  & 0.2   & 27.2 & 11.7    & 0.1   & 25.1 & 10.3    & -0.4  & 26.2 & 7.3  & -0.3  & 34.1 & 5.1  & -0.2  & 28.2 \\
& 25k    & \multicolumn{1}{c}{}   & 1k  & 6.9  & 0.3   & 12.5 & 4.5  & 0.7   & 6.0  & 4.6  & 1.2   & 6.0  & 3.4  & 0.3   & 7.1  & 4.1  & 1.3   & 9.3  \\
& 25k    & \multicolumn{1}{c}{}   & 10k    & 11.2    & -0.1  & 27.6 & 11.2    & 0.2   & 17.3 & 11.4    & 0.2   & 22.8 & 8.6  & 0.7   & 30.4 & 6.7  & 1.1   & 23.5 \\
& 25k    & \multicolumn{1}{c}{}   & 25k    & 11.1    & -1.3  & 29.9 & 12.8    & 0.3   & 24.3 & 12.9    & -0.2  & 28.7 & 9.1  & -0.4  & 37.2 & 7.1  & -0.5  & 29.7 \\
& 50k    & \multicolumn{1}{c}{}   & 1k  & N/A  & N/A   & N/A  & N/A  & N/A   & N/A  & 4.7  & 1.2   & 6.1  & 4.0  & 0.8   & 8.1  & 4.3  & 0.4   & 7.6  \\
& 50k    & \multicolumn{1}{c}{}   & 10k    & N/A  & N/A   & N/A  & N/A  & N/A   & N/A  & 13.4    & 0.1   & 21.3 & 8.4  & -0.1  & 30.4 & 7.5  & 0.6   & 22.4 \\ \vspace{.7em}
& 50k    & \multicolumn{1}{c}{}   & 25k    & N/A  & N/A   & N/A  & N/A  & N/A   & N/A  & 14.4    & -0.8  & 28.0 & 9.6  & 0.1   & 38.3 & 7.9  & -1.0  & 29.6 \\
\multirow{12}{*}{Bible}   & 1k  & \multirow{12}{*}{PMI/Gvt}    & 1k  & 0.1  & 7.5   & 0.3  & -0.4    & 3.7   & 0.4  & -0.4    & 6.4   & 0.2  & -0.6    & 6.4   & -1.4 & -0.2    & 8.4   & -1.0 \\
  & 1k  &   & 10k    & 0.1  & 9.9   & 0.2  & -0.6    & 6.6   & -1.2 & 0.3  & 11.4  & -0.4 & -0.5    & 12.6  & -1.4 & 0.3  & 9.3   & -0.6 \\
  & 1k  &   & 25k    & 0.3  & 10.8  & -0.3 & -0.4    & 7.8   & -0.4 & 0.2  & 11.2  & -0.3 & 0.0  & 15.9  & -1.6 & -0.1    & 10.5  & -1.0 \\
  & 1k  &   & 50k    & N/A  & N/A   & N/A  & N/A  & N/A   & N/A  & -0.3    & 10.7  & -0.5 & 0.1  & 14.6  & -3.7 & -0.1    & 11.1  & -1.0 \\
  & 10k    &   & 1k  & -0.8    & 10.2  & -2   & -1.1    & 4.8   & 0.8  & -1.4    & 6.9   & -0.7 & -1.3    & 8.6   & 0.3  & -1.2    & 5.3   & -2.0 \\
  & 10k    &   & 10k    & 0.1  & 16.8  & -0.6 & -0.2    & 10.3  & -0.7 & -0.7    & 13.0  & -0.4 & -0.9    & 17.3  & -1.5 & -0.2    & 13.2  & -0.9 \\
  & 10k    &   & 25k    & 0.5  & 18.8  & -0.6 & 0.4  & 12.4  & -0.6 & -0.5    & 16.4  & 0.9  & 0.0  & 21.0  & -0.4 & 0.0  & 15.7  & 0.0  \\
  & 10k    &   & 50k    & N/A  & N/A   & N/A  & N/A  & N/A   & N/A  & -0.1    & 16.6  & 0.1  & 0.0  & 24.1  & -0.8 & -0.3    & 17.3  & 0.2  \\
  & 25k    &   & 1k  & -0.6    & 11.1  & -1.6 & -1.1    & 5.2   & 1.3  & -1.9    & 7.8   & -0.5 & -1.2    & 9.0   & 0.2  & -1.1    & 6.6   & 0.0  \\
  & 25k    &   & 10k    & 0.1  & 18.6  & -0.9 & -2.3    & 15.8  & -3.7 & -0.8    & 14.4  & 0.0  & -0.7    & 18.6  & 0.2  & -0.4    & 15.8  & -0.5 \\
  & 25k    &   & 25k    & 0.1  & 19.7  & -0.1 & 0.3  & 13.0  & 0.1  & -0.8    & 16.2  & 0.7  & -0.3    & 23.0  & 0.1  & -0.1    & 17.2  & -0.2 \\
  & 25k    &   & 50k    & N/A  & N/A   & N/A  & N/A  & N/A   & N/A  & -0.8    & 17.8  & 0.1  & -0.3    & 24.7  & -1.7 & 0.0  & 18.1  & -0.5 \\ \bottomrule
\end{tabular*}
\caption{Difference in spBLEU scores between multi-domain FT and single-domain ITTL. The baseline was single-domain ITTL. Here D1 refers to Domain 1 and D2 refers to Domain 2. For multi-domain FT D1 and D2 were combined and fine-tuned.}
\label{tab:mixedFT-single-domain}
\end{table*}
\begin{table}[!htb]
\centering
\resizebox{\textwidth}{!}{
\begin{tabular}{@{}llll @{\hspace{\datgap}} ll  @{\hspace{\datgap}} lll @{\hspace{\datgap}} lll @{\hspace{\datgap}} lll @{\hspace{\datgap}} lll @{\hspace{\datgap}} lll}
\toprule
\multicolumn{4}{c}{\textbf{Stage 1}}  & \multicolumn{2}{c}{\textbf{Stage 2}}    & \multicolumn{3}{c}{\textbf{\lang{ka}}} & \multicolumn{3}{c}{\textbf{\lang{gu}}} & \multicolumn{3}{c}{\textbf{\lang{hi}}} & \multicolumn{3}{c}{\textbf{\lang{si}}} & \multicolumn{3}{c}{\textbf{\lang{ta}}} \\
\cmidrule(r{\datgap}){1-4} \cmidrule(r{\datgap}){5-6} \cmidrule(r{\datgap}){7-9} \cmidrule(r{\datgap}){10-12} \cmidrule(r{\datgap}){13-15} \cmidrule(r{\datgap}){16-18} \cmidrule(){19-21}
  \textbf{D1} & \textbf{D1 size} & \textbf{D2} & \textbf{D2 size} & \textbf{D} & \textbf{D size} & \textbf{FLORES}  & \textbf{Bible} & \textbf{PMI}  & \textbf{FLORES}  & \textbf{Bible} & \textbf{PMI}  & \textbf{FLORES}  & \textbf{Bible} & \textbf{PMI}  & \textbf{FLORES}  & \textbf{Bible} & \textbf{Gvt}  & \textbf{FLORES}  & \textbf{Bible} & \textbf{Gvt}  \\ \midrule
\multirow{8}{*}{CC} & 25k    & \multirow{8}{*}{PMI/Gvt}  & 1k    & \multirow{8}{*}{PMI/Gvt}  & 1k    & 0.9  & 0.1  & 1.0  & 0.4    & 0.1  & 0.3  & 0.9  & 0.3   & 2.1  & 0.2  & 0.1   & 1.0 & -0.3  & 0.0   & 0.3  \\
& 25k    &   & 10k    &   & 10k    & 0.1  & 0.0  & 0.4 & -0.5  & 1.3  & 0.3 & 0.5  & 0.4   & 1.3 & 0.3  & 0.1   & 0.4 & -0.1  & 0.0   & 0.0 \\
& 25k    &   & 25k    &   & 25k    & -0.4  & 0.7  & 0.2 & 0.4  & 0.2   & 0.1 & -0.8  & -0.4   & 0.8 & -0.1  & -0.1   & 0.2 & -0.2  & -0.2   & -0.1  \\
& 25k    &   & 50k    &   & 50k    & N/A  & N/A   & N/A  & N/A  & N/A   & N/A  & -0.5  & -0.1   & 1.4 & -1.7  & -1.3   & 0.7 & 0.0  & 0.0   & 0.0 \\
& 100k   &   & 1k   &   & 1k  & 3.9    & 0.4  & 6.7 & 6.1    & 2.4  & 1.9 & 1.4  & 0.3   & 2.2 & 2.4    & 0.6  & 2.8 & 0.2    & -0.2  & 2.1 \\
& 100k   &   & 10k   &   & 10k    & 2.3    & 1.4  & 5.3 & 2.3   & 1.9  & 2.0 & 1.0  & 0.5  & 1.3 & -0.7  & -1.3   & 2.7 & -0.9  & -0.1   & 0.8 \\
& 100k   &   & 25k   &   & 25k    & 2.1    & 1.2  & 4.3 & 1.4  & 0.9  & 1.4 & 0.2  & 0.4   & 2.0  & -2.6  & -2.3  & 2.8 & -1.1  & -0.6   & 0.2 \\ \vspace{.7em}
& 100k   &   & 50k   &   & 50k    & N/A  & N/A   & N/A  & N/A  & N/A   & N/A  & -0.7  & 0.3   & 2.8 & -2.9  & -2.3   & 3.6 & -1.9  & -1.4   & 0.9 \\
\multirow{6}{*}{CC}   & 25k    & \multirow{6}{*}{Bible}    & 1k    & \multirow{6}{*}{Bible}    & 1k  & 0.2  & 0.4   & 0.0  & -0.3  & 2.1  & 1.8 & -0.4  & 1.4   & -0.3  & -0.2  & 1.5  & -0.1  & 0.0  & 0.1   & 0.1  \\
  & 25k    &   & 10k    &   & 10k    & -0.5  & 0.3  & -0.4  & 1.8  & 0.7   & 1.2  & -0.5    & 0.6   & -0.8  & -1.1  & 0.3  & -0.3 & -0.9  & 0.3   & -0.8  \\
  & 25k    &   & 25k    &   & 25k    & -1.1  & 0.9  & -0.7  & 1.3  & 0.1   & 1.1  & -0.7    & -0.1  & -1.0  & -4.9  & 0.1   & -2.8 & -1.9  & -0.3  & -1.2  \\
  & 100k   &   & 1k   &   & 1k  & 1.1  & 4.5  & 0.4  & 3.2    & 1.6  & 3.9 & 0.2  & 1.6  & -0.5  & 0.6  & 1.4  & 0.1  & -0.1  & 0.9  & -0.3  \\
  & 100k   &   & 10k   &   & 10k    & -0.4  & 6.2  & -1.1    & 4.4   & 1.7 & 2.2   & -4.1  & 2.3 & -8.7    & -5.9  & 1.8 & -4.4  & -0.1  & 0.2  & -0.2   \\ \vspace{.7em}
  & 100k   &   & 25k   &   & 25k    & -2.2  & 4.5  & -1.8  & -0.9  & 2.9  & -1.1  & -6.3    & 1.7  & -11.1 & -8.3    & 1.8  & -6.4 & -9.7   & 1.0  & -7.1 \\
\multirow{24}{*}{PMI/Gvt}  & \multirow{2}{*}{1k}  & \multirow{24}{*}{Bible} & \multirow{2}{*}{1k}   & Bible & 1k  & 0.0  & -0.2  & -0.1  & 0.0  & -0.1   & 0.2  & 0.0  & 0.0  & -0.1  & -0.1  & 0.5   & 0.2  & -0.4  & 0.7   & -0.3  \\
& \multicolumn{1}{c}{}   & \multicolumn{1}{c}{}   & \multicolumn{1}{c}{}  & PMI/Gvt & 1k    & 0.1  & -0.5   & 0.4 & 0.4  & -0.1  & 0.7 & 0.1  & -0.2  & 0.9 & 0.2  & -2.5  & 1.2 & 0.2  & -0.1  & 0.8 \\
& \multirow{2}{*}{1k}   & \multicolumn{1}{c}{}   & \multirow{2}{*}{10k}  & Bible & 10k  & -0.6  & 0.4   & -3.2 & 0.3  & -0.1   & 0.0 & -0.5  & 0.4  & -1.3 & -0.6  & 1.1  & -3.6 & -0.8  & 0.2   & -0.4 \\
& \multicolumn{1}{c}{}   & \multicolumn{1}{c}{}   & \multicolumn{1}{c}{}  & PMI/Gvt & 1k    & 0.3  & -1.0   & 1.9 & 0.1  & 0.0  & 0.0 & 0.3  & -0.2  & 0.7 & 0.5  & -0.1  & 2.6 & 0.8  & -0.4  & 1.4 \\
& \multirow{2}{*}{1k}   & \multicolumn{1}{c}{}   & \multirow{2}{*}{25k}  & Bible & 25k  & -0.7  & 0.2   & -5.7 & -0.4  & -0.1   & -1.5 & -1.8  & 0.5  & -8.2 & -0.9  & -0.2  & -5.7 & -0.4  & -0.5   & -1.0 \\
& \multicolumn{1}{c}{}   & \multicolumn{1}{c}{}   & \multicolumn{1}{c}{}  & PMI/Gvt & 1k    & 0.7  & -2.2   & 3.0 & 0.9  & -0.4  & 1.0 & 1.4  & -1.0  & 4.7 & 0.5  & -0.7  & 1.4 & 1.1  & -0.7  & 1.1 \\
& \multirow{2}{*}{10k}   & \multicolumn{1}{c}{}   & \multirow{2}{*}{1k}  & Bible & 1k  & -0.8  & 1.1   & -1.9 & 0.0  & 0.4   & 0.2 & -0.4  & 1.4  & -0.2 & -0.1  & 0.2  & 0.0 & -0.2  & -0.1   & 0.0 \\
& \multicolumn{1}{c}{}   & \multicolumn{1}{c}{}   & \multicolumn{1}{c}{}  & PMI/Gvt & 10k    & 0.2  & -0.4   & 0.0 & -0.3  & -1.0  & 0.1 & 0.1  & 0.7  & -0.2 & -0.2 & -0.1  & 0.2  & 0.0 & -0.4  & 0.6 \\
& \multirow{2}{*}{10k}   & \multicolumn{1}{c}{}   & \multirow{2}{*}{10k}  & Bible & 10k  & -1.9  & 0.6   & -3.2 & -0.9  & 0.1   & -0.4 & -2.5  & 0.8  & -6.1 & -0.7  & 0.4  & -0.7 & -0.4  & 0.3   & -0.5 \\
& \multicolumn{1}{c}{}   & \multicolumn{1}{c}{}   & \multicolumn{1}{c}{}  & PMI/Gvt & 10k    & 0.1  & -0.4   & 0.1 & -0.2  & -1.1  & 0.1 & 0.1  & 0.0  & 0.1 & -0.2 & -0.3  & 0.2  & 0.7 & 0.0  & 1.0 \\
& \multirow{2}{*}{10k}   & \multicolumn{1}{c}{}   & \multirow{2}{*}{25k}  & Bible & 25k  & -4.6  & 0.6   & -11.9 & -1.1  & 0.1   & -1.1 & -0.2  & -0.3  & -0.3 & -2.6  & -0.1  & -7.3 & -1.1  & -0.4   & -1.0 \\ \vspace{.7em}
& \multicolumn{1}{c}{}   & \multicolumn{1}{c}{}   & \multicolumn{1}{c}{}  & PMI/Gvt & 10k    & 0.4  & -1.3   & 0.2 & 0.3  & -0.2  & 0.4 & 0.3  & -0.4  & 1.2 & 0.4 & -0.5  & 0.6  & 0.4 & 0.0  & 1.0 \\
& \multirow{2}{*}{25k}   & \multicolumn{1}{c}{}   & \multirow{2}{*}{1k}  & Bible & 1k  & -0.6    & 1.0   & -0.8 & -0.1  & 0.5   & -0.2 & 0.0  & 0.0  & 0.0 & -0.7  & 1.0  & 0.0 & 0.0  & 0.8   & 0.0 \\ \vspace{.7em}
& \multicolumn{1}{c}{}   & \multicolumn{1}{c}{}   & \multicolumn{1}{c}{}  & PMI/Gvt & 25k    & 0.0  & -0.1   & 0.2 & 0.2  & -1.0  & 0.0 & 0.0  & -3.8  & 1.5 & 0.3 & 0.0  & 0.4  & 0.4 & -1.1  & -0.4 \\
& \multirow{2}{*}{25k}   & \multicolumn{1}{c}{}   & \multirow{2}{*}{10k}  & Bible & 10k  & -0.9    & 0.0  & -1.3 & -0.7  & 0.3   & -0.8 & -0.8  & 0.6  & -0.6 & -1.1  & 0.4  & -0.8 & -0.2  & 0.3   & -0.6 \\ \vspace{.7em}
& \multicolumn{1}{c}{}   & \multicolumn{1}{c}{}   & \multicolumn{1}{c}{}  & PMI/Gvt & 25k    & -0.2  & -2.2   & 0.0 & 0.3  & -1.0  & 0.4 & -0.2  & -1.0  & -0.1 & 0.0 & -0.2  & 0.0  & 0.3 & -0.4  & -0.2 \\

& \multirow{2}{*}{25k}   & \multicolumn{1}{c}{}   & \multirow{2}{*}{25k}  & Bible & 25k  & -5.7  & 1.0   & -12.7 & -0.8  & 0.5   & -1.2 & -0.1  & 0.0  & 0.1 & -3.3  & 0.3  & -7.5 & -1.1  & 0.0   & -0.8 \\
& \multicolumn{1}{c}{}   & \multicolumn{1}{c}{}   & \multicolumn{1}{c}{}  & PMI/Gvt & 25k    & 1.2  & -4.8   & 1.7 & -0.2  & -2.7  & -0.1 & 0.3  & -2.0  & 0.3 & 0.6 & -5.1  & 1.2  & 0.8 & -1.9  & 0.0 \\
& \multirow{2}{*}{50k}   & \multicolumn{1}{c}{}   & \multirow{2}{*}{1k}  & Bible & 1k  & N/A  & N/A   & N/A & N/A  & N/A   & N/A & -0.5  & 2.0 & -0.3  & -1.0  & 1.1 & -0.2  & -0.8  & 2.0   & -0.2 \\ \vspace{.7em}
& \multicolumn{1}{c}{}   & \multicolumn{1}{c}{}   & \multicolumn{1}{c}{}  & PMI/Gvt & 50k    & N/A  & N/A   & N/A & N/A  & N/A   & N/A & -0.1  & -3.5  & 0.0 & -0.5 & -2.7  & 0.7  & 0.5 & -0.4  & -0.5 \\
& \multirow{2}{*}{50k}   & \multicolumn{1}{c}{}   & \multirow{2}{*}{10k}  & Bible & 10k  & N/A  & N/A   & N/A & N/A  & N/A   & N/A & -0.5    & 0.9  & -0.8 & -1.0  & 0.2  & -0.8 & -4.1  & 0.8   & -4.2 \\ \vspace{.7em}
& \multicolumn{1}{c}{}   & \multicolumn{1}{c}{}   & \multicolumn{1}{c}{}  & PMI/Gvt & 50k    & N/A  & N/A   & N/A & N/A  & N/A   & N/A & 0.1  & -6.9  & 0.7 & 0.2 & -4.8  & 1.0  & -0.1 & -0.1  & -0.1 \\
& \multirow{2}{*}{50k}   & \multicolumn{1}{c}{}   & \multirow{2}{*}{25k}  & Bible & 25k  & N/A  & N/A   & N/A & N/A  & N/A   & N/A & -4.5  & 0.6  & -11.0 & -4.1  & 1.0 & -0.8 & -1.2  & 0.3   & -1.0 \\ \vspace{.7em}
& \multicolumn{1}{c}{}   & \multicolumn{1}{c}{}   & \multicolumn{1}{c}{}  & PMI/Gvt & 50k    & N/A  & N/A   & N/A & N/A  & N/A   & N/A & 0.2  & -7.5  & 1.6 & 0.4 & -5.6  & 1.1  & 0.6 & -4.4  & 1.1 \\ \bottomrule
\end{tabular}
}
\caption{Difference in spBLEU scores between multi-domain FT and multi-domain ITTL. The baseline was multi-domain FT. Here D1 refers to Domain 1 in stage 1 and D2 refers to Domain 2 in stage 1, while D refers to Domain in stage 2 for multi-domain ITTL. D1 and D2 were combined in stage 1 and multi-domain FT was only fine-tuned in stage 1.}
\label{tab:mixedFT-mixed-domainITFT}
\end{table}
\begin{table*}[!htb]
\centering
\uniformtablesize 
\setlength\tabcolsep{2pt}
\renewcommand{\arraystretch}{1.1}

\begin{tabular*}{\textwidth}{@{\extracolsep{\fill}} l l l l lll lll lll lll lll @{}}
\toprule
\multirow{2}{*}{\textbf{D1}}                          & \multirow{2}{*}{\textbf{D1 size}} & \multirow{2}{*}{\textbf{D2}}                       & \multirow{2}{*}{\textbf{D2 size}} & \multicolumn{3}{c}{\textbf{\lang{ka}}} & \multicolumn{3}{c}{\textbf{\lang{gu}}} & \multicolumn{3}{c}{\textbf{\lang{hi}}} & \multicolumn{3}{c}{\textbf{\lang{si}}} & \multicolumn{3}{c}{\textbf{\lang{ta}}} \\
\cmidrule(r{\datgap}){5-7} \cmidrule(r{\datgap}){8-10} \cmidrule(r{\datgap}){11-13} \cmidrule(r{\datgap}){14-16} \cmidrule(){17-19}
   &   & &    & \textbf{F}  & \textbf{B} & \textbf{P}  & \textbf{F}  & \textbf{B} & \textbf{P}  & \textbf{F}  & \textbf{B} & \textbf{P}  & \textbf{F}  & \textbf{B} & \textbf{G}  & \textbf{F}  & \textbf{B} & \textbf{G}\\ \midrule   
\multirow{8}{*}{CC} & 25k   & \multirow{8}{*}{PMI/Gvt}    & 1k    & 1.1 & -0.3  & 1.9  & -0.3    & -1.4  & 0.8  & 2.7 & 2.0   & 1.1  & 1.7 & 0.5   & 1.7  & 2.3 & 0.8   & 1.0  \\
& 25k   & & 10k   & 1.3 & -1.2  & -0.1 & 2.3 & 0.1   & 0.0  & 2.7 & 2.6   & -0.6 & 2.1 & 0.9   & 0.0  & 3.6 & 3.5   & -0.3 \\
& 25k   & & 25k   & 1.2 & -0.1  & -0.6 & 2.6 & 0.2   & -0.7 & 3.0 & 2.9   & -0.2 & 1.2 & 0.4   & 0.3  & 3.6 & 4.1   & 0.7  \\
& 25k   & & 50k   & N/A & N/A   & N/A  & N/A & N/A   & N/A  & 3.0 & 2.8   & -0.3 & 1.6 & 0.7   & -0.4 & 1.5 & 3.3   & 0.6  \\
& 100k  & & 1k    & -0.2    & -0.5  & 1.6  & 1.0 & 0.3   & 0.8  & 2.0 & 0.9   & 1.7  & 0.6 & 0.2   & 1.6  & 0.8 & 0.5   & 0.3  \\
& 100k  & & 10k   & 0.4 & 0.3   & 0.1  & 2.2 & 0.4   & 0.3  & 4.9 & 2.4   & 0.1  & 2.5 & 1.3   & -1.4 & 2.9 & 2.3   & 0.2  \\
& 100k  & & 25k   & 0.8 & 0.1   & 0.0  & 2.0 & 0.5   & -0.3 & 4.4 & 2.5   & 0.2  & 2.3 & 1.4   & 0.9  & 2.0 & 2.1   & 0.5  \\ \vspace{.7em}
& 100k  & & 50k   & N/A & N/A   & N/A  & N/A & N/A   & N/A  & 2.7 & 2.0   & 0.2  & 1.7 & 1.2   & -0.2 & 2.1 & 2.0   & 0.6  \\
\multirow{6}{*}{CC} & 25k   & \multirow{6}{*}{Bible}  & 1k    & 1.0 & 0.4   & 0.5  & 2.3 & 1.5   & 1.5  & 7.8 & 0.2   & 5.5  & 4.9 & -0.5  & 8.3  & 3.4 & 1.9   & 2.1  \\
   & 25k   & & 10k   & 1.0 & -0.4  & 0.7  & 8.0 & 0.9   & 6.8  & 9.9 & 0.9   & 8.8  & 7.1 & 0.2   & 13.9 & 7.2 & 0.3   & 6.6  \\
   & 25k   & & 25k   & 0.7 & 0.1   & 0.4  & 9.0 & 1.0   & 8.5  & 8.8 & -0.4  & 8.3  & 6.8 & 0.0   & 14.6 & 3.2 & -0.1  & 5.1  \\
   & 100k  & & 1k    & 1.3 & 0.6   & 1.0  & 0.7 & 1.4   & 3.2  & 6.9 & 0.8   & 4.2  & 4.6 & 0.3   & 7.3  & 2.1 & 1.1   & 1.8  \\
   & 100k  & & 10k   & 0.9 & 0.0   & 0.3  & 9.0 & -0.3  & 7.9  & 17.4    & 0.4   & 13.4 & 7.4 & 0.8   & 13.3 & 7.0 & 0.1   & 6.7  \\ \vspace{.7em}
   & 100k  & & 25k   & 0.8 & -0.3  & 0.5  & 4.1 & 0.1   & 3.7  & 9.3 & -0.1  & 8.9  & 6.1 & -0.5  & 13.1 & 5.6 & -0.9  & 6.6  \\
\multirow{12}{*}{PMI/Gvt}  & 1k    & \multirow{12}{*}{Bible} & 1k    & 0.6 & -0.5  & 5.1  & 1.9 & 0.2   & 3.8  & 2.8 & 0.2   & 6.8  & 1.9 & -0.5  & 6.3  & 1.4 & 4.0   & 5.4  \\
& 1k    & \multicolumn{1}{c}{}    & 10k   & 0.9 & 0.6   & 6.7  & 4.4 & -0.1  & 14.4 & 3.6 & 0.0   & 15.2 & 2.4 & 0.0   & 18.2 & 0.7 & -0.6  & 9.6  \\
& 1k    & \multicolumn{1}{c}{}    & 25k   & 0.9 & 0.6   & 4.5  & 3.8 & 0.4   & 14.6 & 3.1 & -0.3  & 14.7 & 0.9 & 0.6   & 12.3 & 0.7 & 0.2   & 9.9  \\
& 10k   & \multicolumn{1}{c}{}    & 1k    & 5.1 & 0.3   & 13.9 & 4.9 & -0.2  & 7.2  & 4.6 & 0.8   & 6.3  & 3.1 & 2.0   & 7.2  & 3.4 & 0.5   & 8.8  \\
& 10k   & \multicolumn{1}{c}{}    & 10k   & 6.8 & 0.3   & 23.0 & 9.6 & 0.1   & 20.1 & 9.2 & 0.6   & 21.6 & 4.4 & 0.8   & 23.7 & 4.4 & 0.0   & 23.5 \\
& 10k   & \multicolumn{1}{c}{}    & 25k   & 4.2 & 0.5   & 14.9 & 10.6    & 0.2   & 24.0 & 9.3 & -0.3  & 25.6 & 7.0 & 0.1   & 34.6 & 2.5 & 0.0   & 21.0 \\
& 25k   & \multicolumn{1}{c}{}    & 1k    & 6.1 & 1.1   & 11.4 & 4.4 & 0.8   & 5.6  & 4.3 & 1.4   & 5.5  & 3.4 & 0.5   & 6.7  & 3.3 & 1.1   & 7.7  \\
& 25k   & \multicolumn{1}{c}{}    & 10k   & 10.2    & 0.0   & 26.2 & 10.5    & 0.8   & 16.4 & 11.2    & 0.5   & 22.2 & 7.7 & 0.7   & 28.7 & 5.6 & 1.5   & 22.7 \\
& 25k   & \multicolumn{1}{c}{}    & 25k   & 5.4 & -0.3  & 17.2 & 12.0    & 0.8   & 23.1 & 12.2    & -0.2  & 27.9 & 9.1 & -0.2  & 37.9 & 3.8 & -0.2  & 22.2 \\
& 50k   & \multicolumn{1}{c}{}    & 1k    & N/A & N/A   & N/A  & N/A & N/A   & N/A  & 3.3 & 1.6   & 5.2  & 3.5 & 0.9   & 5.0  & 3.2 & 1.1   & 7.0  \\
& 50k   & \multicolumn{1}{c}{}    & 10k   & N/A & N/A   & N/A  & N/A & N/A   & N/A  & 9.0 & 0.8   & 16.8 & 7.9 & 0.8   & 29.6 & 6.5 & 1.0   & 21.9 \\ \vspace{.7em}
& 50k   & \multicolumn{1}{c}{}    & 25k   & N/A & N/A   & N/A  & N/A & N/A   & N/A  & 13.2    & -1    & 27.7 & 5.1 & 0.7   & 27.3 & 3.8 & 0.0   & 21.6 \\
\multirow{12}{*}{Bible}    & 1k    & \multirow{12}{*}{PMI/Gvt}   & 1k    & 0.2 & 7.0   & 0.7  & -0.2    & 3.4   & 1.3  & -0.4    & 6.4   & 0.2  & -0.4    & 8.0   & 1.4  & 0.0 & 5.9   & 0.2  \\
   & 1k    & & 10k   & 1.4 & 10.5  & 0.8  & -0.4    & 5.7   & -0.4 & 0.3 & 11.4  & -0.4 & -0.1    & 14.6  & 0.5  & 0.3 & 10.2  & 1.2  \\
   & 1k    & & 25k   & 0.5 & 10.9  & 0.2  & -0.2    & 7.2   & -0.2 & 0.5 & 11.8  & 0.2  & 0.0 & 11.9  & 0.3  & 0.3 & 11.7  & 1.0  \\
   & 1k    & & 50k   & N/A & N/A   & N/A  & N/A & N/A   & N/A  & 0.3 & 12.3  & 0.2  & 0.0 & 13.0  & -0.9 & -0.5    & 8.8   & 0.1  \\
   & 10k   & & 1k    & 0.0 & 9.7   & 0.9  & -1.4    & 4.4   & 0.1  & -1.6    & 6.6   & -1.1 & -0.9    & 8.1   & 0.7  & -0.4    & 5.7   & 1.1  \\
   & 10k   & & 10k   & 0.6 & 16.4  & -0.3 & -0.4    & 9.2   & -0.6 & -0.7    & 13.0  & -0.9 & -0.4    & 18.7  & -0.2 & -0.2    & 13.4  & 0.0  \\
   & 10k   & & 25k   & 0.4 & 16.5  & -0.5 & 0.7 & 11.1  & -0.1 & -0.5    & 16.4  & 0.9  & -0.1    & 20.6  & 0.6  & 0.0 & 15.5  & 0.0  \\
   & 10k   & & 50k   & N/A & N/A   & N/A  & N/A & N/A   & N/A  & 0.2 & 16.7  & 0.4  & 0.1 & 17.2  & -0.1 & -0.1    & 12.3  & 0.9  \\
   & 25k   & & 1k    & 0.6 & 9.3   & 2.6  & -0.3    & 4.6   & 1.6  & -1.7    & 6.9   & -0.6 & -0.2    & 6.9   & 2.4  & -0.6    & 5.9   & 1.4  \\
   & 25k   & & 10k   & 0.6 & 17.6  & -0.3 & 0.7 & 10.2  & 0.9  & -0.9    & 13.9  & -0.4 & -0.3    & 17.5  & 0.6  & 0.0 & 15.0  & 0.0  \\
   & 25k   & & 25k   & 1.3 & 14.9  & 1.6  & 0.1 & 10.3  & 0.0  & -1.2    & 16.2  & 0.7  & -0.1    & 20.8  & -0.2 & 0.5 & 12.1  & 1.0  \\
   & 25k   & & 50k   & N/A & N/A   & N/A  & N/A & N/A   & N/A  & -0.8    & 18.3  & -0.6 & -0.1    & 17.2  & -0.1 & 0.4 & 12.5  & 0.6  \\ \bottomrule
\end{tabular*}
\caption{Difference in spBLEU scores between multi-domain ITTL and single-domain ITTL. Here D1 refers to Domain 1 and D2 refers to Domain 2. For multi-domain ITTL D1 and D2 were combined and fine-tuned and then further fine-tuned on D2.FLORES=F Bible=B PMI=P Gvt=G}
\label{tab:mixedFT-single-domain}
\end{table*}
\begin{table}[!htb]
\centering
\uniformtablesize
\setlength\tabcolsep{1.5pt}
\renewcommand{\arraystretch}{1.05}

\begin{tabular*}{\textwidth}{@{\extracolsep{\fill}} l l lll lll lll @{}}

\toprule
\multirow{2}{*}{\textbf{Fine-tuning Set}}
& \multirow{2}{*}{\textbf{Fine-tuning Set size}} & \multicolumn{3}{c}{\textbf{\lang{ka}}} & \multicolumn{3}{c}{\textbf{\lang{si}}} & \multicolumn{3}{c}{\textbf{\lang{hi}}} \\  \cmidrule(r{\datgap}){3-5} 
\cmidrule(r{\datgap}){6-8} \cmidrule(r{\datgap}){9-11}
        &                        & \textbf{FLORES}  & \textbf{Bible} & \textbf{PMI}  & \textbf{FLORES} & \textbf{Bible} & \textbf{Gvt}  & \textbf{FLORES}  & \textbf{Bible} & \textbf{PMI} \\ \midrule
\multirow{2}{*}{CC}  &25k & 0.9 &0.0& 0.5& 8.1 &1.7& 13.2& 12.4 &4.4& 11.2 \\\vspace{.7em}
  & 100k       &4.7& 0.2& 3.7 &14.8& 4.1& 26.1& 21.9& 6.7& 17.2 \\

\multirow{4}{*}{PMI/Gvt} & 1k  &1.0	&0.1&	6.5	&3.9&0.5	&20.2&	7.1	&1.5& 18.2\\
  & 10k &10.4&	1.3	&29.4	&8.8&	1.0	&38.2	&14.7&	2.6	&31.4\\
  & 25k &14.1&	1.7	&33.6&	11.2	&1.2&	44.7&	17.3&	2.9	&35.2\\ \vspace{.7em}  & 50k  &N/A & N/A&N/A& 12.4&	1.6&	49.5	&19.0	&3.4&	37.0\\

\multirow{3}{*}{Bible}& 1k& 0.3	&8.2	&0.2&	0.9	&17.0	&1.3&	2.6	&14.0	&2.3\\
  & 10k&1.6	&22.6&	0.7&	1.8	&33.3	&0.8&	2.9	&26.9	&2.2\\
 & 25k  &2.2&	27.8&	0.9	&1.9	&37.9	&0.9&3.2&	31.4&	2.0\\\bottomrule
\end{tabular*}
\caption{Results (spBLEU) for vanilla fine-tuning.}
\label{tab:vanilla-fine-tuning}
\end{table}


\subsection{{Result Tables for XX-EN}\label{sec:app-b2}}

\begin{table}[!htb]
\centering
\uniformtablesize
\setlength\tabcolsep{2pt}
\renewcommand{\arraystretch}{1.05}

\begin{tabular*}{\textwidth}{@{\extracolsep{\fill}} l l l l lll lll @{}}

\toprule

\multirow{2}{*}{\textbf{D1}}           & \multirow{2}{*}{\textbf{D1 size}}    & \multirow{2}{*}{\textbf{D2}}    & \multirow{2}{*}{\textbf{D2 size}} & \multicolumn{3}{c}{\textbf{\lang{ka}}} & \multicolumn{3}{c}{\textbf{\lang{si}}} \\
\cmidrule(r{\datgap}){5-7} \cmidrule(lr){8-10}
  & & &  & \textbf{FLORES} & \textbf{Bible} & \textbf{PMI}  & \textbf{FLORES} & \textbf{Bible}  & \textbf{Gvt}  \\ \midrule
\multirow{9}{*}{CC} & \multirow{3}{*}{0k}   & \multirow{9}{*}{Bible}   & 1k          & 0.0    & 0.8   & 0.0  & 2.4    & 13.8   & 1.8  \\
 & & & 10k            & 0.9    & 19.0  & 0.6  & 2.5    & 30.3   & 1.4  \\ \vspace{0.5em}
 & & & 25k            & 1.4    & 28.5  & 1.1  & 2.5    & 35.1   & 1.1  \\
 & \multirow{3}{*}{25k}   & & 1k          & 0.7    & 2.2   & 0.4  & 6.1    & 15.3   & 9.4  \\
 & & & 10k            & 1.4    & 22.2  & 0.9  & 3.2    & 31.1   & 2.4  \\ \vspace{0.5em}
 & & & 25k            & 1.6    & 28.5  & 1.2  & 3.0    & 35.4   & 1.4  \\
 & \multirow{3}{*}{100k}   &           & 1k          & 5.1    & 10.7  & 4.4  & 9.4    & 17.0   & 16.9 \\
 & & & 10k            & 2.7    & 23.8  & 2.1  & 4.1    & 31.1   & 4.2  \\ 
 & & & 25k            & 2.2    & 29.0  & 1.7  & 3.6    & 35.3   & 2.2  \\ \bottomrule
\end{tabular*}

\caption{{Results for single-domain ITTL from other languages to English. Here D1 refers to Domain 1 and D2 refers to Domain 2.}}
\label{tab:single-domainITFT-xx-en}
\end{table}


\begin{table}[!htb]
\centering
\uniformtablesize
\setlength\tabcolsep{2pt}
\renewcommand{\arraystretch}{1.05}

\begin{tabular*}{\textwidth}{@{\extracolsep{\fill}} l l l l lll lll @{}}

\toprule
\multirow{2}{*}{\textbf{D1 + D2}}           & \multirow{2}{*}{\textbf{D1 + D2 size}} & \multirow{2}{*}{\textbf{D2}}           & \multirow{2}{*}{\textbf{D2 size}} & \multicolumn{3}{c}{\textbf{\lang{ka}}} & \multicolumn{3}{c}{\textbf{\lang{si}}} \\
\cmidrule(r{\datgap}){5-7} \cmidrule(r{\datgap}){8-10}
      &          &             &              &     \textbf{FLORES}  & \textbf{Bible}    & \textbf{PMI}    & \textbf{FLORES}  & \textbf{Bible}   & \textbf{Gvt} \\ \midrule
\multirow{8}{*}{CC + Bible}     & 25k + 1k          & \multirow{8}{*}{Bible}   & 1k    & 0.5    & 3.1   & 0.5    & 9.7  & 16.3    & 13.5 \\
                 & 25k + 10k         &              & 10k   & 2.5    & 22.7  & 1.7    & 8.8  & 30.0   & 13.4 \\ \vspace{.7em}
                 & 25k + 25k         &              & 25k   & 2.8    & 28.2  & 1.9    & 8.1  & 34.9    & 13.0 \\
                 & 100k + 1k         &              & 1k    & 0.6    & 6.3   & 0.7    & 13.9    & 17.5    & 21.3 \\
                 & 100k + 10k        &              & 10k   & 1.4    & 22.4  & 1.9    & 14.0   & 30.1    & 21.6 \\
                 & 100k + 25k        &              & 25k   & 3.4    & 26.1  & 3.4    & 13.4    & 33.3    & 20.7 \\ \bottomrule
\end{tabular*}
\caption{{Results for multi-domain FT from other languages to English. Here D1 refers to Domain 1 and D2 refers to Domain 2.}}
\label{tab:mixed-domainFT-xx-en}
\end{table}


\begin{table}[!htb]
\centering
\uniformtablesize
\setlength\tabcolsep{2pt}
\renewcommand{\arraystretch}{1.05}

\begin{tabular*}{\textwidth}{@{\extracolsep{\fill}} l l l l lll lll @{}}
\toprule

\multirow{2}{*}{\textbf{D1 + D2}}                          & \multirow{2}{*}{\textbf{D1 + D2 size}} & \multirow{2}{*}{\textbf{D2}}                       & \multirow{2}{*}{\textbf{D2 size}} & \multicolumn{3}{c}{\textbf{\lang{ka}}} & \multicolumn{3}{c}{\textbf{\lang{si}}} \\
\cmidrule(r{\datgap}){5-7} \cmidrule(r{\datgap}){8-10}
            &                         &                            &                             &           \textbf{FLORES}     & \textbf{Bible}    & \textbf{PMI}    & \textbf{FLORES}  & \textbf{Bible}   & \textbf{Gvt} \\ \midrule
\multirow{8}{*}{CC + Bible}           & 25k + 1k                   & \multirow{8}{*}{Bible}      & 1k       & 0.6       & 3        & 0.5    & 9.7     & 17      & 13.5 \\
                                      & 25k + 10k                  &                             & 10k      & 2.7       & 23.5     & 2      & 7.1     & 31.2    & 11.7 \\ \vspace{.7em}
                                      & 25k + 25k                  &                             & 25k      & 2.4       & 28.2     & 1.6    & 5.8     & 35.4    & 8.9 \\
                                      & 100k + 1k                  &                             & 1k       & 5.4       & 9.8      & 4.8    & 14      & 18.3    & 21.3 \\
                                      & 100k + 10k                 &                             & 10k      & 4.2       & 24.5     & 3.2    & 11.4    & 30.8    & 20.2 \\
                                      & 100k + 25k                 &                             & 25k      & 3.2       & 29.4     & 2.5    & 5.5     & 35.7    & 7.3 \\ \bottomrule
\end{tabular*}
\caption{{Results for multi-domain ITTL from other languages to English. Here D1 refers to Domain 1 and D2 refers to Domain 2.}}
\label{tab:mixed-domainITFT-xx-en}
\end{table}

\clearpage

\section{Graphical Results of Performance}%
\label{sec:graphicalresults}
\subsection{Single-domain ITTL Performance Plots\label{single-domain-itft}}
\begin{figure*}[h!]
     \centering
     \begin{subfigure}[b]{ \textwidth}
         \centering
         \includegraphics[width=\textwidth]{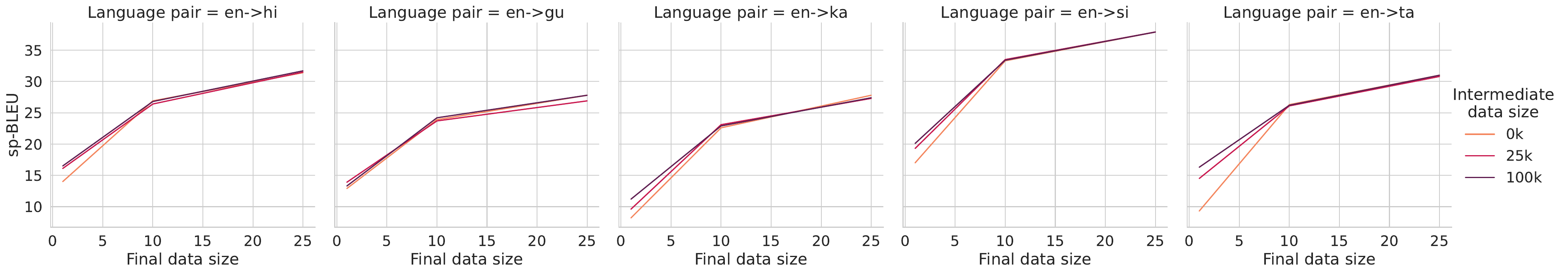}
         \vspace{-6mm}
        \caption{Single-domain ITTL where CC is used as Stage 1 data set, and Bible is used as Stage 2 data set. The spBLEU scores correspond to the test on Bible test set.}
        \vspace{4mm}
     \end{subfigure}
     \begin{subfigure}[b]{ \textwidth}
         \centering
         \includegraphics[width=\textwidth]{images/two-stage/two-stage-cc-pmo-pmo.pdf}
         \vspace{-6mm}
        \caption{Single-domain ITTL where CC is used as Stage 1 data set, and PMI/Gvt is used as Stage 2 data set. The spBLEU scores correspond to the test on the PMI/Gvt test set.}
        \vspace{4mm}
     \end{subfigure}
     \begin{subfigure}[b]{ \textwidth}
         \centering
         \includegraphics[width=\textwidth]{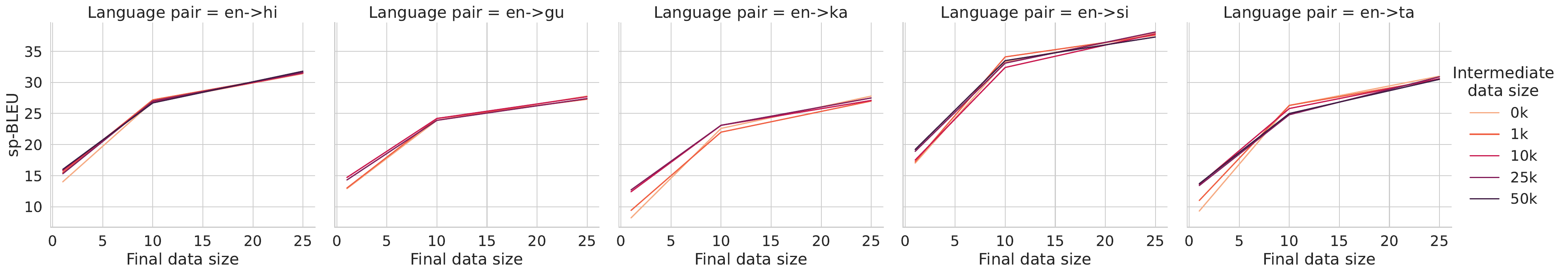}
         \vspace{-6mm}
        \caption{Single-domain ITTL where PMI/Gvt is used as Stage 1 data set, and Bible is used as Stage 2 data set. The spBLEU scores correspond to the test on Bible test set.}
        \vspace{4mm}
     \end{subfigure}
     \begin{subfigure}[b]{ \textwidth}
         \centering
         \includegraphics[width=\textwidth]{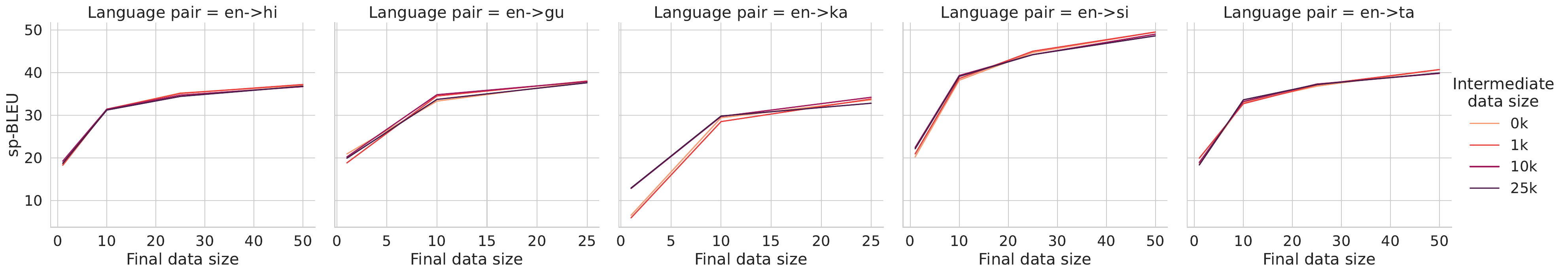}
        \caption{Single-domain ITTL where Bible is used as Stage 1 data set, and PMI/Gvt is used as Stage 2 data set. The spBLEU scores correspond to the test on the PMI/Gvt test set.}
     \end{subfigure}
     \caption{Single-domain ITTL results on in-domain cases.}
     \label{fig:singleITFT-indomain}
\end{figure*}

\begin{figure*}[h!]
     \centering
     \begin{subfigure}[b]{ \textwidth}
         \centering
         \includegraphics[width=\textwidth]{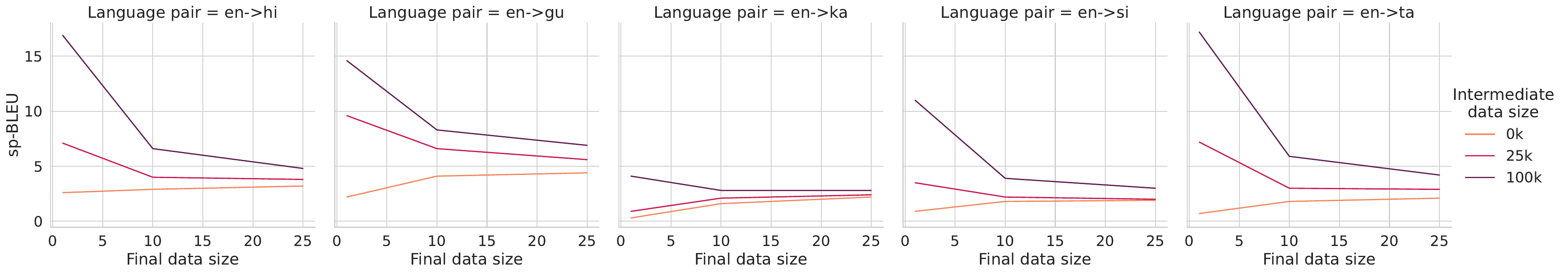}
         \vspace{-6mm}
        \caption{Single-domain ITTL where CC is used as Stage 1 data set, and Bible is used as Stage 2 data set. The spBLEU scores correspond to the test on the FLORES test set.}
        \vspace{4mm}
     \end{subfigure}
     \begin{subfigure}[b]{ \textwidth}
         \centering
         \includegraphics[width=\textwidth]{images/two-stage/two-stage-cc-pmo-flores.pdf}
         \vspace{-6mm}
        \caption{Single-domain ITTL where CC is used as Stage 1 data set, and PMI/Gvt is used as Stage 2 data set. The spBLEU scores correspond to the test on the FLORES test set.}
        \vspace{4mm}
     \end{subfigure}
     \begin{subfigure}[b]{ \textwidth}
         \centering
         \includegraphics[width=\textwidth]{images/two-stage/two-stage-pmo-bible-flores.pdf}
         \vspace{-6mm}
        \caption{Single-domain ITTL where PMI/Gvt is used as Stage 1 data set, and Bible is used as Stage 2 data set. The spBLEU scores correspond to the test on the FLORES test set.}
        \vspace{4mm}
     \end{subfigure}
     \begin{subfigure}[b]{ \textwidth}
         \centering
         \includegraphics[width=\textwidth]{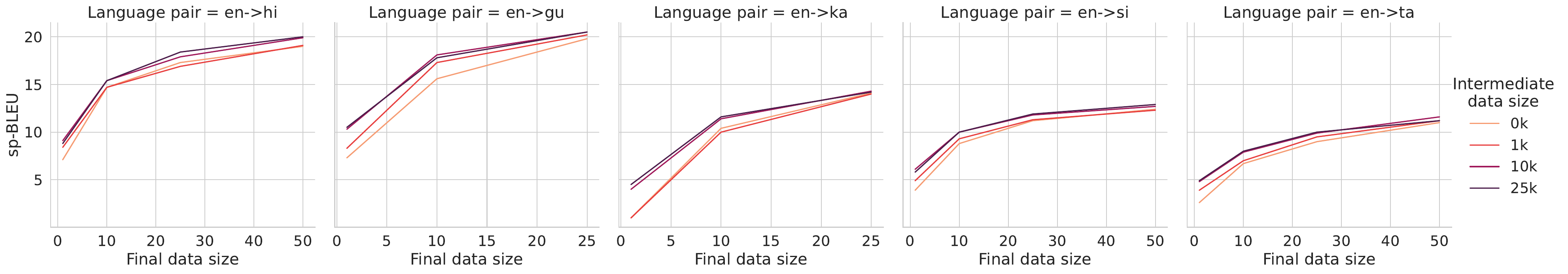}
         \vspace{-6mm}
        \caption{Single-domain ITTL where Bible is used as Stage 1 data set, and PMI/Gvt is used as Stage 2 data set. The spBLEU scores correspond to the test on the FLORES test set.}
        \vspace{4mm}
     \end{subfigure}
          \begin{subfigure}[b]{ \textwidth}
         \centering
         \includegraphics[width=\textwidth]{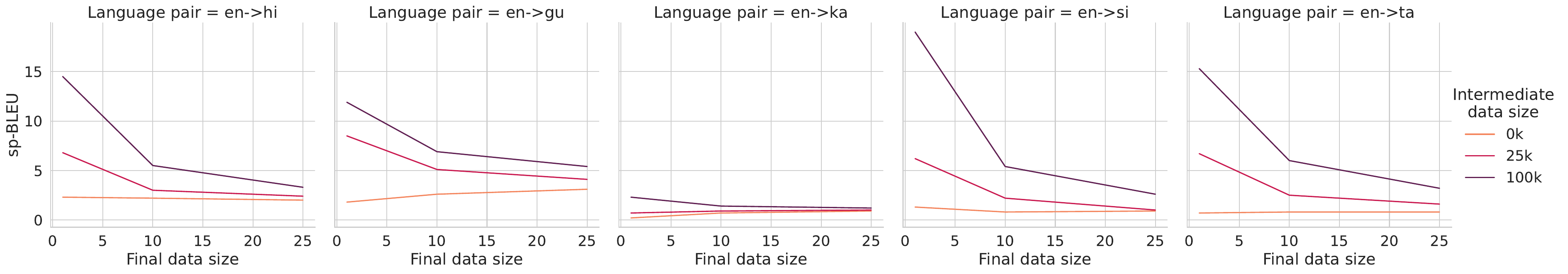}
        \vspace{-6mm}
        \caption{Single-domain ITTL where CC is used as Stage 1 data set, and Bible is used as Stage 2 data set. The spBLEU scores correspond to the test on the PMI/Gvt test set.}
        \vspace{4mm}
     \end{subfigure}
     \begin{subfigure}[b]{ \textwidth}
         \centering
         \includegraphics[width=\textwidth]{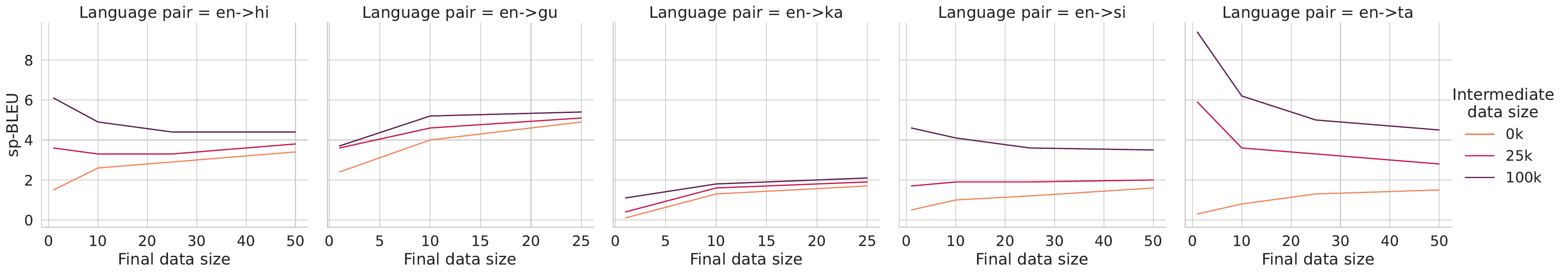}
        \caption{Single-domain ITTL where CC is used as Stage 1 data set, and PMI/Gvt is used as Stage 2 data set. The spBLEU scores correspond to the test on Bible test set.}
     \end{subfigure}
     \vspace{-6mm}
     \caption{Single-domain ITTL results for out-domain cases.}
     \vspace{4mm}
     \label{fig:singleITFT-outdomain}
\end{figure*}

Figure~\ref{fig:singleITFT-indomain} and Figure~\ref{fig:singleITFT-outdomain} show the single-domain ITTL results for in-domain and out-domain cases (respectively).

\clearpage

\subsection{Muti-domain FT Performance Plots\label{mixed_ft}}
\begin{figure*}[h!]
     \centering
     \begin{subfigure}[b]{ \textwidth}
         \centering
         \includegraphics[width=\textwidth]{images/mixed-domain/stage1/mixed-domain1-cc-pmo-pmo.pdf}
         \vspace{-6mm}
        \caption{Multi-domain FT where a mix of CC and PMI/Gvt are used are used to fine-tune the model. The spBLEU scores correspond to the test on the PMI/Gvt test set.}
        \vspace{4mm}
     \end{subfigure}
     \begin{subfigure}[b]{ \textwidth}
         \centering
         \includegraphics[width=\textwidth]{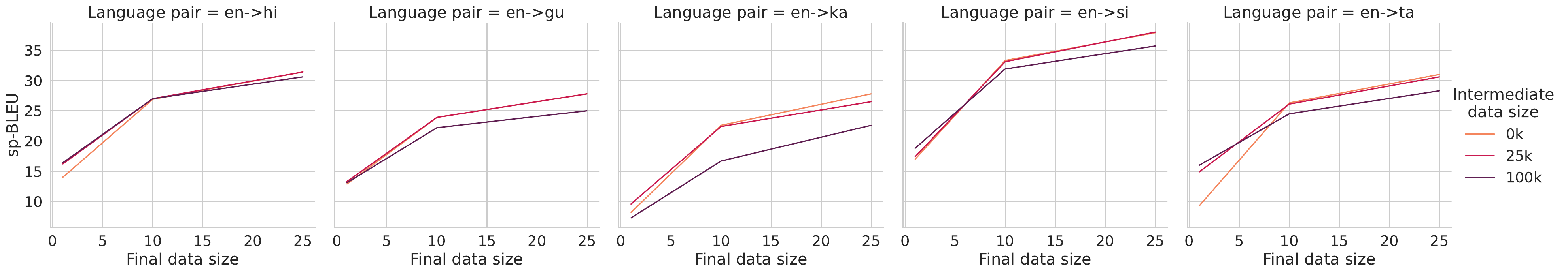}
         \vspace{-6mm}
        \caption{Multi-domain FT where a mix of CC and Bible are used to fine-tune the model. The spBLEU scores correspond to the test on Bible test set.}
        \vspace{4mm}
     \end{subfigure}
     \begin{subfigure}[b]{ \textwidth}
         \centering
         \includegraphics[width=\textwidth]{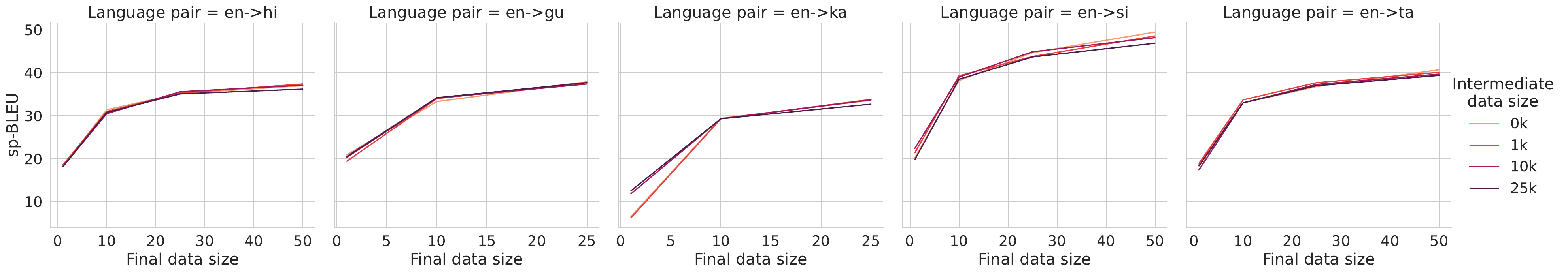}
         \vspace{-6mm}
        \caption{Multi-domain FT where a mix of PMI/Gvt and Bible are used to fine-tune the model. The spBLEU scores correspond to the test on the PMI/Gvt test set.}
        \vspace{4mm}
     \end{subfigure}
     \begin{subfigure}[b]{ \textwidth}
         \centering
         \includegraphics[width=\textwidth]{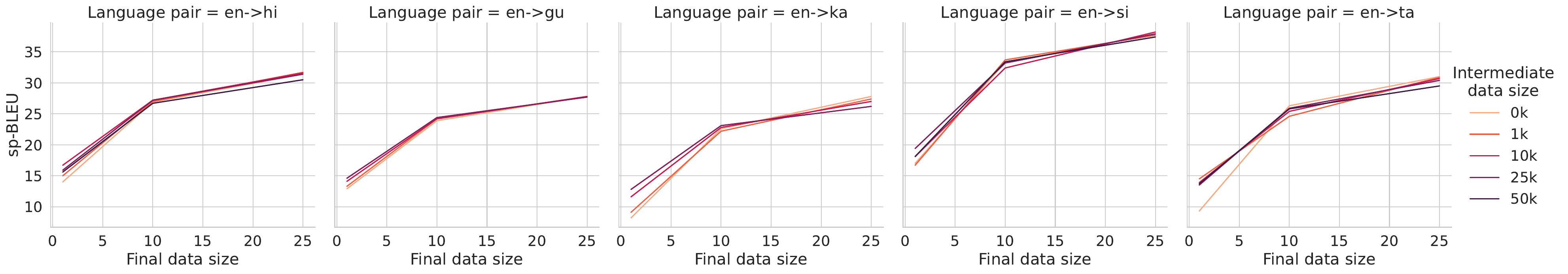}
        \caption{Multi-domain FT where a mix of PMI/Gvt and Bible are used to fine-tune the model. The spBLEU scores correspond to the test on Bible test set.}
     \end{subfigure}
     \vspace{-6mm}
     \caption{Results for multi-domain FT for in-domain cases.}
     \vspace{4mm}
     \label{fig:mixedFT-indomain}
\end{figure*}

\begin{figure*}[h!]
     \centering
     \begin{subfigure}[b]{ \textwidth}
         \centering
         \includegraphics[width=\textwidth]{images/mixed-domain/stage1/mixed-domain1-cc-pmo-flores.pdf}
         \vspace{-6mm}
        \caption{Multi-domain FT where a mix of CC and PMI/Gvt are used to fine-tune the model. The spBLEU scores correspond to the test on the FLORES test set.}
        \vspace{4mm}
     \end{subfigure}
     \begin{subfigure}[b]{ \textwidth}
         \centering
         \includegraphics[width=\textwidth]{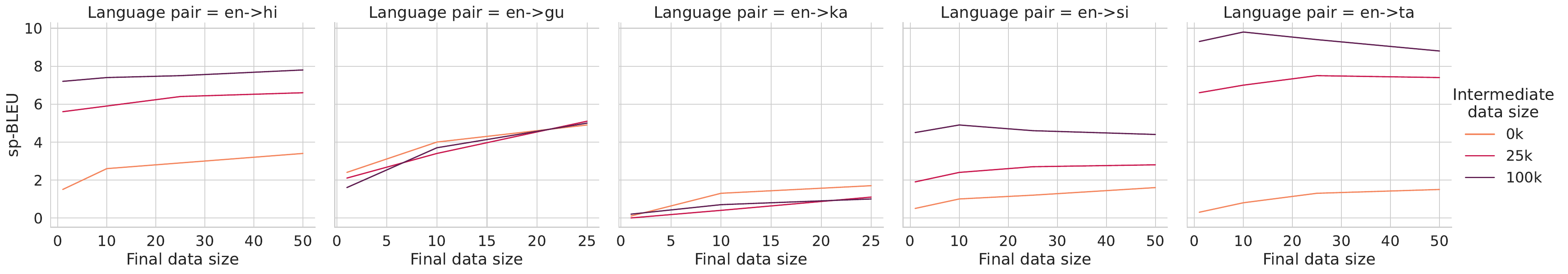}
         \vspace{-6mm}
        \caption{Multi-domain FT where a mix of CC and PMI/Gvt are used to fine-tune the model. The spBLEU scores correspond to the test on Bible test set.}
        \vspace{4mm}
     \end{subfigure}
     \begin{subfigure}[b]{ \textwidth}
         \centering
         \includegraphics[width=\textwidth]{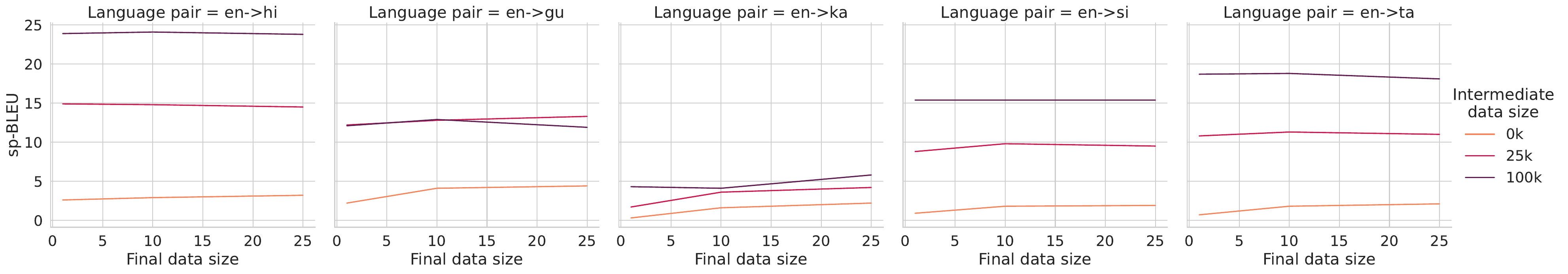}
         \vspace{-6mm}
        \caption{Multi-domain FT where a mix of CC and Bible are used to fine-tune the model. The spBLEU scores correspond to the test on the FLORES test set.}
        \vspace{4mm}
     \end{subfigure}
     \begin{subfigure}[b]{ \textwidth}
         \centering
         \includegraphics[width=\textwidth]{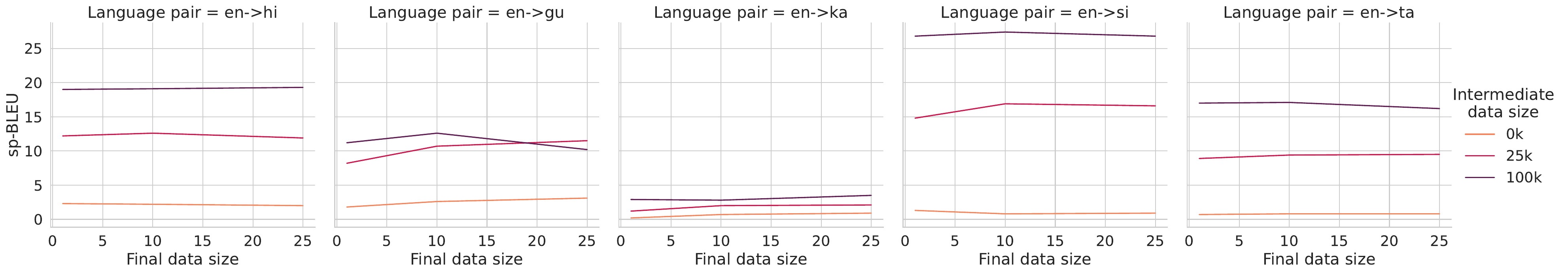}
         \vspace{-6mm}
        \caption{Multi-domain FT where a mix of CC and Bible are used to fine-tune the model. The spBLEU scores correspond to the test on the PMI/Gvt test set.}
        \vspace{4mm}
     \end{subfigure}
     \begin{subfigure}[b]{ \textwidth}
         \centering
         \includegraphics[width=\textwidth]{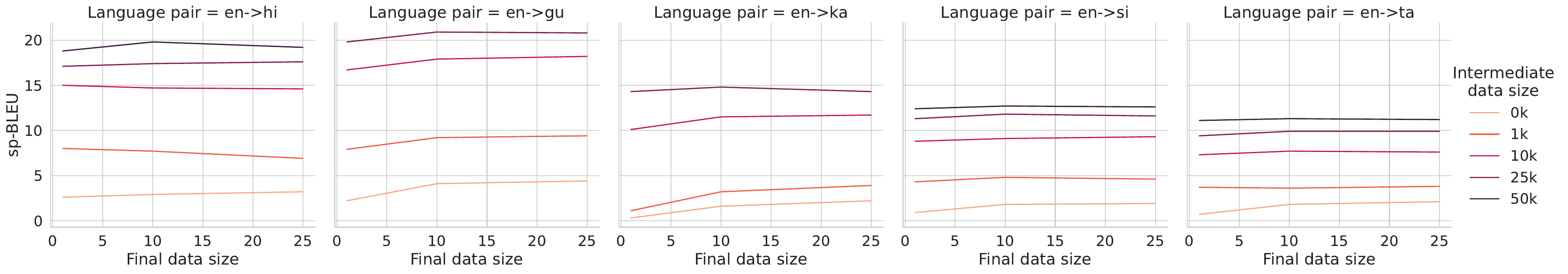}
        \caption{Multi-domain FT where a mix of PMI/Gvt and Bible are used to fine-tune the model. The spBLEU scores correspond to the test on the FLORES test set.}
     \end{subfigure}
     \vspace{-6mm}
     \caption{Results for multi-domain FT for out-domain cases.}
     \vspace{4mm}
     \label{fig:mixedFT-outdomain}
\end{figure*}

Figure~\ref{fig:mixedFT-indomain} and Figure~\ref{fig:mixedFT-outdomain} show results of multi-domain FT for in-domain and out-domain cases (respectively). 

\clearpage

\subsection{Multi ITTL Performance Plots}
\begin{figure*}[h!]
     \centering
     \begin{subfigure}[b]{ \textwidth}
         \centering
         \includegraphics[width=\textwidth]{images/mixed-domain/stage2/mixed-domain-cc-pmo-pmo.pdf}
         \vspace{-6mm}
        \caption{Multi-domain ITTL where a mix of CC and PMI/Gvt is used as Stage 1 data set, and PMI/Gvt is used as Stage 2 data set. The spBLEU scores correspond to the test on the PMI/Gvt test set.}
        \vspace{4mm}
     \end{subfigure}
     \begin{subfigure}[b]{ \textwidth}
         \centering
         \includegraphics[width=\textwidth]{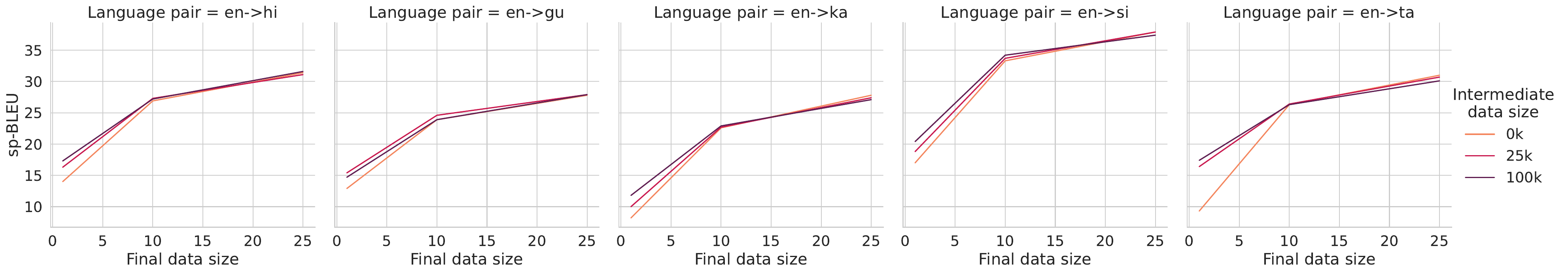}
         \vspace{-6mm}
        \caption{Multi-domain ITTL where a mix of CC and Bible are used as Stage 1 data set, and Bible is used as Stage 2 data set. The spBLEU scores correspond to the test on Bible test set.}
        \vspace{4mm}
     \end{subfigure}
     \begin{subfigure}[b]{ \textwidth}
         \centering
         \includegraphics[width=\textwidth]{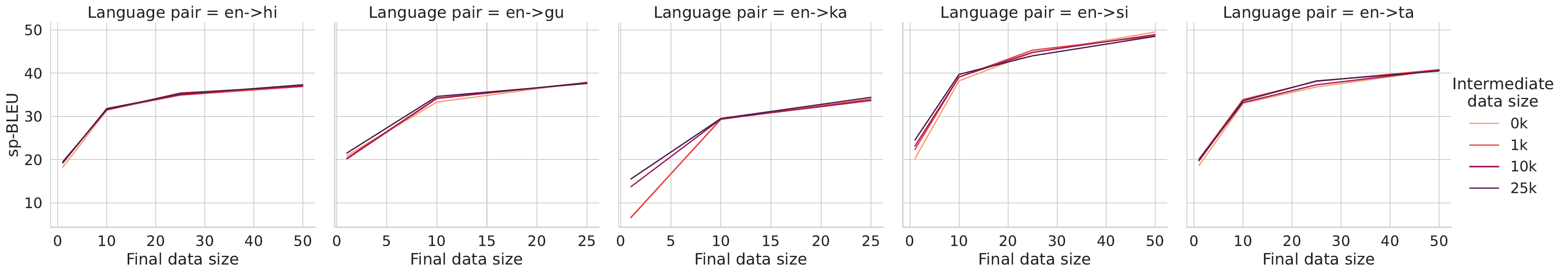}
         \vspace{-6mm}
        \caption{Multi-domain ITTL where a mix of Bible and PMI/Gvt is used as Stage 1 data set, and PMI/Gvt is used as Stage 2 data set. The spBLEU scores correspond to the test on the PMI/Gvt test set.}
        \vspace{4mm}
     \end{subfigure}
     \begin{subfigure}[b]{ \textwidth}
         \centering
         \includegraphics[width=\textwidth]{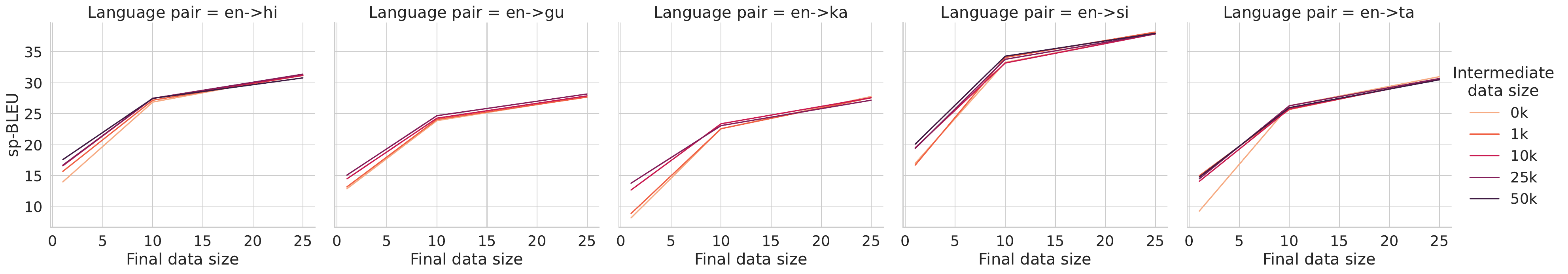}
         \vspace{-6mm}
        \caption{Multi-domain ITTL where a mix of PMI/Gvt and Bible are used as Stage 1 data set and Bible is used as Stage 2 data set. The spBLEU scores correspond to the test on Bible test set.}
     \end{subfigure}
     \caption{Results for multi-domain ITTL for in-domain cases.}
     \label{fig:mixedITFT-indomain}
\end{figure*}

\begin{figure*}[h!]
     \centering     
     \begin{subfigure}[b]{ \textwidth}
         \centering
         \includegraphics[width=\textwidth]{images/mixed-domain/stage2/mixed-domain-cc-pmo-flores.pdf}
         \vspace{-6mm}
        \caption{Multi-domain ITTL where a mix of CC and PMI/Gvt is used as Stage 1 data set, and PMI/Gvt is used as Stage 2 data set. The spBLEU scores correspond to the test on the FLORES test set.}
        \vspace{4mm}
     \end{subfigure}
     \begin{subfigure}[b]{ \textwidth}
         \centering
         \includegraphics[width=\textwidth]{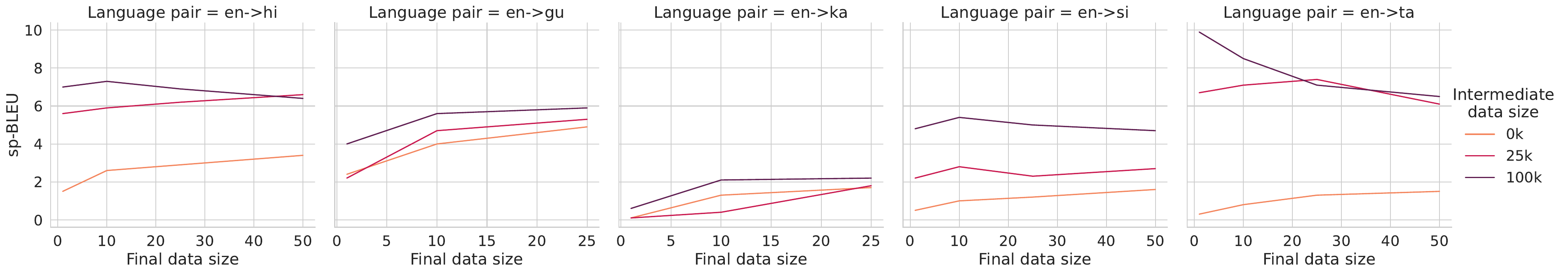}
         \vspace{-6mm}
        \caption{Multi-domain ITTL where a mix of CC and PMI/Gvt is used as Stage 1 data set, and PMI/Gvt is used as Stage 2 data set. The spBLEU scores correspond to the test on Bible test set.}
        \vspace{4mm}
     \end{subfigure}
     \begin{subfigure}[b]{ \textwidth}
         \centering
         \includegraphics[width=\textwidth]{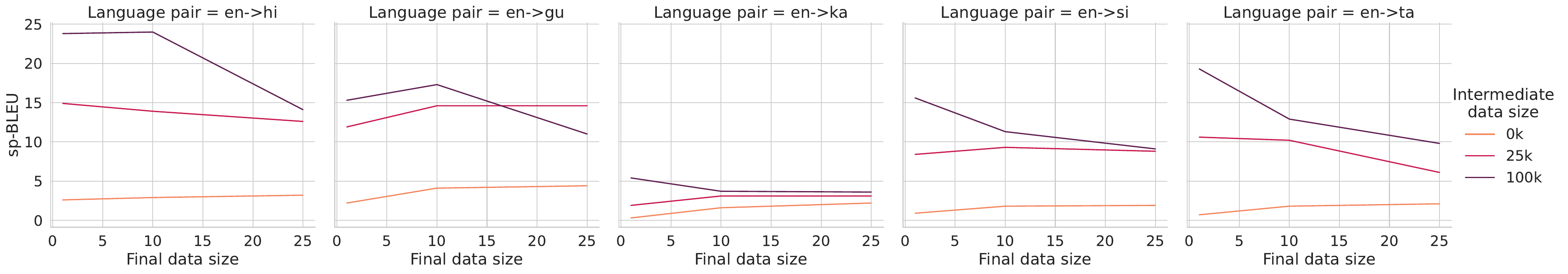}
         \vspace{-6mm}
        \caption{Multi-domain ITTL where a mix of CC and Bible are used as Stage 1 data set, and Bible is used as Stage 2 data set. The spBLEU scores correspond to the test on the FLORES test set.}
        \vspace{4mm}
     \end{subfigure}
     \begin{subfigure}[b]{ \textwidth}
         \centering
         \includegraphics[width=\textwidth]{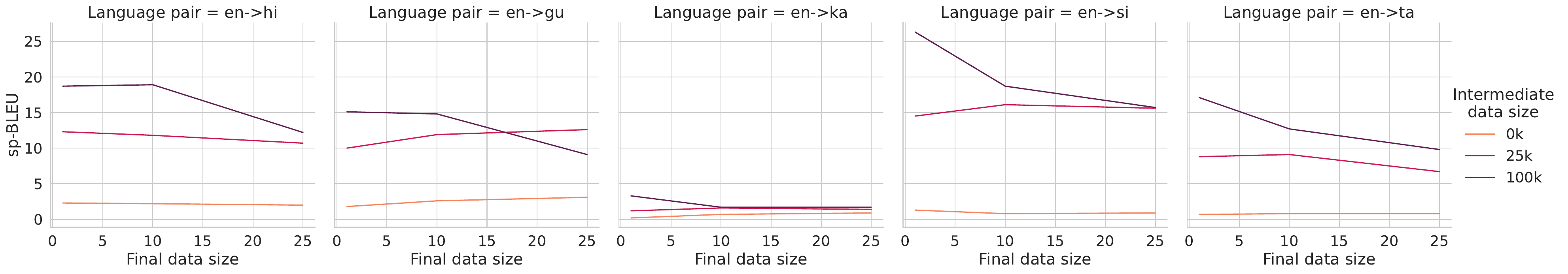}
         \vspace{-6mm}
        \caption{Multi-domain ITTL where a mix of CC and Bible is used as Stage 1 data set, and Bible is used as Stage 2 data set. The spBLEU scores correspond to the test on the PMI/Gvt test set.}
        \vspace{4mm}
     \end{subfigure}
     \begin{subfigure}[b]{ \textwidth}
         \centering
         \includegraphics[width=\textwidth]{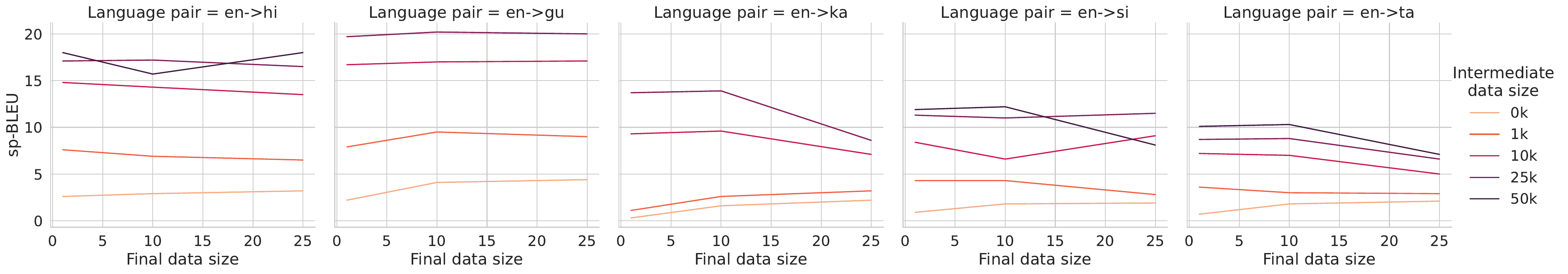}
         \vspace{-6mm}
        \caption{Multi-domain ITTL where a mix of Bible and PMI/Gvt is used as Stage 1 data set, and Bible is used as Stage 2 data set. The spBLEU scores correspond to the test on the FLORES test set.}
        \vspace{4mm}
     \end{subfigure}
     \begin{subfigure}[b]{ \textwidth}
         \centering
         \includegraphics[width=\textwidth]{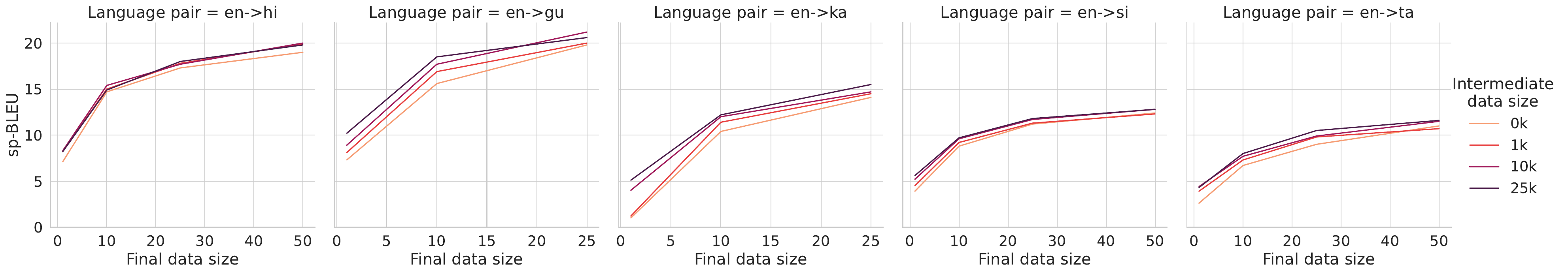}
         \vspace{-6mm}
        \caption{Multi-domain ITTL where a mix of Bible and PMI/Gvt is used as Stage 1 data set, and PMI/Gvt is used as Stage 2 data set. The spBLEU scores correspond to the test on the FLORES test set.}
     \end{subfigure}
     \caption{Results for Multi-domain ITTL for out-domain cases.}
     \label{fig:mixedITFT-outdomain}
\end{figure*}

Figure~\ref{fig:mixedITFT-indomain} and Figure~\ref{fig:mixedITFT-outdomain} show results of multi-domain FT for in-domain and out-domain cases (respectively). 

\end{document}